\documentclass{article}

\usepackage{arxiv}

\usepackage[utf8]{inputenc} 
\usepackage[T1]{fontenc}    
\usepackage{hyperref}       
\usepackage{url}            
\usepackage{booktabs}       
\usepackage{amsfonts}       
\usepackage{nicefrac}       
\usepackage{microtype}      
\usepackage{graphicx}
\usepackage{natbib}
\usepackage{doi}
\usepackage{multirow}
\usepackage{xurl}

\title{A survey on attention mechanisms for medical applications: are we moving towards better algorithms?}

\author{ \href{https://orcid.org/0000-0003-4744-9174}{\includegraphics[scale=0.06]{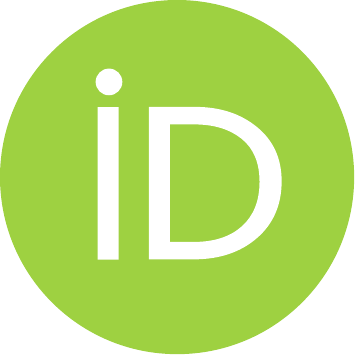}\hspace{1mm}Tiago Gonçalves} \\
	INESC TEC and University of Porto \\
	Porto, Portugal\\
	\texttt{tiago.f.goncalves@inesctec.pt} \\
	\And
	\href{https://orcid.org/0000-0002-2302-8597}{\includegraphics[scale=0.06]{orcid.pdf}\hspace{1mm}Isabel Rio-Torto} \\
	INESC TEC and University of Porto \\
	Porto, Portugal\\
	\texttt{isabel.riotorto@inesctec.pt} \\
	\And
	\href{https://orcid.org/0000-0002-4050-7880}{\includegraphics[scale=0.06]{orcid.pdf}\hspace{1mm}Luís F. Teixeira} \\
	INESC TEC and University of Porto \\
	Porto, Portugal\\
	\texttt{luis.f.teixeira@inesctec.pt} \\
	\And
	\href{https://orcid.org/0000-0002-3760-2473}{\includegraphics[scale=0.06]{orcid.pdf}\hspace{1mm}Jaime S. Cardoso} \\
	INESC TEC and University of Porto \\
	Porto, Portugal\\
	\texttt{jaime.cardoso@inesctec.pt} \\
}

\date{}

\renewcommand{\headeright}{}
\renewcommand{\undertitle}{}
\renewcommand{\shorttitle}{A survey on attention mechanisms for medical applications: are we moving towards better algorithms?}

\hypersetup{
pdftitle={A survey on attention mechanisms for medical applications: are we moving towards better algorithms?},
pdfsubject={q-bio.NC, q-bio.QM},
pdfauthor={Tiago Gonçalves, Isabel Rio-Torto, Luís F. Teixeira, Jaime S. Cardoso},
pdfkeywords={artificial intelligence, attention mechanisms, computer vision, deep learning, medical applications, medical image analysis, transformers},
}

\begin{document}
\maketitle

\begin{abstract}
The increasing popularity of attention mechanisms in deep learning algorithms for computer vision and natural language processing made these models attractive to other research domains. In healthcare, there is a strong need for tools that may improve the routines of the clinicians and the patients. Naturally, the use of attention-based algorithms for medical applications occurred smoothly. However, being healthcare a domain that depends on high-stake decisions, the scientific community must ponder if these high-performing algorithms fit the needs of medical applications. With this motto, this paper extensively reviews the use of attention mechanisms in machine learning (including Transformers) for several medical applications. This work distinguishes itself from its predecessors by proposing a critical analysis of the claims and potentialities of attention mechanisms presented in the literature through an experimental case study on medical image classification with three different use cases. These experiments focus on the integrating process of attention mechanisms into established deep learning architectures, the analysis of their predictive power, and a visual assessment of their saliency maps generated by \textit{post-hoc} explanation methods. This paper concludes with a critical analysis of the claims and potentialities presented in the literature about attention mechanisms and proposes future research lines in medical applications that may benefit from these frameworks.
\end{abstract}


\section*{Introduction}\label{sec:introduction}
The concept of \textit{attention} is not new and has been part of multidisciplinary debates across psychology, physiology, and neuroscience~\cite{koch1987shifts}. The inspiration behind the translation of this concept into artificial intelligence (AI) algorithms comes from biological vision systems, which appear to employ a serial computational strategy when inspecting complex scenes~\cite{itti2000saliency}. Historically, the pursuit of an \textit{attention mechanism} seems to be deeply related to the problem of \textit{visual search}, where the main objective may be to find a small object in a disorderly environment (e.g., a pen on a desk). A possible solution to this problem is to use images with high resolution and wide field-of-view. However, this results in very high dimensional inputs~\cite{minut2001reinforcement}. Hence, it was essential to devise a strategy capable of selecting the most important parts of the input, deciding where to look next during the processing pipeline, and managing limited computational resources. These aspects motivated the study of \textit{selective visual attention}, which deals with the computational complexity of computer vision tasks and often requires several basic components, such as the selection of a region of interest in the image and the selection of the dimension of features and values of interest~\cite{tsotsos1995modeling}. Besides, theoretically, such mechanisms have the potential to guide the research into focusing more on sparse local models~\cite{ballard1991animate}. Nevertheless, the scientific community proposed several research lines that certainly influenced the modern methodologies of attention-based mechanisms for deep learning. Bandera et al.~\cite{bandera1996residual} and Minut and Mahadevan~\cite{minut2001reinforcement} proposed \textit{reinforcement learning} policies to regularise the training of computer vision systems in the task of target recognition, Koch and Ullman~\cite{koch1987shifts} and Tsotsos et al.~\cite{tsotsos1995modeling} proposed a set of rules that promote visual attention via selective tuning and shifts in the processing focus~\cite{koch1987shifts} during the analysis of a visual scene, and Itti and Koch~\cite{itti2000saliency} addressed this challenge using a \textit{saliency map} based search mechanism. Recently, the increase of available computational power and the democratised access to \textit{big data} allowed the resurgence of deep learning algorithms and their applications~\cite{lecun_deep_2015, schmidhuber_deep_2015}. With this paradigm shift, current methods now seek to integrate \textit{learnable} attention mechanisms on an end-to-end basis~\cite{jetley2018learn}, with interesting impacts in diverse applications~\cite{makarov2022self}.

As the available literature on attention mechanisms continues to grow, the first reviews and surveys on the topic were published and gained visibility. Hence, it is relevant to introduce the reader to the context of such studies and their content. Besides, this overview helps the reader gain intuition about the work developed in this research field while understanding the research questions that this survey aims to answer. Borji and Itti~\cite{borji2012state} reviewed several models of visual attention based on non-deep learning methods and presented a taxonomy indicating their approaches an capabilities. Cho et al.~\cite{cho2015describing} presented an early study on the use of attention mechanisms in convolutional neural networks (CNNs) and recurrent neural networks (RNNs) with a special focus on \textit{soft-attention} for applications such as neural machine translation, image captioning, video description generation or end-to-end neural speech recognition. Wang and Tax~\cite{wang2016survey} published a survey that explored the use of attention mechanisms in RNNs for sequence-to-sequence challenges. Chaudhari et al.~\cite{chaudhari2019attentive} presented a survey of attention mechanisms for natural language processing, wherein they proposed a taxonomy and reviewed several neural network architectures in which attention has been incorporated while discussing applications in which modelling attention has shown significant impact. Yang~\cite{yang2020overview} performed a brief overview about attention mechanisms for computer vision and introduced a preliminary taxonomy. Lindsay~\cite{lindsay2020attention} published a review on attention mechanisms that provides intuition about the definition of attention in the neuroscience and psychology literature and covers several use cases of attention in machine learning, indicating their biological counterparts where they exist.

In 2017, Vaswani et al.~\cite{vaswani2017attention} proposed the~\textit{Transformer}, a deep neural network architecture composed solely of attention operations that discarded recurrence and convolutions entirely. In the original paper, the authors applied this architecture to the task of neural machine translation and explored the use of attention operations as a means of reducing computational complexity when compared to recurrent or convolution operations. The establishment of a similar architecture for computer vision occurred almost naturally, when Dosovitskiy et al.~\cite{dosovitskiy2020image} proposed the~\textit{Vision Transformer} architecture. Han et al. was one of the pioneers in analysing the advantages and disadvantages associated with the use of Transformer-based architectures in computer vision with a particular interest in the \textit{self-attention} mechanism~\cite{han2020survey}. Khan et al.~\cite{khan2021transformers} also provided a  comprehensive overview of Transformer-based models from the perspective of their application and architectural design. Guo et al.~\cite{guo2021attention} published a comprehensive and systematic review of attention mechanisms for computer vision and created a taxonomy that categorises attention mechanisms according to their data domain. Xu et al.~\cite{xu2022transformers} revisited the Transformer architectures with particular emphasis on low-level vision and generation. Recently, Shamshad et al.~\cite{shamshad2022transformers} explored the use of Transformer-based architectures for medical image analysis and published a pioneering review that provides insight on the applications of these algorithms in several dimensions of the clinical daily routine. Almost all of the previous papers focus on the current state-of-the-art, performing a comparative analysis based solely on published results, thus, anticipating future and open challenges from a theoretical perspective. 

In this survey, we perform an extensive review of attention mechanisms for medical applications (including Transformer-based architectures) with a theoretical discussion and a strong focus on the methodologies and their applications. In other works of the same genre, the authors usually focus on enumerating state-of-the-art strategies with their reported results. Besides, to our knowledge, this is the first work that approaches the topic of attention mechanisms from a broad perspective since we establish the connection between the natural language processing and computer vision domains, given the multi-modality property of medical data. Also, we complement this discussion with the Transformer-based architectures, as they currently are the most popular attention-based models. Hence, we approach this topic with a critical view and ask the following research questions:
\begin{enumerate}
	\item Will attention mechanisms automatically improve the predictive power of deep learning algorithms for medical image applications?

	\item What is the impact of integrating attention mechanisms on model complexity?
	
	\item Can we improve the degree of interpretability of deep learning models solely through attention mechanisms?

	\item How practical is it to design and build attention mechanisms for deep learning applications?
\end{enumerate}

To answer these, we develop an experimental protocol with three case studies on medical image classification, using established deep learning architectures with and without state-of-the-art attention blocks and a Transformer-based model. We address the open challenges of interpretability and transparency with the help of \textit{post-hoc} explanation methods and perform a visual analysis of the impact of different attention mechanisms in these outputs. We consider this survey a timely opportunity for the community to start questioning the extent of several claims made in the literature. In addition, our experimental protocol works as a reliable proxy for this analysis.

The main contributions of this paper are:
\begin{enumerate}
	
	\item An extensive review of the use of attention mechanisms in machine learning methods (including Transformers) for several medical applications based on the types of tasks that can integrate several work pipelines of the medical domain: medical image classification, medical image segmentation, medical report understanding, and other tasks (medical image detection, medical image reconstruction, medical image retrieval, medical signal processing, and physiology and pharmaceutical research);
    
	\item An experimental case study on medical image classification with three different use cases that focus on the integration of attention mechanisms into established deep learning architectures, the analysis of their predictive power, and a visual assessment of their saliency maps generated by \textit{post-hoc} explanation methods;
    
	\item A critical analysis of the claims and potentialities of attention mechanisms presented in the literature;
	
	\item A discussion on the future challenges of medical applications and how they may benefit from attention mechanisms.

\end{enumerate}

Besides the Introduction, the remainder of this article is organised as follows: Background introduces a historical overview and the fundamental concepts of attention mechanisms in machine learning, Literature Review presents an exhaustive review of the integration of attention mechanisms in algorithms for medical applications, Case Studies details the experimental work that accompanies this paper, and Conclusions and Future Challenges summarises the main conclusions of this survey and points to future directions towards the study of attention-based algorithms for medical applications.

\section*{Background}\label{sec:background}
This section presents a historical overview and the fundamental concepts regarding attention mechanisms in machine learning.

Although there is no clear nor unified concept for~\textit{attention}, we can refer to it as the ability to flexibly manage limited computational resources. This research topic spans several domains of knowledge and has been studied in conjunction with many other topics in neuroscience and psychology~\cite{chun2011taxonomy}, including awareness, vigilance, saliency, executive control, learning, and, more recently, artificial intelligence. Therefore, we have different ways of perceiving the properties of attention mechanisms at biological, psychological or computational levels~\cite{lindsay2020attention}. From the perspective of the domains of \textit{arousal}, \textit{alertness} or \textit{vigilance}, attention could be described as a general measure of alertness or the ability to engage with the surroundings~\cite{oken2006vigilance}. From the perspective of \textit{sensorial stimuli}, attention is often deployed as an alert subject to specific sensory input,  allowing tight control over both the stimuli and the locus of attention (e.g., selective audio attention, visual attention)~\cite{kanwisher2000visual,bronkhorst2015cocktail,hutmacher2019there}. From the perspective of \textit{executive control}~\cite{miller2014neural}, the assumption is that there are multiple simultaneous competing tasks and that a central controller is needed to decide which to engage in and when, hence, attention can be thought of as the output of executive control, since the executive control system must select the targets of attention and communicate that to the systems responsible for implementing it~\cite{lindsay2020attention}. From the perspective of \textit{memory}, attention is understood as the ability of the brain to select the subset of information that is well-matched to the needs of the memory system, thus being a choice regarding the deployment of limited resources~\cite{aly2017hippocampal,lindsay2020attention}. From the perspective of \textit{artificial intelligence}, the first successful implementations were accomplished with encoder-decoder structures in deep neural network architectures, where the main goal is to learn the best attention weights that connect the encoder to the decoder~\cite{cho2015describing}.

In the context of deep learning, RNNs were the pioneering attention-based structures. This is mainly due to their ability to learn and process sequential data (e.g., data with a temporal component), which makes RNNs a class of machines with dynamic states (i.e.,  their state depends on both the input to the system and the current state), such that a signal received at a given moment can alter the behaviour of the network at a much later point in time~\cite{graves2014neural}. These properties allow RNNs to process variable-length structures without further modification, a property that has been extensively explored in several cognitive problems such as speech recognition~\cite{graves2013speech, graves2014towards}, text generation~\cite{sutskever2011generating}, handwriting generation~\cite{graves2013generating} and machine translation~\cite{sutskever2014sequence}. The work described in~\cite{bahdanau2014neural} is perhaps the most classic example of the design and implementation of computational attention mechanisms with RNNs. In the original paper, the authors focused on the task of neural machine translation and proposed an RNN-based encoder-decoder architecture in which the task of the encoder is to compress the input data (in this case, an English sentence), while the task of the decoder is to receive and decode this data to achieve a meaningful translation (in this case, a French sentence). The principal motivation for introducing an attention mechanism in this learning pipeline is related to the difficulty of achieving a perfect alignment between the input and output sequences, which is an almost impossible goal in most cases. The role of the attention mechanism is to learn weights that make the decoder focus on the relevant parts of the input sequences to produce meaningful outputs. At the time of publication, this methodology improved the state-of-the-art and paved the way for new research questions and exploration of novel attention-based frameworks.

The study of attention mechanisms comprises two main research lines: \textit{language, text, and speech}; and \textit{computer vision}. Since the clinical practice (the subject of this survey) involves multiple modalities of data, we examine these two main dimensions to help the reader become familiar with the various nomenclatures, taxonomies, and practices. We believe this introductory approach is important to understand the context of the discussion presented in the following sections.

\subsubsection*{Attention mechanisms for language, text, and speech}
\begin{figure}[ht]
    \centering
    \includegraphics[scale=0.6]{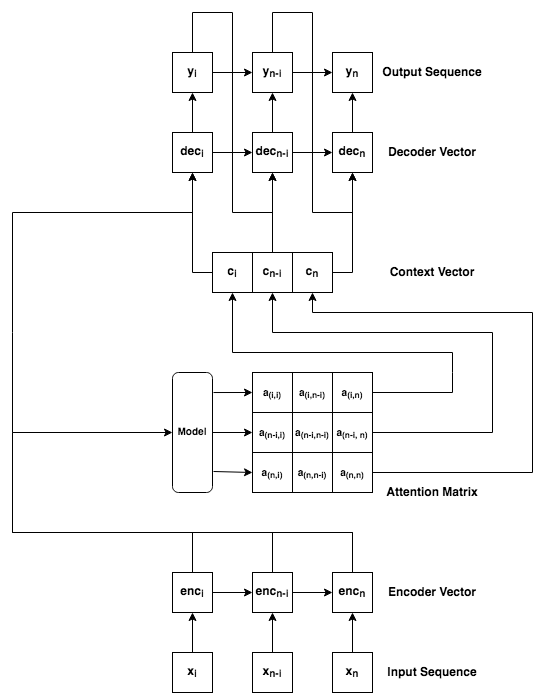}
    \caption{Block diagram of the general attention mechanism for language, text and speech, according to~\cite{chaudhari2019attentive}.}
    \label{fig:att-mech-nlp}
\end{figure}

An insightful and comprehensive taxonomy for categorising attention mechanisms for language, text, and speech has been proposed by Chaudhari et al.~\cite{chaudhari2019attentive}. The authors relate most of the concepts to the encoder-decoder model. Hence, for better understanding, we need to define the following: \textit{input sequences}, the input vectors of the model (see Figure~\ref{fig:att-mech-nlp}'s ``Input Sequence''); \textit{output sequences}, the output vectors of the model (see Figure~\ref{fig:att-mech-nlp}'s ``Output Sequence''); \textit{candidate states}, the hidden states of the encoder (see Figure~\ref{fig:att-mech-nlp}'s ``Encoder Vector''); \textit{query states}, the hidden states of the decoder (see Figure~\ref{fig:att-mech-nlp}'s ``Decoder Vector'').

Following this rationale, Chaudhari et al.~\cite{chaudhari2019attentive} organise the different types of attention mechanisms into different (non-mutually exclusive) categories:

\begin{itemize}

    \item \textbf{Number of abstraction levels:} This category takes into account the number of representation/feature levels where the model will learn the attention weights. At this level, we may consider the following types of attention: \textit{single-level attention}, when the attention weights are computed only for the original input sequence; and \textit{multi-level attention}, when we apply the attention mechanism on multiple levels of abstraction of the input sequence, usually in a sequential manner.

    \item \textbf{Number of positions:} This category takes into account the number of positions of the input sequence where the attention weights are learned. At this level, we may consider the following types of attention: \textit{soft attention}, which consists of a weighted average of all the hidden states of the input sequence (i.e., the same as candidate states) to build the context vector (see Figure~\ref{fig:att-mech-nlp}'s ``Context Vector''); \textit{global attention}, which is the term used in the machine translation field to describe soft attention; \textit{hard attention}, which consists of a stochastic sampling of the hidden states to build the context vector; \textit{local attention}, also part of the machine translation field, consists of the detection of an attention point and on the use of a window around that point to compute local attention weights.

    \item \textbf{Number of representations:} This category considers the number of different feature representations of the input sequence used in the learning task. At this level, we can consider the following types of attention: \textit{single-representational attention}, which is the most common and consists in learning the attention weights using only a single feature representation; \textit{multi-representational attention}, which consists in learning the attention weights using multiple representations (of the same input sequence) and creating a context vector that is the result of a weighted combination of these multiple representations and their attention weights; \textit{multi-dimensional attention}, which consists in computing the attention weights along several dimensions of the representation of the input sequence, which will result in the computation of the relevance of each dimension to the learning task.

    \item \textbf{Number of sequences:} This category takes into account the number of input and output sequences. At this level, we may consider the following types of attention: \textit{distinctive attention}, when the candidate and query states belong to two different input and output sequences respectively; \textit{co-attention}, when we consider multiple input sequences at the same time and the main goal is to jointly learn their attention weights; \textit{self-attention}, when the query and candidate states belong to the same input sequence.

\end{itemize}

An analysis of the literature shows that, until very recently, most of the works in attention mechanisms for language, text, and speech derived from the framework proposed by Bahdanau et al.~\cite{bahdanau2014neural}. In~\cite{luong2015effective}, the authors trained a long short-term memory network (LSTM) on the task of neural machine translation (i.e., English to German) and presented an explicit comparative study of local and global attention mechanisms. In~\cite{chorowski2015attention}, the authors considered speech recognition as a sequence generation task (i.e., speech to transcription) and developed an attention mechanism with convolutional features to refine the quality of the output sequences. On the other hand, oppositely to this generalized use of RNN-based architectures, in~\cite{ping2017deep}, the authors proposed a fully-convolutional network (FCN) with an attention block that uses the dot-product operation trained to generate text from speech. Lately, the successful introduction of the~\textit{Transformer} architecture~\cite{vaswani2017attention}, based solely on attention mechanisms, started a new paradigm for their study. Recent studies invest their efforts in the computational or design improvements to the baseline Transformer architecture~\cite{sukhbaatar2019augmenting,sukhbaatar2019adaptive,fan2020addressing}, or in the proposal of novel Transformer-based architectures~\cite{mehta2020delight,wang2020linformer,beltagy2020longformer,kitaev2020reformer,tay2021synthesizer}. On the other hand, opposite to the trend on the increase of model complexity, Merity~\cite{merity2019single} revisited the fundamentals of LSTMs and proposed a~\textit{single-head attention} that could represent an alternative to the use of Transformer-based architectures.

\subsubsection*{Attention mechanisms for computer vision}
Attention mechanisms for computer vision got their inspiration from the human visual system, which can detect whether certain features of an image are relevant or not~\cite{xu2015show}. In contrast to the literature on attention mechanisms for language, text, and speech, there are already a number of interesting review articles on this topic~\cite{yang2020overview,xu2022transformers,han2020survey,khan2021transformers,guo2021attention}. However, only the work of Guo et al.~\cite{guo2021attention} has presented and discussed a complete taxonomy that organizes the different types of attention mechanisms for computer vision into different categories:
\begin{itemize}
    
    \item \textbf{Channel attention:} This category follows the assumption that, in deep CNNs, different channels in different feature maps usually represent different objects~\cite{chen2017sca}. Therefore, the job of channel attention is to calibrate the weight of each channel adaptively. This mechanism acts as an object selector, thus determining \textit{what to pay attention to}.
    
    \item \textbf{Spatial attention:} This category is analogous to channel attention. Here, the job of the attention mechanism is to calibrate the weight of each region of the image adaptively. This mechanism acts as an adaptive spatial region selection process, thus determining \textit{where to pay attention}.
    
    \item \textbf{Temporal attention:} This category considers that data has a temporal dimension, therefore, in the case of computer vision tasks, this type of attention mechanism is often used for video processing. The job of temporal attention is to calibrate the weight of each time frame adaptively. This mechanism acts as a dynamic time selection process, thus determining \textit{when to pay attention}. 
    
    \item \textbf{Branch attention:} This category considers multi-branched deep learning architectures. The job of branch attention is to calibrate the weight of each branch adaptively. This mechanism acts as a dynamic branch selection process, thus determining \textit{which to pay attention to}.
    
    \item \textbf{Channel and spatial attention:} This category combines the advantages of both channel and spatial attention. This mechanism acts as a dynamic spatial region and object selection process, thus determining \textit{what and where to pay attention}.
    
    \item \textbf{Spatial and temporal attention:} This category combines the advantages of both spatial and temporal attention. This mechanism acts as a dynamic spatial region and time-frame selection process, thus determining \textit{where and when to pay attention}.

\end{itemize}

The application of attention mechanisms in computer vision is almost contemporaneous with the first proposals for attention mechanisms in the language, text, and speech domains. Below we present several representative examples from each category.

With respect to \textit{channel attention}, Hu et al.~\cite{hu2018squeeze} proposed the~\textit{squeeze-and-excitation} block. This module adaptively recalibrates channel-wise feature responses by explicitly modelling dependencies between channels. This strengthens the representational power of CNNs by improving the quality of spatial encodings  across the feature hierarchy. These properties have been explored as fundamental components of attention mechanisms in~\cite{cao2019gcnet,quader2020weight}. Still on the scope of channel attention, a different research line, related to the \textit{self-attention} mechanism, was further explored in~\cite{bello2019attention}. In this work, the authors proposed \textit{attention augmentation} mechanisms combining both convolutions and self-attention by concatenating the convolutional feature maps with a set of feature maps generated by self-attention. This approach showed improvements in image classification and object detection while maintaining the number of parameters. In~\cite{wang2017multi}, the authors addressed the topic of one-shot learning and proposed a neural network architecture that uses the semantic representation (i.e., the embedding) of the label to obtain attention maps that are used to generate image features. They also developed a \textit{multiple-attention} scheme to extract meaningful information from the input images and used an auxiliary training set to learn the attention weights.

Concerning \textit{spatial attention}, Mnih et al.~\cite{mnih2014recurrent} designed a novel RNN capable of extracting information from an image or video by adaptively selecting a sequence of regions or locations and only processing these regions at a high resolution. Hence, this method can contribute to a high degree of translation invariance and the ability to process variable input sizes. Dai et al.~\cite{dai2017deformable} developed the \textit{deformable convolution}, which adds flexibility to the regular (spatial) sampling grid of the standard convolution. The intuition behind this approach is that CNNs with these modules extract better features related to the objects of interest in the images (with particular focus on non-rigid objects), thus increasing their robustness to affine transformations. Wang et al.~\cite{wang2018non} revisited the concept of non-local attention and designed the non-local networks, which compute the response at a given spatial location as a weighted sum of the features at all positions in the features. Hu et al.~\cite{hu2018gather} addressed the issue of \textit{context exploitation} in CNNs and proposed a framework consisting of a pair of operators: the \textit{gather} operator, which aggregates contextual information across
large neighbourhoods of each feature map, and the \textit{excite} operator (inspired by~\cite{hu2018squeeze}), which redistributes the pooled information to local features.

Regarding \textit{temporal attention}, Xu et al.~\cite{xu2017jointly} have proposed the \textit{attentive spatial-temporal pooling network} architecture, which operates on video sequences by learning a similarity metric over the features extracted from recurrent and convolutional layers, and computing the attention weights on spatial (i.e., regions in each frame) and temporal (i.e., frames over sequences) levels. Similarly, Chen et al.~\cite{chen2018video} developed an attention mechanism to optimize the similarity learning among short-video snippets, and Zhang et al.~\cite{zhang2019scan} proposed the \textit{self-and-collaborative} attention network, which also learns a similarity metric between feature representations of video-pairs.

Regarding \textit{branch attention}, Srivastava et al.~\cite{srivastava2015training} proposed the concept of \textit{highway networks} which consists of deep neural networks with an LSTM-inspired attention gate that allows for computation paths along which information can flow across many layers (i.e., \textit{information highways}). This work can be considered the precursor of this concept of branch attention. Later, Li et al.~\cite{li2019selective} designed \textit{selective kernel networks}, which contain multiple CNN branches with different kernel sizes that are fused using an attention mechanism, guided by the information in these branches. Zhang et al.~\cite{zhang2020resnest} presented a deep neural architecture that contains several replicated branches (i.e., the \textit{cardinal} branches) that contain more replicated branches (i.e., the \textit{split} branches). In this work, the information fusion is performed with an attention mechanism inspired by~\cite{hu2018squeeze}. Chen et al.~\cite{chen2020dynamic} proposed the \textit{dynamic convolution}, which employs multiple parallel convolution kernels dynamically and aggregates this information through a non-linear attention mechanism.

With respect to \textit{channel and spatial attention}, Woo et al.~\cite{woo2018cbam} have designed a convolutional block attention module that contains a channel attention module followed by a spatial attention module. According to the authors, this block can be easily integrated into any architecture and enables the regression of finer features for object detection. Following this research line, Zhao et al.~\cite{zhao2018psanet} have proposed the~\textit{point-wise spatial attention network} for the task of image segmentation. This network processes contextual information from all positions in the feature map and uses a \textit{self-adaptive} attention mechanism to connect all of them. Using generative adversarial networks (GANs), Zhang et al.~\cite{zhang2019self} explored a \textit{self-attention} mechanism to enable both the generator and the discriminator to efficiently model relationships between widely separated spatial regions.

In terms of \textit{spatial and temporal attention}, Du et al.~\cite{du2017recurrent} introduced the \textit{recurrent spatial-temporal attention network} in which they used an end-to-end LSTM-based network with a spatio-temporal module to process video data, and developed an attention-based mechanism to fuse this information. Song et al.~\cite{song2017end} used RNNs and LSTMs together to process video data and added spatial and temporal attention modules, whose information is later fused. Gao et al.~\cite{gao2019hierarchical} addressed the problem of \textit{visual captioning} with hierarchical LSTM models using spatio-temporal attention to select specific regions or frames to predict the associated words and with adaptive attention to model the importance of visual information or the language contextual information. Yan et al.~\cite{yan2019stat} have also addressed the problem of visual captioning with an architecture based on a CNN-encoder and an LSTM-decoder with spatio-temporal attention to extract spatial and temporal features in a video and guide the decoder to automatically select the most significant regions in the most relevant temporal segments for word prediction.

In addition, several works focus on multi-modal data, thus mixing both the computer vision and the language, text, and speech domains. In~\cite{xu2015show}, the authors revisited the concepts of \textit{hard} and \textit{soft} attention. They implemented an architecture consisting of a convolutional block to extract image features and an LSTM network with attention to generate the image description (i.e., automatic image captioning). 

Although CNN-based architectures had achieved high performance, the transition to Transformer-based architectures occurred almost naturally. Wu et al.~\cite{wu2020visual} discussed three open challenges that motivated the natural implementation of Transformer-based architectures for computer vision tasks: convolutions uniformly process all image patches regardless of their importance, which can lead to spatial inefficiency (e.g., image classification models should prioritise foreground over background); not all images have all concepts, therefore, applying high-level filters to all images could be computationally inefficient (e.g., features related to an image of a person may not be present in an image of a flower); and, the convolution operation is not good at establishing relationships between spatially-distant concepts, which is an essential property in computer vision. For this reason, Dosovitskiy et al.~\cite{dosovitskiy2020image} have proposed the \textit{Vision Transformer} architecture. Analogously to the original Transformer, to work with a Vision Transformer, one needs to split an image into patches and provide the sequence of linear embeddings of these patches as the input. Contemporaneously, Wu et al.~\cite{wu2020visual} proposed a similar architecture that aims to encode semantic concepts in a smaller number of \textit{tokens} and establish a relationship between spatially-distant concepts through an attention mechanism in this token space. Several modifications of the baseline architecture and their application to different tasks have already been reported~\cite{carion2020end, d2021convit,zhou2021deepvit, touvron2021training}.

\section*{Literature Review}\label{sec:sota}
This section provides a comprehensive review of the integration of attention mechanisms into algorithms for medical applications. For the sake of clarity, we structure this section into different types of tasks that can integrate different workflows of the medical domain: medical image classification, medical image segmentation, medical report understanding, and other tasks (medical image detection, medical image reconstruction, medical image retrieval, medical signal processing, and physiological and pharmaceutical research). The work presented in this section results from the analysis of existing surveys (referenced in the Introduction) and a search in Google Scholar\footnote{\url{https://scholar.google.com/}} using the terms ``attention mechanism'', ``deep learning'', and ``medical''.

\subsection*{Medical Image Classification}
Below, we group the works of the most common approaches in recent works.

\subsubsection*{Breast Lesion Classification}
In a multi-database and multi-class classification (i.e., benign, malignant, and normal) approach for breast lesions in ultrasound images, Gheflati and Rivaz~\cite{gheflati2021vision} studied the impact of using data augmentation, and transfer learning on Transformer- and CNN-based architectures, and reported similar, and sometimes improved, results. In a multiple instance learning approach (MIL), Ilse et al.~\cite{ilse2018attention} proposed a gated attention mechanism in deep neural networks for the classification of histopathological images (the authors also investigated the application of this methodology to colon cancer).

\subsubsection*{COVID-19 Classification}
In the task of Coronavirus disease (COVID-19) classification, Han et al.~\cite{han2020accurate} proposed a MIL algorithm for the automated screening of COVID-19 from volumetric chest computed tomography (CT) scans that employs an attention mechanism in the pooling operation to select the instances that may be meaningful for the final classification. Perera et al.~\cite{perera2021pocformer} proposed a lightweight Vision Transformer that uses ultrasound images to discriminate between COVID-19, bacterial pneumonia and healthy specimens. Jiang and Lin~\cite{jiang2021covid} proposed an architecture that combines two Transformer-based models to classify chest X-ray images as COVID-19, pneumonia, and healthy specimens. Liu and Yin~\cite{liu2021automatic} used X-ray image data and applied a Transformer architecture with transfer learning to train a model capable of discriminating between normal and COVID-19 cases. Sangjoon et al.~\cite{park2021federated} explored the potential of Vision Transformer architectures in a federated learning setting for the classification of COVID-19 in chest X-ray images and reported that these architectures are suitable for collaborative learning in medical imaging. Hsu et al.~\cite{hsu2021visual} approached chest CT scan data with an architecture that uses Transformer-based blocks to learn the context of the CT slices, the information at the pixel level, and the spatial-context features with the aid of a deep Wilcoxon signed-rank test to determine the importance of each slice. Zhang and Wen~\cite{zhang2021transformer} proposed a framework for 3D CT scans that consists of a U-Net~\cite{ronneberger2015u} module to segment the lungs and a Transformer-based module to extract features and perform classification (i.e., COVID-19 diagnosis). Mondal et al.~\cite{mondal2021xvitcos} proposed a Vision Transformer with a multi-stage transfer learning strategy that uses CT and X-ray image modalities to discriminate between COVID-19, pneumonia, and healthy specimens. Gao et al.~\cite{gao2021covid} tested a Vision Transformer on 2D and 3D medical image data from CT scans and reported improved results against common CNN architectures for COVID-19 classification (i.e., COVID-19 against non-COVID-19). Similarly, Shome et al.~\cite{shome2021covid} also employed a Vision Transformer on 2D chest CT X-ray images and tested its predictive performance in both binary (i.e., COVID-19 against healthy specimens) and multi-class (i.e., COVID-19 against bacterial pneumonia against healthy specimens) settings, reporting improved results against common CNN architectures. Ambita et al.~\cite{ambita2021covit} used a Vision Transformer to detect COVID-19 in CT scan images and employed a data augmentation strategy that relied on the generation of synthetic images with a self-attention-based GAN. Park et al.~\cite{park2021vision} explored the use of a Vision Transformer architecture for COVID-19 classification that uses low-level features generated using a backbone network. The authors also performed several experiments using chest X-ray databases from different institutions.

\subsubsection*{Whole Slide Image Classification}
Lu et al.~\cite{lu2021smile} addressed the problem of glioma subtype classification (i.e., a multi-class task) in whole slide image (WSI) data using a contrastive training framework that employs a CNN backbone to learn relevant patch-level feature representations, and a sparse-attention block to aggregate the features of these multiple patches (i.e., multiple instances). Chen et al.~\cite{chen2021gashis} employed a Transformer-CNN-based architecture to classify gastric histopathology WSIs in a binary setting (i.e., normal against abnormal). Jiang et al.~\cite{jiang2021method} addressed the diagnosis of acute lymphocytic leukemia through the classification of leukemic B-lymphoblast cells (i.e., cancer cells) and B-lymphoid precursors (i.e., normal cells) with a Transformer-CNN ensemble, and a data enhancement method that tackles the problem of class imbalance. Zheng et al.~\cite{zheng2021deep} proposed a novel graph-based Vision Transformer architecture to classify lung WSIs (i.e., adenocarcinoma, squamous cell carcinoma, and normal histology).

\subsubsection*{Retinal Disease Classification}
Bodapati et al.~\cite{bodapati2021composite} proposed an architecture for the automatic diagnosis of diabetic retinopathy that uses multiple pre-trained CNNs with spatial pooling to extract features from color fundus retinal images and integrates gated attention blocks to guide the model to focus more on lesion portions of the retinal images while paying less attention to the non-lesion regions. Yu et al.~\cite{yu2021mil} addressed the task of retinal disease classiﬁcation using fundus image data with a Vision Transformer trained under a MIL setting. Similarly, in diabetic retinopathy classification, Sun et al.~\cite{sun2021lesion} proposed a lesion-aware Transformer architecture that jointly learns to detect the presence of diabetic retinopathy and the location of lesion discovery, using an encoder-decoder structure. Several authors have addressed the task of diabetic retinopathy recognition in a multi-class setting (i.e., no diabetic retinopathy, mild non-proliferative diabetic retinopathy, moderate diabetic retinopathy, severe non-proliferative diabetic retinopathy, and proliferative diabetic retinopathy) with a Vision Transformer architecture \cite{wu2021vision,aldahoul2021encoding}. Yang et al.~\cite{yang2021fundus} used a hybrid CNN-Transformer architecture to tackle ophthalmic image data in a multi-class setting (i.e., normal, diabetes, glaucoma, cataract, age-related macular degeneration, hypertension, myopia, and other abnormalities) using different data pre-processing strategies.

\subsubsection*{Miscellaneous Classification Tasks}
In lung cancer classification in screening CT, Al-Shabi et al.~\cite{al2020procan} have extended the study of non-local attention and proposed an architecture that is initially trained on simple examples and gradually grows to increase its ability to handle the task at hand, as the classification task becomes more difficult. Moranguinho et al.~\cite{moranguinho2021attention} applied this methodology for the classification of lung biopsy histopathological images with respect to cancer classification and used post-model interpretability algorithms to assess the regions of interest for the predictions of their trained deep neural network. He et al.~\cite{he2021automatic} proposed a deep learning architecture for depression recognition that combines a backbone CNN to extract features, a local attention module that focuses on parts of the input images, a global attention module to learn global patterns from the entire input images, and a weighted spatial pyramid pooling layer to learn the depression patterns after the feature aggregation operation. Dai et al.~\cite{dai2021transmed} explored the task of multi-modal and multi-class classification in head, neck, and knee magnetic resonance imaging (MRI) data using a hybrid CNN-Transformer model, where the CNN module is used as a low-level feature extractor. Datta et al.~\cite{datta2021soft} performed a comparative study of the effect of~\textit{soft-attention} combined with different backbone architectures in skin cancer classification and reported several advantages against conventional methodologies. Barhoumi and Ghulam~\cite{barhoumi2021scopeformer} proposed a model that aggregates several feature maps extracted using multiple Xception CNNs~\cite{chollet2017xception} and uses these features to train a Vision Transformer for the intracranial hemorrhage classification problem, using CT images. Liang and Gu~\cite{liang2021computer} added an attention mechanism to a backbone network based on the ResNet~\cite{he2016deep} to aid the diagnosis of Alzheimer’s disease in brain MRI data.

\subsection*{Medical Image Segmentation}
Using well-known data sets related to different use cases and working on top of~\cite{hu2018squeeze}, Roy et al.~\cite{roy2018recalibrating} verified that the addition of a spatial-channel squeeze and excitation block works as an attention mechanism in FCNs and improves the quality of the segmentation maps. Interestingly, in most recent papers, the authors approach the problem of medical image segmentation using encoder-decoder structures that generally comprise a CNN and a Transformer to extract volumetric spatial feature maps, perform global feature modeling and predict refined segmentation maps \cite{wang2021transbts,jia2021bitr,peiris2021volumetric,hatamizadeh2022swin}, although there are already some works that replace the CNN-based modules at the encoder or decoder levels and integrate other attention mechanisms to extract features and model long-range dependencies~\cite{li2021more}. More recent methodologies on medical image segmentation are taking advantage of a hybrid use of the Vision Transformer and the U-Net with improved results regarding the quality of segmentation maps~\cite{chen2021transattunet,xu2021levit}. Wu et al.~\cite{wu2022fat} proposed a \textit{feature adaptive} model that comprises a dual encoder with CNN and Transformer branches to simultaneously capture both local features and global context information. Below, we group the works of the most common approaches in recent works.

\subsubsection*{Breast Lesion Segmentation}
Using breast ultrasound images, Vakanski et al.~\cite{vakanski2020attention} integrated an attention block into the well-known architecture U-Net to learn semantic representations that prioritise spatial regions for the task of breast tumour segmentation. In breast tumour segmentation, using ultrasound imaging data, Zhu et al.~\cite{zhu2021region} developed a Transformer-based architecture that incorporates prior information of the region of tumours to obtain accurate segmentation. Following the most recent approaches of Transformer-based architectures for medical image analysis, Liu et al.~\cite{liu20213d} proposed a U-Net-like model that integrates an attention module to focus on the tumour region and combines CNN and Transformer blocks to extract and process features that may lead to improved segmentation maps. 

\subsubsection*{Computational Pathology}
Using microscopic images, Prangemeier et al.~\cite{prangemeier2020attention} proposed an architecture based on a CNN-encoder and Transformer-decoder for the \textit{classification} and \textit{instance segmentation} of yeast cells. Using microscopy images of corneal endothelial cells, Zhang et al.~\cite{zhang2021multi} revisited the problem of cell segmentation with the proposal of the multi-branch hybrid Transformer network, which follows the main structure of its predecessors (i.e., it contains both CNN and Transformer modules) to extract and process local and spatial features that may lead to refined segmentation maps. Shao et al.~\cite{shao2021transmil} addressed MIL using a Transformer-based architecture that explores morphological and spatial information and considers the correlation between instances. The authors tested their framework in different computational pathology problems. Using high-resolution images, Nguyen et al.~\cite{nguyen2021evaluating} performed a comparative analysis using architectures based on CNN and Transformer modules and reported improved results with the Transformer-based approaches. In a similar approach, using hyperspectral image data, Yun et al.~\cite{yun2021spectr} proposed an encoder-decoder architecture with CNN and Transformer blocks to extract and model spatial and spectral features, achieving competitive results against other methodologies. 

\subsubsection*{Cardiac Segmentation}
Li et al.~\cite{li2020generalisable} addressed the task of cardiac image segmentation using MRI data from different domains with a generative adversarial model with an attention loss, proposed to translate the images from existing source domains, and a stack of data augmentation techniques to simulate real-world transformation to boost the segmentation performance for unseen domains. Concurrently, Kong and Shadden~\cite{kong2020generalizable} tackled the same problem with a GAN with a cycle consistency loss (i.e., CycleGAN~\cite{zhu2017unpaired}) to generate images in different styles, exchanging the low-frequency features of images from different domains, and using an architecture based on an attention-gated U-Net that should learn to focus on cardiac structures of varying shapes and sizes while suppressing irrelevant regions. These strategies have been followed by the community in other contexts as well~\cite{campello2021multi}. In the task of cardiac segmentation, using ultrasound images (i.e., echocardiography), Deng et al.~\cite{deng2021transbridge} developed an architecture that combines a CNN-based encoder-decoder to extract multi-level features, a Transformer structure to fuse these features, and a patch embedding layer created to reduce the number of parameters of the embedding layer and the size of the token sequence. Chen et al.~\cite{chen2021transunet} proposed a U-Net-like architecture that studied the potential of several settings (e.g., Transformer as encoder, hybrid CNN-Transformer as encoder) that reported increased performances for cardiac segmentation (the authors also reported increased performances on multi-organ segmentation). Huang et al.~\cite{huang2021missformer} also proposed a Transformer-based model that tackles several tasks: the alignment of features, the enhancement of the long-range dependencies and local context, and the modelling of the long-range dependencies and local context of multi-scale features. They experimented with this architecture on cardiac segmentation databases (also on multi-organ segmentation).

\subsubsection*{Multi-task Segmentation}
Yao et al.~\cite{yao2020claw} proposed a U-Net-based architecture with deep feature concatenation and an attention mechanism branched into several attention gates for the task of scleral blood vessel segmentation. Following this methodology, Chang et al.~\cite{chang2021transclaw} added a Transformer module to the encoder block of the baseline model and reported improved results on multi-organ segmentation tasks. Using CT images, Xie et al.~\cite{xie2021cotr} proposed a framework that connects a CNN and a Transformer to extract both feature representations and model the long-range dependency on these feature maps. This model also employs a self-attention mechanism that focuses on specific positions of the image, thus reducing the computational complexity. Similarly, for kidney segmentation using CT images, Shen et al.~\cite{shen2021automated} also proposed an encoder-decoder architecture that leverages the interaction of CNN and Transformer modules to learn multi-scale features to achieve improved segmentation results. Following this work, Zhang et al.~\cite{zhang2021transfuse} developed an architecture for multi-organ segmentation that runs a CNN-based encoder and Transformer-based segmentation network in parallel and fuses the features from these two branches to jointly make predictions. Tang et al.~\cite{tang2021self} proposed a U-Net shaped architecture with a Transformer module for multi-organ segmentation with 3D images and employs a specific pre-training strategy with contrastive learning, masked volume inpainting, and 3D rotation prediction. Using data from different modalities and organs, Karimi et al.~\cite{karimi2021convolution} proposed an architecture that solely uses Transformer modules to obtain the segmentation maps, thus showing that it is possible to develop models that can achieve competitive results without relying on the convolution operation. Cao et al.~\cite{cao2021swin} also proposed a U-Net architecture composed solely of Transformer-based modules for multi-organ segmentation that uses skip-connections for local-global semantic feature learning. Using images from different modalities (i.e., MRI and CT) and organs (i.e., brain cortical plate, hippocampus, pancreas), Karimi et al.~\cite{karimi2021convolution} proposed an architecture based entirely on Transformers that uses self-attention between neighboring image patches, without requiring any convolution operations. On colonoscopy images, Sang et al.~\cite{sang2021ag} proposed an architecture that takes advantage of a ResNeSt~\textit{backbone}~\cite{zhang2020resnest} connected to the coupled U-Nets~\cite{tang2018cu} architecture and several attention gates to combine multi-level features and yield accurate polyp segmentation. In a multi-task segmentation approach, Zhang et al.~\cite{zhang2021pyramid} developed a method that seeks to integrate multi-scale attention and CNN feature extraction using a pyramidal network architecture, by working on multi-resolution images. Lin et al.~\cite{lin2021ds} proposed an encoder-decoder architecture comprised of hierarchical Transformer-based modules to model the long-range dependencies and the multi-scale context connections during the process of down-sampling and up-sampling. Li et al.~\cite{li2021medical} proposed a different framework based on Transformers, that uses a squeeze-and-excitation attention module to both regularize the self-attention mechanism of Transformers and learn diversified representations. Ranjbarzadeh et al.~\cite{ranjbarzadeh2021brain} addressed brain tumor segmentation in multi-modal MRI data using a cascade CNN that processes both local and global features in two different routes, combined with a distance-wise attention mechanism that considers the effect of the location of the center of the tumor and the brain. Petit et al.~\cite{petit2021u} designed a U-shaped architecture for image segmentation with self-attention and cross-attention from Transformers (i.e., U-Transformer). They performed experiments on two abdominal CT-image databases and obtained superior performance when compared to U-Net-based models. Regarding skin lesion segmentation, Wang et al.~\cite{wang2021boundary} proposed a \textit{boundary-aware} Transformer that aims to take advantage of \textit{boundary-wise priors} through a boundary-wise attention gate that highlights the ambiguous boundaries to generate attention maps to guide the training. For tooth root segmentation, Li et al.~\cite{li2021gt} proposed a Transformer-based architecture with the structure of the U-Net backbone and optimised the training of this model with a Fourier Descriptor loss function that takes advantage of prior knowledge related to shape. Li et al.~\cite{li2021agmb} explored the same task through an anatomy-guided multi-branch Transformer with multi-head attention that employs a polynomial curve fitting segmentation strategy (based on keypoint detection) to extract anatomy features. A similar approach was employed in~\cite{guo2021sa} for retinal vessel segmentation. This methodology was extended in~\cite{jin2020ra}, which employs a 3D U-Net-based architecture with \textit{residual} blocks and an attention module to extract volumes of interest in liver and brain CT imaging data and perform tumor segmentation. Sobirov et al.~\cite{sobirov2022automatic} applied the Transformer-based model proposed by Hatamizadeh et al.~\cite{hatamizadeh2022swin} to the tasks of head and neck tumour segmentation using multi-modal data (i.e., CT and PET images) and compared their results with traditional CNN-based approaches. Yan et al.~\cite{yan2022after} also employed a U-Net-based structure for the task of multi-organ segmentation in 3D medical image data, however, with slightly different changes: in their approach, they used a CNN-encoder and CNN-decoder with a Transformer model in between to fuse contextual information in the neighboring image slices.

\subsection*{Medical Report Understanding}

\subsubsection*{Medical Report Generation}
Using two medical databases containing radiological and pathological images, Jing et al.~\cite{jing2017automatic} proposed a methodology consisting of a multi-task deep learning model that jointly performs the prediction of tags and the generation of paragraphs, a~\textit{co-attention} mechanism for locating regions containing abnormalities and generating narrations for them, and a hierarchical LSTM model for generating long paragraphs. One of the first mentions of  using Transformer architectures to generate medical image reports is described in~\cite{xiong2019reinforced}. In this work, the authors used X-ray data and proposed a hierarchical framework trained with reinforcement learning. This framework consists of an encoder that extracts visual features through a bottom-up attention mechanism aimed at identifying regions of interest and extracting top-down visual features, and a Transformer-based non-recurrent captioning decoder aimed at generating a coherent paragraph of medical image reports. Lovelace and Mortazavi~\cite{lovelace2020learning} developed a chest X-ray report generation algorithm that consists of two training stages: the training of the report generation model (based on a Transformer architecture) using a standard language generation objective, and, after, the fine-tuning of this model on clinical observations extracted from reports sampled from the same model to regularise the degree of coherence between the observations from the generated and ground truth reports. In a similar task, Srinivasan et al.~\cite{srinivasan2020hierarchical} proposed a deep framework that uses a CNN to extract features of the images and generate tag embeddings for each image, and uses Transformers to encode the image and tag features with self-attention to get a finer representation. Alternatively, Miura et al.~\cite{miura2020improving} applied a Transformer-based architecture trained with reinforcement learning which comprises two different rewards, namely, one that encourages the system to generate radiology domain entities consistent with the reference, and one that uses natural language inference to encourage these entities to be described in inferentially consistent ways. On the other hand, Chen et al.~\cite{chen2020generating} proposed to generate radiology reports with a memory-driven Transformer, where a relational memory is designed to record key information of the generation process and a memory-driven conditional layer normalisation is applied to incorporate the memory into the decoder of Transformer. Also in automatic chest X-ray report generation, Liu et al.~\cite{liu2021contrastive} created a~\textit{contrastive-attention} methodology that compares a given input image with normal images to distill contrastive information that better represents the visual features of abnormal regions, thus providing more accurate descriptions for an interpretable diagnosis and improving the quality of the generated reports. Najdenkoska et al.~\cite{najdenkoska2021variational} tackled the same problem with~\textit{variational topic inference} that consists of modelling a set of topics as latent variables to guide sentence generation by aligning image and language modalities in a latent space. Each topic contributes to the generation of a sentence in the report. The entire pipeline is refined with a visual attention module that enables the model to attend to different locations in the image and generate more informative descriptions. Alfarghaly et al.~\cite{alfarghaly2021automated} proposed a Transformer-based pipeline that involves the use of a fine-tuned CheXNet model~\cite{rajpurkar2017chexnet} to predict specific tags from the image, the computation of weighted semantic features from the predicted tag’s pre-trained embeddings, and the conditioning of a pre-trained distilled GPT2~\cite{radford2019language} model on the visual and semantic features to generate the full medical reports. Amjoud et al.~\cite{amjoud2021automatic} proposed an encoder-decoder framework that combines the benefits of transfer learning for feature extraction and the Transformer architecture for medical report generation. Tipirneni et al.~\cite{tipirneni2021self} proposed a Transformer-based architecture to address multivariate clinical time series with missing values. A different approach was discussed in~\cite{wang2021confidence}, where the authors propose an encoder-decoder architecture that comprises both CNN and Transformer modules and aims to explicitly quantify the inherent visual-textual uncertainties existing in the multi-modal task of radiology report generation both at the report and sentence levels, and explore a novel method to measure the semantic similarity of radiology reports, which seems to better capture the characteristics of diagnosis information. A different research line focuses on imitating the working patterns of radiologists, which typically consist of examining the abnormal regions and assigning the disease topic tags to these regions and then relying on the years of prior medical knowledge and prior working experience to write reports~\cite{liu2021exploring}. Hence, in~\cite{liu2021exploring}, the authors proposed a \textit{posterior-and-prior knowledge exploring-and-distilling} approach that consists of three modules: the \textit{posterior knowledge explorer}, which aims to provide explicit abnormal visual regions to alleviate visual data bias; the \textit{prior knowledge explorer}, which aims to explore the prior knowledge from the prior medical knowledge graph (i.e., medical knowledge) and prior radiology reports (i.e., working experience) to alleviate textual data bias; and the \textit{multi-domain knowledge distiller} that aggregates this processed knowledge and generates the final reports. A very close framework was described by Liu et al.~\cite{liu2021auto}, in which the authors proposed an \textit{unsupervised model knowledge graph auto-encoder} which accepts independent sets of images and reports during training, and consists of three modules: the \textit{pre-constructed knowledge graph}, that works as the shared latent space and aims to bridge the visual and textual domains; the \textit{knowledge-driven encoder} which projects medical images and reports to the corresponding coordinates in that latent space; and the \textit{knowledge-driven decoder} that generates a medical report given a coordinate in that latent space. This modular structure is also employed by Nguyen et al.~\cite{nguyen2021automated}, which added three complementary modules: a CNN-based \textit{classification} module that produces an internal checklist of disease-related topics (i.e., \textit{the enriched disease embedding}); a Transformer-based \textit{generator} that generates the medical reports from the enriched disease embedding and produces a \textit{weighted embedding representation}; and an \textit{interpreter} that uses the weighted embedding representation to ensure consistency concerning disease-related topics. Similarly, You et al.~\cite{you2021aligntransformer} proposed a framework, which includes two different attention-based modules: the \textit{align hierarchical attention} module that first predicts the disease tags from the input image and then learns the multi-grained visual features by hierarchically aligning the visual regions and disease tags; and the \textit{multi-grained Transformer} module that uses the multi-grained features to generate the medical reports. Still, on this line, Hou et al.~\cite{hou2021ratchet} proposed a CNN-RNN-based medical Transformer trained end-to-end, capable of extracting image features and generating text reports that aim to fit seamlessly into clinical workflows. 

\subsubsection*{Miscellaneous Tasks}
Recently, Zhou et al.~\cite{zhou2022generalized} proposed a cross-supervised methodology that acquires free supervision signals from original radiology reports accompanying the radiography images, and employs a Vision Transformer, designed to learn joint representations from multiple views within every patient study. Regarding automatic surgical instruction generation, Zhang et al.~\cite{zhang2021surgical} introduced a Transformer-based encoder-decoder architecture trained with \textit{self-critical} reinforcement learning to generate instructions from surgical images, and reported improvements against the existing baselines over several caption evaluation metrics. Wang et al.~\cite{wang2021self} proposed a Transformer-base training framework designed for learning the representation of both the image and text data, either paired (e.g., images and texts from the same source) or unpaired (e.g., images from one source coupled with texts from another source). They applied their methodology to different chest X-ray databases and performed experiments in three different applications, i.e., classification, retrieval, and image regeneration.

\subsection*{Other Tasks}
This subsection includes medical tasks whose interest in the field is currently increasing: medical image detection, medical image reconstruction, medical image retrieval, medical signal processing, and physiology and pharmaceutical research.

\subsubsection*{Medical Image Detection}
Regarding colorectal polyp detection, Shen et al.~\cite{shen2021cotr} approached several colonoscopy databases and proposed an architecture constituted by a CNN for feature extraction, Transformer encoder layers interleaved with convolutional layers for feature encoding and recalibration, Transformer decoder layers for object
querying, and a feed-forward network for detection prediction. Similarly, Liu et al.~\cite{liu2021transformer} addressed the same task using an encoder-decoder architecture with a ResNet50-based CNN backbone as the encoder, and a Transformer-based module as the decoder and Mathai et al.~\cite{mathai2021lymph} addressed the task of lymph node detection using MRI data employing analogous strategies.

\subsubsection*{Medical Image Reconstruction}
Gong et al.~\cite{gong2021direct} addressed the reconstruction of linear parametric images from dynamic positron emission tomography (PET) and implemented a 3D U-Net~\cite{cciccek20163d} model with an attention layer that focuses on prior anatomical information during training to improve the quality of the reconstruction of dynamic PET images of the human brain. Liang and Gu~\cite{liang2021computer} also explored the reconstruction of brain MRI data as a regularisation strategy to aid the diagnosis of Alzheimer’s disease.

\subsubsection*{Medical Image Retrieval}
Grundmann et al.~\cite{grundmann2021self} and Kim and Ganapathi~\cite{kim2021read} developed new methods to perform an automatic retrieval and analysis of clinical notes. Li et al.~\cite{li2021anatomy} and Ma et al.~\cite{ma2021transformer} developed new models for the automated evaluation of root canal therapy and significant stenosis detection in coronary CT angiography of coronary arteries, respectively.

\subsubsection*{Medical Signal Processing}
Nauta et al.~\cite{nauta2019causal} explored the combination of attention mechanisms and causal discovery techniques. Using simulated functional MRI data containing blood-oxygen-level dependent data sets for 28 different underlying brain networks, this work proposed a deep learning framework based on attention that learns a causal graph structure by discovering causal relationships in observational time series data.

\subsubsection*{Physiology and Pharmaceutical Research}
Lately, deep learning has also revealed itself to be a popular framework for physiology and pharmaceutical research (e.g., drug combination prediction, protein residue-residue contact prediction). Naturally, since this field also involves high-stake decisions, the increase in the predictive performance of deep learning architectures in several problems of this topic~\cite{shrestha2019assessing, zheng2019deep, hou2019protein, senior2020improved} motivated the need to explain the internal mechanisms of these algorithms and improve their explainability. Interestingly, the strategy employed by machine learning practitioners is based on attention mechanisms. Therefore, we consider it relevant to include works on this matter in this survey, even though they may not be considered direct medical applications. Filipavicius et al.~\cite{filipavicius2020pre} studied the use of Transformer-based language models for protein classification using long sequences. The authors also performed several experiments on the compression of the protein sequences and prepared new pre-training and protein-protein binding prediction databases. Chen et al.~\cite{chen2021combination} presented an attention-based (i.e., regional and sequence attention) CNN for protein contact prediction and for the identification of interpretable patterns that contain useful insights into the key fold-determining residues in proteins. Wang et al.~\cite{wang2021deepdds} recently proposed a deep learning framework based on graph neural networks with an attention mechanism to identify drug combinations that can effectively inhibit the viability of specific cancer cells. Khan and Lee~\cite{khan2021gene} used a Transformer-based architecture for the analysis of high-dimensional gene expression data and reported improved results on the task of lung cancer sub-types classification.

\section*{Case Studies}\label{sec:case_studies}
Most of the works of the previous section are often based on predictive performance (i.e., accuracy) rather than showing that these algorithms are moving towards the increase of transparency or interpretability. In a high-stake decision area such as healthcare, the authors must care about framing their proposals to the context of the application. Therefore, we decided to design an experimental set of case studies, involving three different use cases (i.e., breast lesion, chest pathology, and skin lesion), wherein the main goal is to provide insight into the impact of using attention mechanisms in deep learning algorithms for medical image applications. On the one hand, we assess this impact from the perspective of predictive performance (i.e., accuracy). On the other hand, we report post-model attribution maps obtained with different interpretability frameworks to visualise the image regions that contributed most to the final predictions. We acknowledge that the visual assessment of these attribution maps is subject to a high degree of subjectivity. However, we argue that this analysis can serve as an indirect evaluation of the degree of interpretability of these models.

\subsection*{Data}
To ensure that our conclusions are independent of the data, we selected medical image classification databases related to three different use-cases: diabetic retinopathy, chest pathologies and skin lesion classification. Table~\ref{tab:data-info} presents the number of training, validation and test examples for each data set, and the meaning of the labels we used in this classification case study.

\subsubsection*{APTOS2019}
The APTOS2019 data set contains retinography images taken under a variety of imaging conditions (mydriatic vs. non-mydriatic) and from different camera models and ethnicities. The data set was generated by Aravind Eye Hospital in India in cooperation with Asia Pacific Tele-Ophthalmology Society (APTOS). Each retinal image has been graded independently by two ophthalmologists, with any disagreements adjudicated by a third ophthalmologist. A retinopathy severity score was assigned to each image according to the NHS diabetic eye screening guidelines\footnote{\url{https://www.gov.uk/government/collections/diabetic-eye-screening-commission-and-provide}} as R0 (no diabetic retinopathy), R1 (background), R2 (pre-proliferative), R3 (proliferative), M0 (no visible maculopathy), M1 (maculopathy). A curated version of this data set was used in the \textit{APTOS 2019 Blindness Detection} hosted on Kaggle\footnote{\url{https://www.kaggle.com/c/aptos2019-blindness-detection}}. For the binary classification case, we grouped all the non-R0 classes into the ``diabetic retinopathy'' class (see Table~\ref{tab:data-info}).

\subsubsection*{ISIC2020}
The ISIC2020~\cite{rotemberg2021patient} data set contains dermoscopic training images of unique benign and malignant skin lesions from over 2000 patients. Each image is associated with one of these individuals using a unique patient identifier. All malignant diagnoses have been confirmed via histopathology, and benign diagnoses have been confirmed using either expert agreement, longitudinal follow-up, or histopathology. The data set was generated by the International Skin Imaging Collaboration (ISIC) and images were collected by the following sources: Hospital Clínic de Barcelona, Medical University of Vienna, Memorial Sloan Kettering Cancer Center, Melanoma Institute Australia, University of Queensland, and the University of Athens Medical School. The data set was curated for the \textit{SIIM-ISIC Melanoma Classification Challenge} hosted on Kaggle\footnote{\url{https://www.kaggle.com/c/siim-isic-melanoma-classification/overview}}.

\subsubsection*{MIMIC-CXR}\label{sssec:chest}
The MIMIC Chest X-ray (MIMIC-CXR) Database v2.0.0~\cite{johnson2019mimic} is a large publicly available data set of chest radiographs in the Digital Imaging and Communications in Medicine (DICOM) format with free-text radiology reports. The data set contains 227835 imaging studies for 65379 patients of the Beth Israel Deaconess Medical Centre Emergency Department (Boston, MA) between 2011–2016. Each imaging study can contain one or more images, usually a frontal and a lateral view. A total of 377,110 images are available in the data set. Studies are made available with a semi-structured free-text radiology report that describes the radiological findings of the images, written by a practicing radiologist during routine clinical care. The data set is de-identified to satisfy the US Health Insurance Portability and Accountability Act of 1996 (HIPAA) Safe Harbor requirements. Protected health information (PHI) has been removed. For the binary classification case we used a subset of images of the anterior-posterior view related to the presence or absence of pleural efusion (see Table~\ref{tab:data-info}).

\begin{table}[th]
\centering
\caption{Data splits and meaning of labels for each data set.}
\label{tab:data-info}
\begin{tabular}{@{}lccccc@{}}
\cmidrule(l){2-6}
\multirow{2}{*}{\textbf{Data set}} & \multicolumn{3}{c}{\textbf{Splits}}                  & \multicolumn{2}{c}{\textbf{Labels}} \\ \cmidrule(l){2-6} 
                                   & \textbf{Train} & \textbf{Validation} & \textbf{Test} & \textbf{0}       & \textbf{1}       \\
\textbf{APTOS2019} & 2334  & 778  & 550  & Normal & Diabetic Retinopathy \\
\textbf{ISIC2020}  & 19593 & 6568 & 6965 & Benign & Malign \\
\textbf{MIMIC-CXR} & 61203 & 534  & 1072 & Normal & Pleural Efusion \\ \bottomrule
\end{tabular}%
\end{table}

\subsection*{Methodology}
We divided the experimental part of this survey into several phases, described below. We conducted all the experiments using the PyTorch library~\cite{NEURIPS2019_9015} for Python.

\subsubsection*{Backbone Models}
In the first phase of this study, we trained two deep learning backbones (i.e., DenseNet-121~\cite{huang2017densely}, ResNet-50~\cite{he2016deep}) on each data set. DenseNet-121's architecture~\cite{huang2017densely} allows connecting all layers (with matching feature-map sizes) directly with each other, thus improving the flow of information and gradients throughout the network, and facilitating their training. ResNet50's architecture introduced the \textit{deep residual learning framework}, which consists of adding \textit{skip connections} that perform \textit{identity mapping}, and adding their outputs to the outputs of the stacked layers. We refer the reader to the original papers for more details. We used these backbone models in our experiments because they are well known and widely used in the deep learning community. In this work, we refer to these models as \textit{DenseNet-121} and \textit{ResNet-50}.

\subsubsection*{Squeeze-and-Excitation Attention Block}
\begin{figure}[ht]
    \centering
    \includegraphics[scale=0.6]{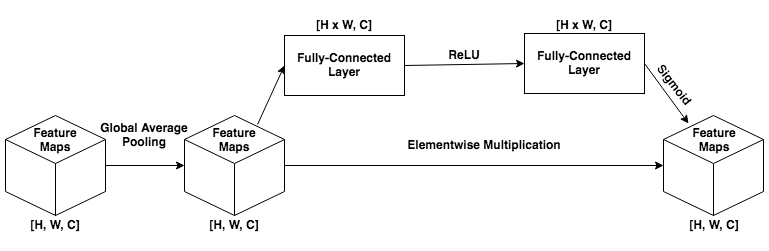}
    \caption{Block diagram of the \textit{Squeeze-and-Excitation} (SE) attention block, according to~\cite{hu2018squeeze}.}
    \label{fig:scheme-selayer}
\end{figure}

In the second phase, we adapted the \textit{Squeeze-and-Excitation} (SE) attention block~\cite{hu2018squeeze} (see Figure~\ref{fig:scheme-selayer}) to each of the backbones and trained these new architectures on the three data sets. According to the authors of the original paper~\cite{hu2018squeeze}, this attention block was designed to adaptively recalibrate channel-wise feature responses by explicitly modelling interdependencies between channels. Since one of the original implementations uses the ResNet-50, we used the author's implementation in our study, which consists of integrating the SE block between the \textit{residual layers} and the activation function (in this case, the \textit{rectified linear unit} (ReLU)). Following this rationale, in the case of DenseNet-121, we integrated the SE block between the \textit{dense} and the \textit{transition} blocks. We decided to use the SE block in our experiments because it was one of the first proposals of channel attention for computer vision, and it is widely used by the community as a comparison reference. Besides, since we are building our case studies in medical image classification using RGB images, our interests also rely on studying the effects of attention on the channel dimension. In this work, we refer to these models as \textit{SEDenseNet-121} and \textit{SEResNet-50}.

\subsubsection*{Convolutional Block Attention Module}
\begin{figure}[ht]
    \centering
    \includegraphics[scale=0.6]{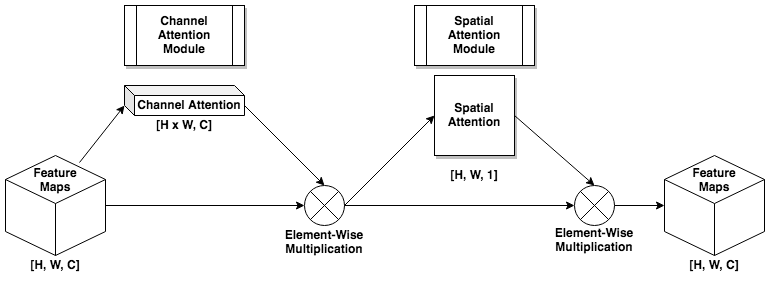}
    \caption{Block diagram of the \textit{Convolutional Block Attention Module} (CBAM), according to~\cite{woo2018cbam}.}
    \label{fig:scheme-cbamlayer}
\end{figure}

\begin{figure}[ht]
    \centering
    \includegraphics[scale=0.6]{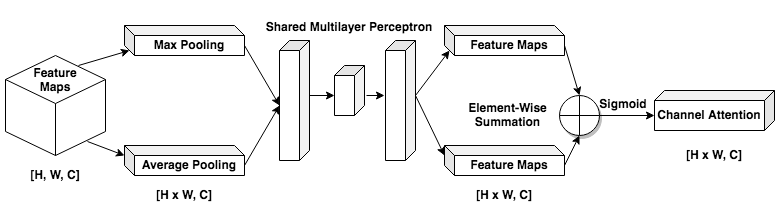}
    \caption{Block diagram of the \textit{Channel Attention Module} of the \textit{Convolutional Block Attention Module} (CBAM), according to~\cite{woo2018cbam}.}
    \label{fig:scheme-camlayer}
\end{figure}

\begin{figure}[ht]
    \centering
    \includegraphics[scale=0.6]{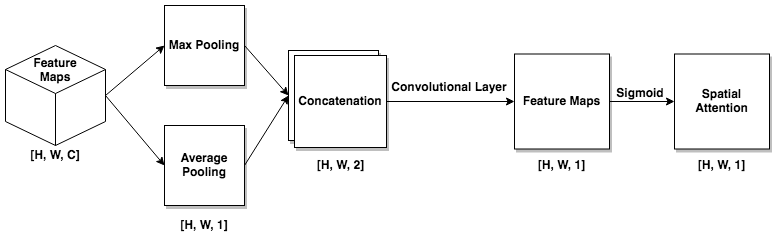}
    \caption{Block diagram of the \textit{Spatial Attention Module} of the \textit{Convolutional Block Attention Module} (CBAM), according to~\cite{woo2018cbam}.}
    \label{fig:scheme-samlayer}
\end{figure}

In the third phase, we adapted the \textit{Convolutional Block Attention Module} (CBAM)~\cite{woo2018cbam} (see Figure~\ref{fig:scheme-cbamlayer}) to each of the backbones and trained these new architectures on the three data sets. According to the authors of the original paper, the CBAM mechanism integrates two specific attention blocks: the \textit{channel attention module}, which aims to produce a channel attention map by exploiting the inter-channel relationship of features and is considered as a feature detector (see Figure~\ref{fig:scheme-camlayer}); and the \textit{spatial attention module}, which aims to generate a spatial attention map by utilizing the inter-spatial relationship of features, thus being complementary to the channel attention (see Figure~\ref{fig:scheme-samlayer}). Like the SE attention block, we integrated the CBAM mechanism at the same topographical location of the architectures of the backbone models. Following the decision on the SE block, we considered it important to study the spatial dimension of images. Therefore, since CBAM integrates both the attention and spatial dimensions in its pipeline, we can perceive this module as an upgrade to the SE block. Besides, it is important to note that this module is also well-known in the community, and, therefore, it is validated as well. In this work, we refer to these models as \textit{CBAMDenseNet-121} and \textit{CBAMResNet-50}.

\subsubsection*{Data-efficient image Transformer}
In the fourth phase, we tested a Transformer-based architecture composed solely of attention mechanisms. The Data-efficient image Transformer (DeiT)~\cite{touvron2021training} is an architecture inspired by the Vision Transformer~\cite{dosovitskiy2020image}, and trained with fewer parameters. In this case, we used the \textit{DeiT-Ti} variation~\cite{touvron2021training}, which has a comparable number of parameters against the chosen CNN backbones. Besides being validated by the community (as in previous examples), this Vision Transformer architecture has a proportional number of parameters (see Table~\ref{tab:model-complexity}) to the backbone models, thus allowing us to mitigate eventual model complexity effects on the predictive performance of this model. In this work, we refer to this model as \textit{DeiT}.

\subsubsection*{Post-model Interpretability}
In the fifth phase, we generated visual explanation maps (i.e., saliency maps) using the \textit{Deep Learning Important FeaTures} (DeepLIFT)~\cite{shrikumar2017learning} and the \textit{Layer-wise Relevance Propagation} (LRP)~\cite{bach2015pixel} \textit{post-hoc} methods. DeepLIFT~\cite{shrikumar2017learning} compares the activation of each neuron to its related \textit{reference activation}, and assigns contribution scores according to the difference. LRP~\cite{bach2015pixel} is a methodology that aims to create visualizations of the contributions of single pixels to predictions. We decided to employ this strategy as a proxy for the degree of interpretability of models, thus allowing our work to be comparable to current research. We used the Captum~\cite{kokhlikyan2020captum} library for Python to generate these \textit{post-hoc} saliency maps for the backbones and their attention-based versions. Regarding the DeiT, we refer the reader to~\cite{chefer2021transformer}, which proposed a framework to generate LRP attributions for Transformer-based architectures.

\subsubsection*{Data Processing}
Regarding the data processing pipeline, all the images were resized to the dimensions of $224\times224$, and a z-normalization was applied to each RGB channel. The \textit{mean} and \textit{standard deviation} parameters used in this operation depend on the initialization of the weights of the model. In the case of DenseNet-121, ResNet-50, SEDenseNet-121, SEResNet-50, CBAMDenseNet-121 and CBAMResNet-50, the weights were initialized from a pre-training on the ImageNet database~\cite{deng2009imagenet} with $\textup{mean} = \left [ 0.485, 0.456, 0.406 \right ]$, and $\textup{std} = \left [ 0.229, 0.224, 0.225 \right ]$. In the case of DeiT, the weights were initialised from a pre-training on the ImageNet database with $\textup{mean} = \left [ 0.500, 0.500, 0.500 \right ]$, and $\textup{std} = \left [ 0.500, 0.500, 0.500 \right ]$.

\subsubsection*{Data Augmentation}
Regarding the data augmentation strategy, we decided to create a pipeline composed of several random affine transformations, i.e., random rotations, random translations, random scaling and random horizontal flip. Table~\ref{tab:dataug} presents the parameters of data augmentation used in our experiments.

\begin{table}[ht]
\centering
\caption{Data augmentation parameters used during the training step of all the phases.}
\label{tab:dataug}
\begin{tabular}{@{}lc@{}}
\toprule
\multicolumn{1}{c}{\textbf{Parameter}} & \textbf{Value}                                               \\ \midrule
Angle of rotation in degrees           & $\left [ -10,10 \right ]$    \\
Horizontal translation shift & $0.05$ \\
Vertical translation shift   & $0.1$  \\
Scaling factor                         & $\left [ 0.95,1.05 \right ]$ \\
Horizontal flip probability  & $0.5$  \\ \bottomrule
\end{tabular}
\end{table}

\subsubsection*{Training}
We trained all the models in the different databases using the same optimisation hyper-parameters. During training, we saved the best model weights according to a decrease in the validation loss. Table~\ref{tab:hyperparameters} presents the optimisation hyper-parameters used in our experiments. We also conducted a \textit{low data regime} experiment, hence, all the models were trained using $1\%$, $10\%$, $50\%$ and $100\%$ of the training data. In imbalanced data sets (i.e., ISIC2020, MIMIC-CXR) we applied class-weights in the loss function.

\begin{table}[t]
\centering
\caption{Set of hyper-parameters used during the training step of all the phases. Please note that we present the batch size as a closed interval since we had to adapt its value during some of the training runs due to computational constraints (e.g., availability of GPU RAM).}
\label{tab:hyperparameters}
\begin{tabular}{@{}lc@{}}
\toprule
\multicolumn{1}{c}{\textbf{Hyper-parameter}} & \textbf{Value} \\ \midrule
Number of epochs                             & 300            \\
Loss function                                & Cross-entropy  \\
Optimiser                                    & Adam           \\
Learning rate                                & $1\times10^{-6}$           \\
Batch size                                   & $\left [ 2,32 \right ]$          \\ \bottomrule
\end{tabular}
\end{table}

\subsection*{Results}

\subsubsection*{Predictive Performance}
\begin{figure}[ht]
    \centering
    \includegraphics[scale=0.7]{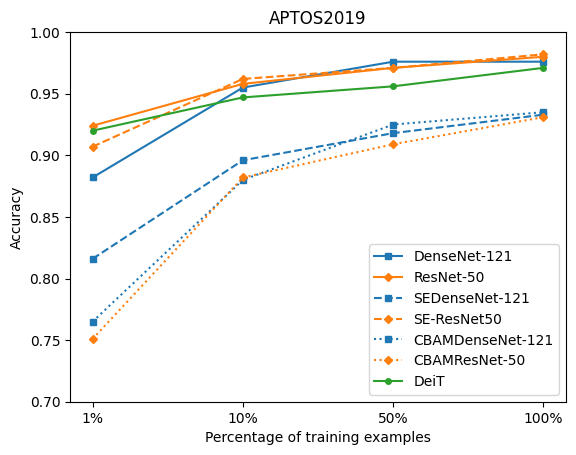}
    \caption{Predictive performance (i.e., accuracy) results of the different models on the APTOS2019 data set, using different percentages of training data.}
    \label{fig:acc-aptos}
\end{figure}

\begin{figure}[ht]
    \centering
    \includegraphics[scale=0.7]{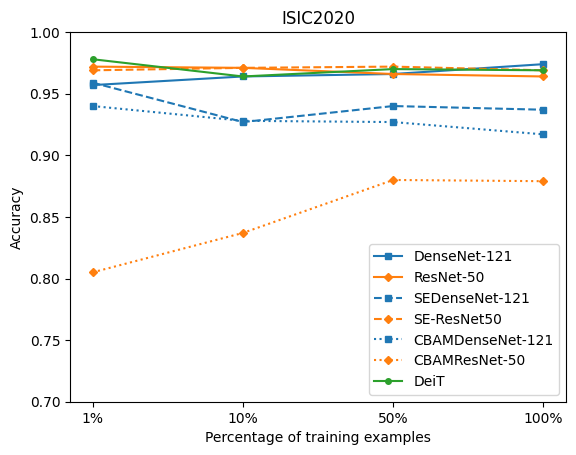}
    \caption{Predictive performance (i.e., accuracy) results of the different models on the ISIC2020 data set, using different percentages of training data.}
    \label{fig:acc-isic}
\end{figure}

\begin{figure}[ht]
    \centering
    \includegraphics[scale=0.7]{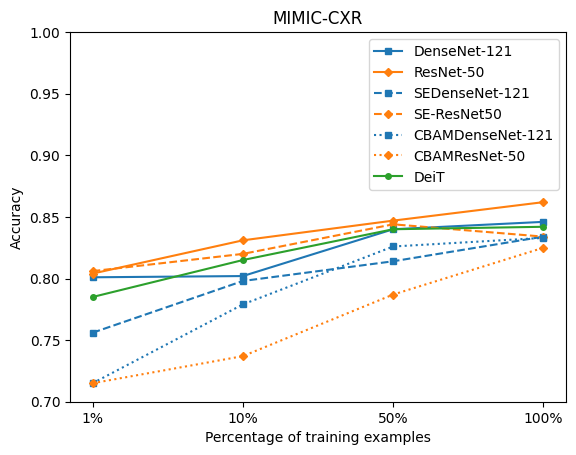}
    \caption{Predictive performance (i.e., accuracy) results of the different models on the MIMIC-CXR data set, using different percentages of training data.}
    \label{fig:acc-mimiccxr}
\end{figure}

Figures~\ref{fig:acc-aptos},~\ref{fig:acc-isic} and~\ref{fig:acc-mimiccxr} respectively present the predictive performance (i.e., accuracy) results of the different models on the APTOS2019, ISIC2020 and MIMIC-CXR data sets, using different percentages of training data.

\subsubsection*{Model Complexity}
Table~\ref{tab:model-complexity} presents the model complexity (i.e., the number of parameters) information for the models used in our experiments.

\begin{table}[t]
\centering
\caption{Model complexity (i.e., the number of parameters) information for the models used in our experiments.}
\label{tab:model-complexity}
\begin{tabular}{@{}lc@{}}
\toprule
\multicolumn{1}{c}{\textbf{Model}} & \textbf{Number of parameters} \\
\midrule
DenseNet-121     & 7,054,210 \\
ResNet-50        & 23,512,130 \\
SEDenseNet-121   & 7,357,314 \\
SEResNet-50      & 26,027,074 \\
CBAMDenseNet-121 & 7,360,706 \\
CBAMResNet-50    & 26,044,722 \\
DeiT             & 5,486,786 \\ \bottomrule
\end{tabular}
\end{table}

\subsubsection*{Post-Model Interpretability}
Tables~\ref{tab:aptos-gt0_pred0},~\ref{tab:aptos-gt1_pred1} and~\ref{tab:aptos-gt0_pred1} present examples of LRP and DeepLIFT \textit{post-hoc} saliency maps obtained for images of the APTOS2019 data set with labels $0$ and $1$ correctly classified, and with label $0$ incorrectly classified by all models, respectively. The same structure is followed for examples of the ISIC2020 (Tables~\ref{tab:isic-gt0_pred0},~\ref{tab:isic-gt1_pred1}, ~\ref{tab:isic-gt0_pred1} and~\ref{tab:isic-gt1_pred0}) and MIMIC-CXR (Tables~\ref{tab:mimiccxr-gt0_pred0},~\ref{tab:mimiccxr-gt1_pred1}, ~\ref{tab:mimiccxr-gt0_pred1} and~\ref{tab:mimiccxr-gt1_pred0}) data sets, with the addition of examples from the label $1$ incorrectly classified by all models. We used the Matplotlib~\cite{Hunter2007} library for Python to generate these saliency maps visualizations, using the ``bwr'' color map, which represents negative values in blue, zero values in white and positive values in red.

\setlength{\tabcolsep}{3pt}
\begin{table*}[ht]
\sffamily
\centering
\caption{Example of LRP and DeepLIFT \textit{post-hoc} saliency maps for an image of the APTOS2019 data set with the label $0$ correctly classified as $0$ by all models.}
\label{tab:aptos-gt0_pred0}
\begin{tabular}{c@{\hspace{.8\tabcolsep}}c@{\hspace{.3\tabcolsep}}cccc@{\hspace{1.5\tabcolsep}}ccc}
& & \multicolumn{3}{c}{\textbf{DENSENET}} & & \multicolumn{3}{c}{\textbf{RESNET}} \\
\cmidrule{3-5} \cmidrule{7-9} \\
\begin{tabular}{c}
     Original Image\\
     \includegraphics[width=.115\linewidth]{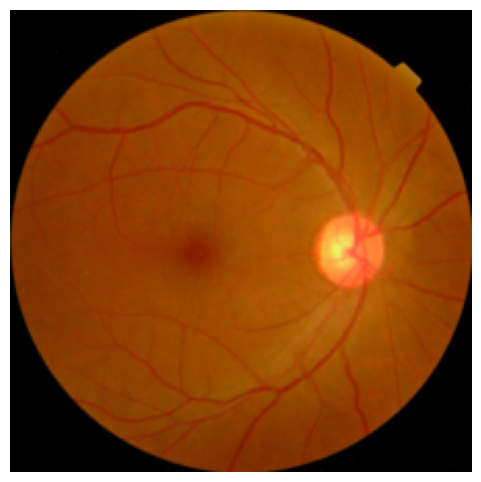}
\end{tabular} & \rotatebox[origin=c]{90}{\textbf{LRP}} & 
\begin{tabular}{c}
     No Attention\\
     \includegraphics[width=.115\linewidth]{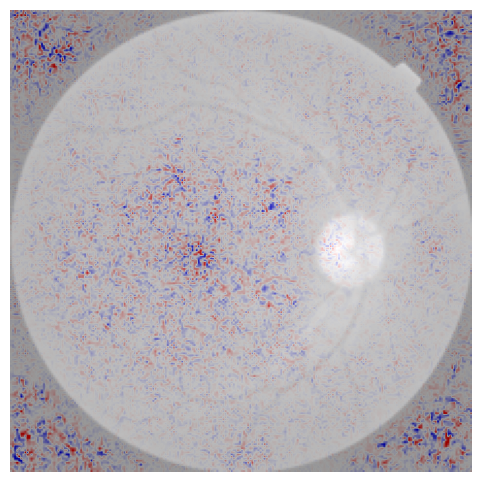}
\end{tabular} & 
\begin{tabular}{c}
     $+$ SE Layer\\
     \includegraphics[width=.115\linewidth]{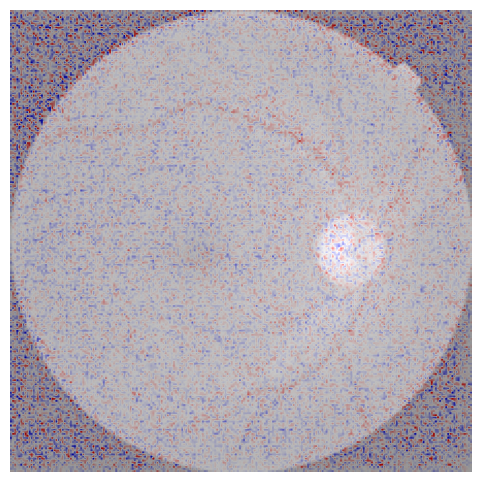}
\end{tabular} &
\begin{tabular}{c}
    $+$ CBAM Layer\\
     \includegraphics[width=.115\linewidth]{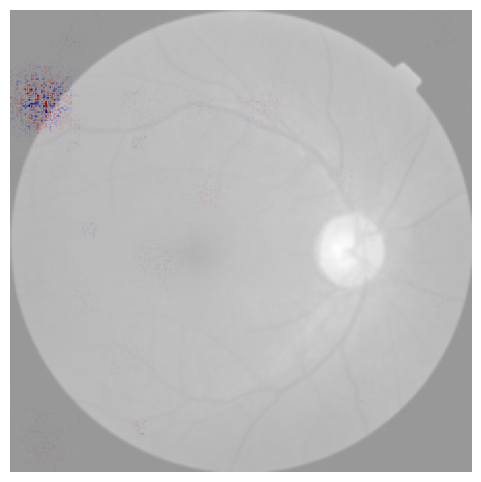}
\end{tabular} & &
\begin{tabular}{c}
     No Attention\\
     \includegraphics[width=.115\linewidth]{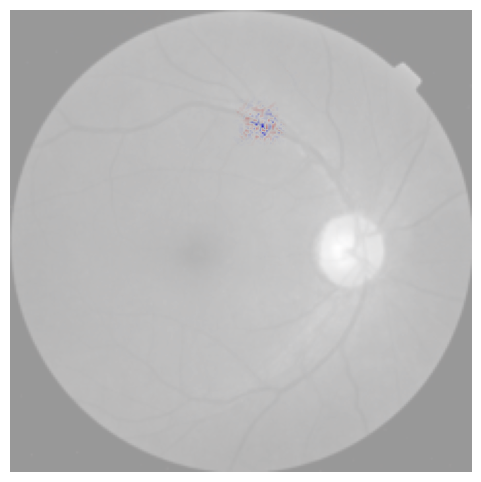}
\end{tabular} &
\begin{tabular}{c}
     $+$ SE Layer\\
     \includegraphics[width=.115\linewidth]{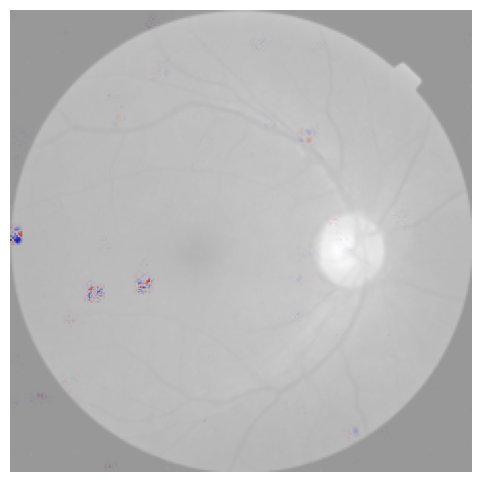}
\end{tabular} &
\begin{tabular}{c}
     $+$ CBAM Layer\\
     \includegraphics[width=.115\linewidth]{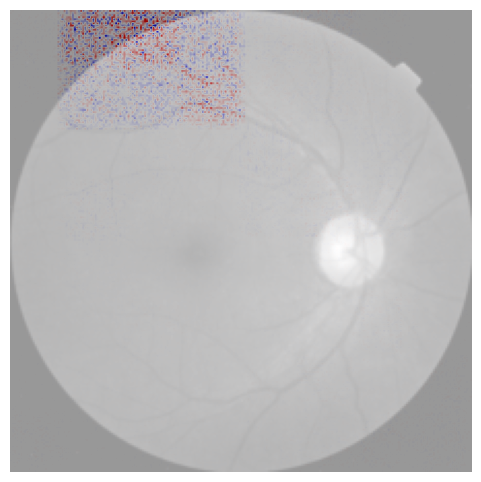}
\end{tabular}
\\
\\
\begin{tabular}{c}
    DeiT (LRP)\\
    \includegraphics[width=.115\linewidth]{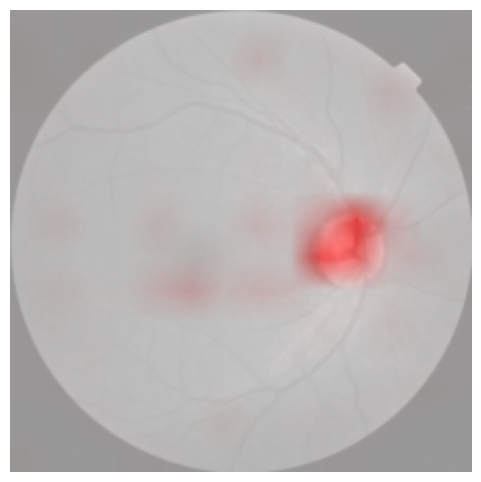}
\end{tabular} & \rotatebox[origin=c]{90}{\textbf{DeepLIFT}} &

\begin{tabular}{c}
    No Attention\\
     \includegraphics[width=.115\linewidth]{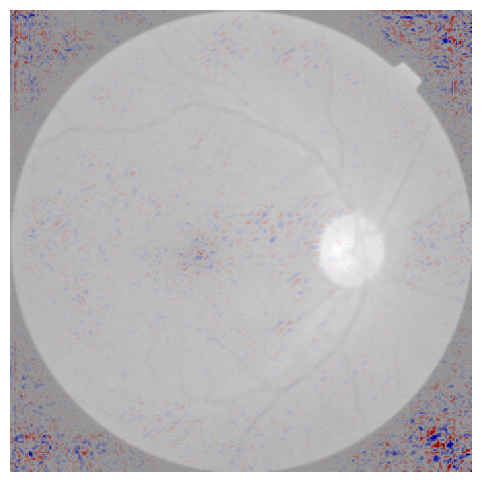}
\end{tabular} &
\begin{tabular}{c}
     $+$ SE Layer\\
     \includegraphics[width=.115\linewidth]{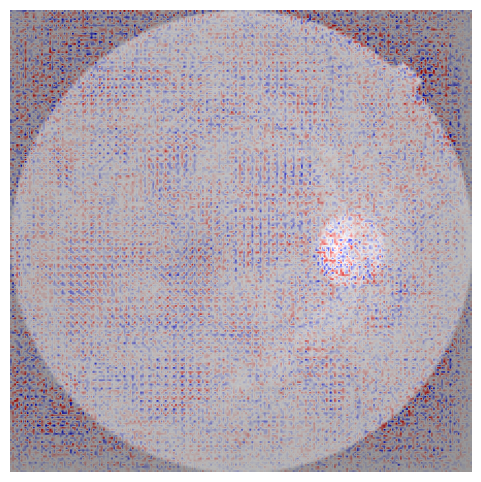}
\end{tabular} &
\begin{tabular}{c}
     $+$ CBAM Layer\\
     \includegraphics[width=.115\linewidth]{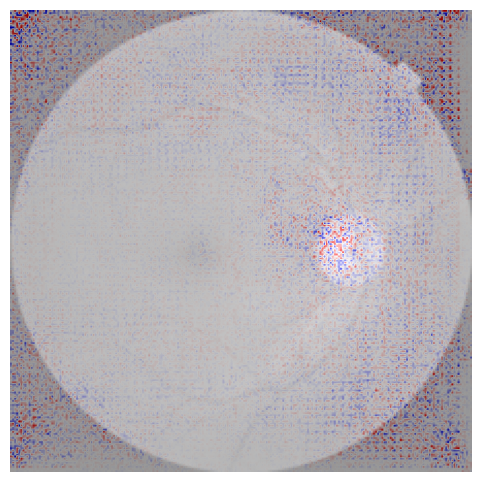}
\end{tabular} & &
\begin{tabular}{c}
     No Attention\\
     \includegraphics[width=.115\linewidth]{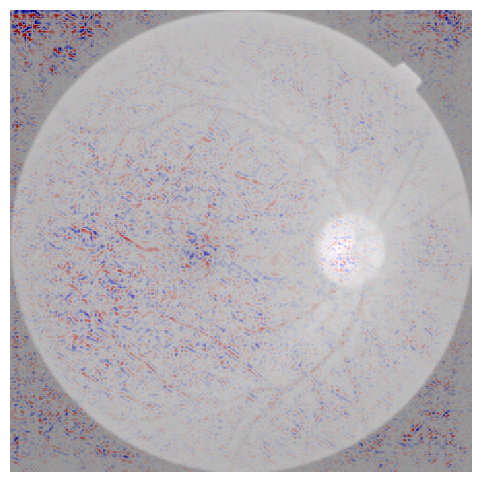}
\end{tabular} &
\begin{tabular}{c}
     $+$ SE Layer\\
     \includegraphics[width=.115\linewidth]{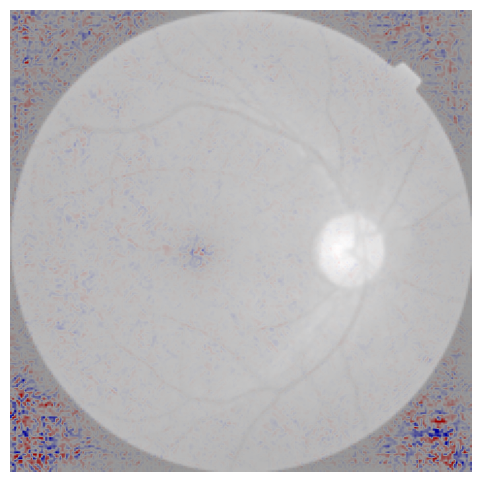}
\end{tabular} &
\begin{tabular}{c}
     $+$ CBAM Layer\\
     \includegraphics[width=.115\linewidth]{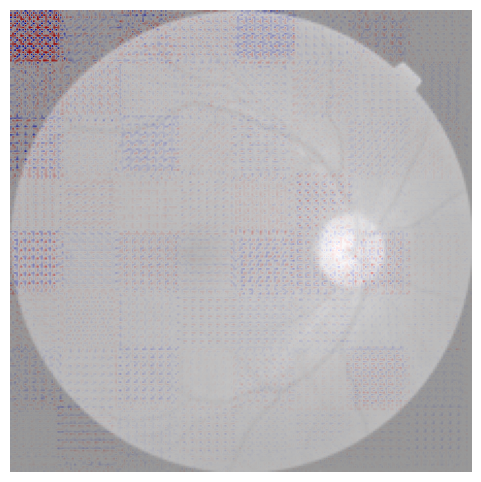}
\end{tabular}

\end{tabular}
\end{table*}

\setlength{\tabcolsep}{3pt}
\begin{table*}[ht]
\sffamily
\centering
\caption{Example of LRP and DeepLIFT \textit{post-hoc} saliency maps for an image of the APTOS2019 data set with the label $1$ correctly classified as $1$ by all models.}
\label{tab:aptos-gt1_pred1}
\begin{tabular}{c@{\hspace{.8\tabcolsep}}c@{\hspace{.3\tabcolsep}}cccc@{\hspace{1.5\tabcolsep}}ccc}
& & \multicolumn{3}{c}{\textbf{DENSENET}} & & \multicolumn{3}{c}{\textbf{RESNET}} \\
\cmidrule{3-5} \cmidrule{7-9} \\
\begin{tabular}{c}
     Original Image\\
     \includegraphics[width=.115\linewidth]{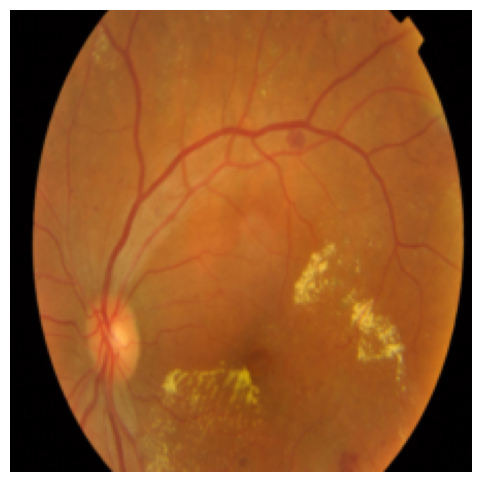}
\end{tabular} & \rotatebox[origin=c]{90}{\textbf{LRP}} & 
\begin{tabular}{c}
     No Attention\\
     \includegraphics[width=.115\linewidth]{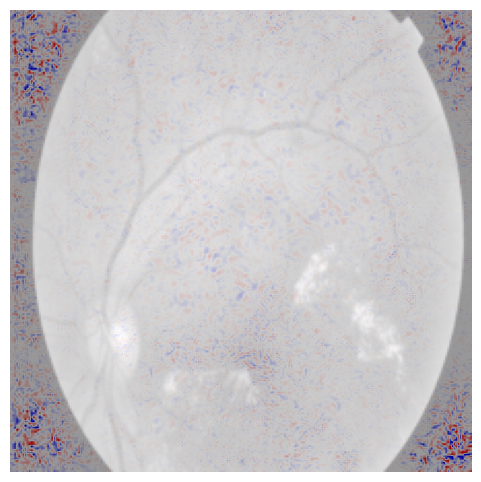}
\end{tabular} & 
\begin{tabular}{c}
     $+$ SE Layer\\
     \includegraphics[width=.115\linewidth]{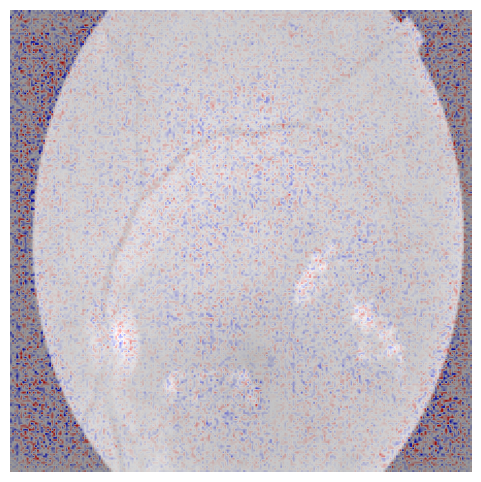}
\end{tabular} &
\begin{tabular}{c}
    $+$ CBAM Layer\\
     \includegraphics[width=.115\linewidth]{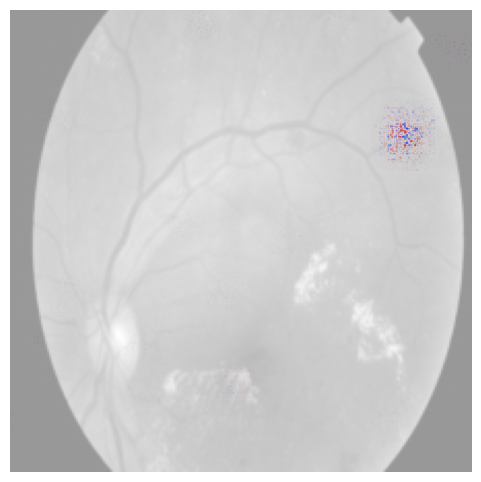}
\end{tabular} & &
\begin{tabular}{c}
     No Attention\\
     \includegraphics[width=.115\linewidth]{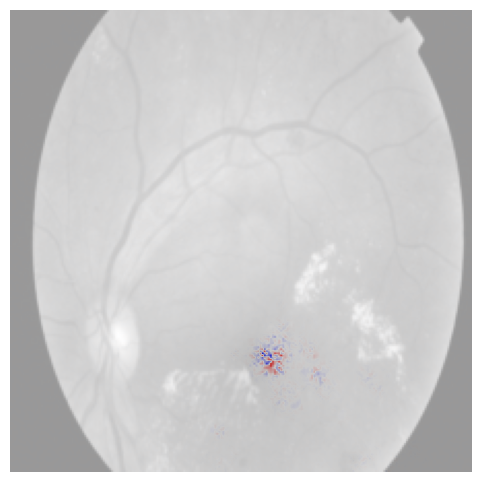}
\end{tabular} &
\begin{tabular}{c}
     $+$ SE Layer\\
     \includegraphics[width=.115\linewidth]{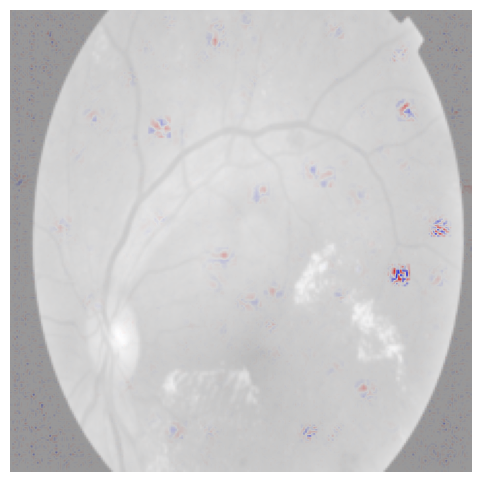}
\end{tabular} &
\begin{tabular}{c}
     $+$ CBAM Layer\\
     \includegraphics[width=.115\linewidth]{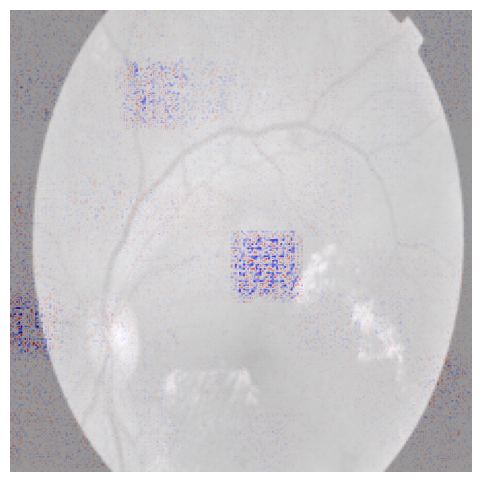}
\end{tabular}
\\
\\
\begin{tabular}{c}
    DeiT (LRP)\\
    \includegraphics[width=.115\linewidth]{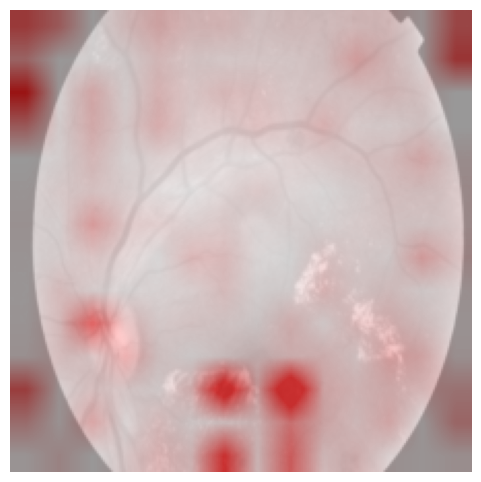}
\end{tabular} & \rotatebox[origin=c]{90}{\textbf{DeepLIFT}} &

\begin{tabular}{c}
    No Attention\\
     \includegraphics[width=.115\linewidth]{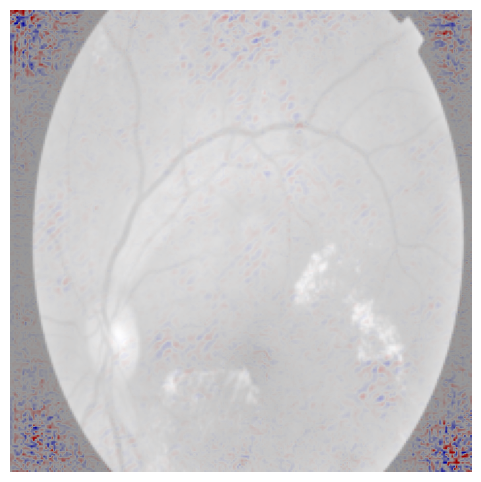}
\end{tabular} &
\begin{tabular}{c}
     $+$ SE Layer\\
     \includegraphics[width=.115\linewidth]{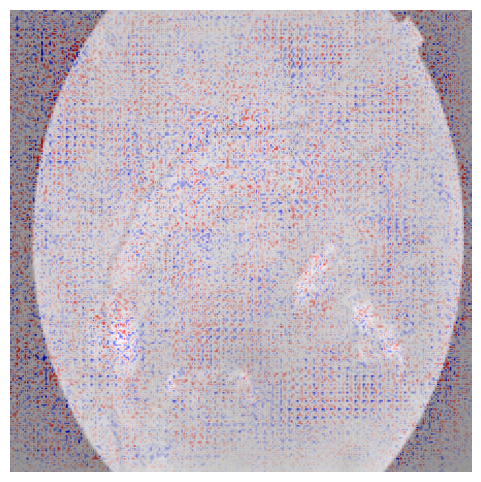}
\end{tabular} &
\begin{tabular}{c}
     $+$ CBAM Layer\\
     \includegraphics[width=.115\linewidth]{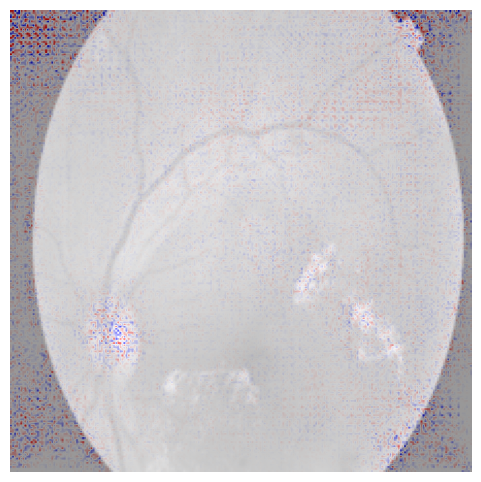}
\end{tabular} & &
\begin{tabular}{c}
     No Attention\\
     \includegraphics[width=.115\linewidth]{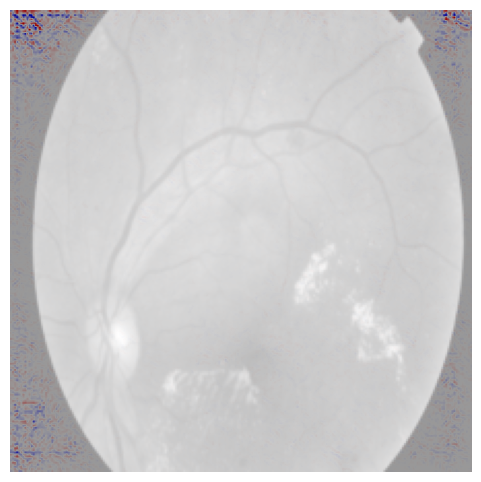}
\end{tabular} &
\begin{tabular}{c}
     $+$ SE Layer\\
     \includegraphics[width=.115\linewidth]{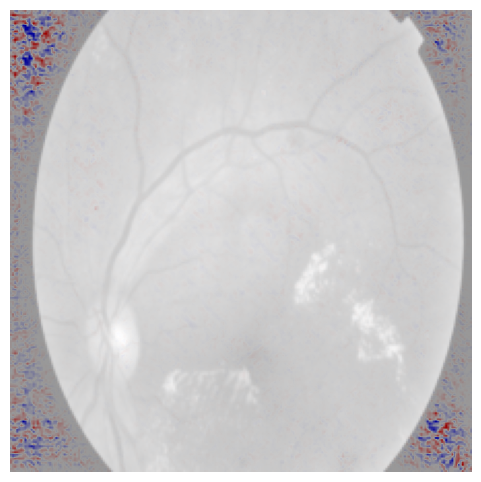}
\end{tabular} &
\begin{tabular}{c}
     $+$ CBAM Layer\\
     \includegraphics[width=.115\linewidth]{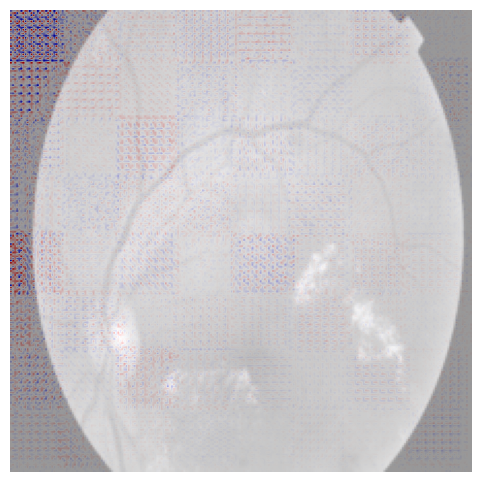}
\end{tabular}

\end{tabular}
\end{table*}

\setlength{\tabcolsep}{3pt}
\begin{table*}[ht]
\sffamily
\centering
\caption{Example of LRP and DeepLIFT \textit{post-hoc} saliency maps for an image of the APTOS2019 data set with the label $0$ incorrectly classified as $1$ by all models.}
\label{tab:aptos-gt0_pred1}
\begin{tabular}{c@{\hspace{.8\tabcolsep}}c@{\hspace{.3\tabcolsep}}cccc@{\hspace{1.5\tabcolsep}}ccc}
& & \multicolumn{3}{c}{\textbf{DENSENET}} & & \multicolumn{3}{c}{\textbf{RESNET}} \\
\cmidrule{3-5} \cmidrule{7-9} \\
\begin{tabular}{c}
     Original Image\\
     \includegraphics[width=.115\linewidth]{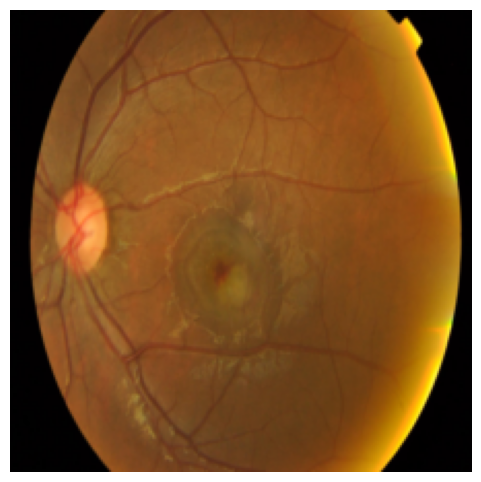}
\end{tabular} & \rotatebox[origin=c]{90}{\textbf{LRP}} & 
\begin{tabular}{c}
     No Attention\\
     \includegraphics[width=.115\linewidth]{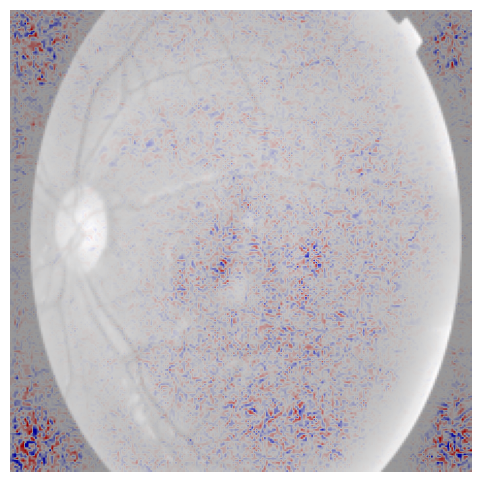}
\end{tabular} & 
\begin{tabular}{c}
     $+$ SE Layer\\
     \includegraphics[width=.115\linewidth]{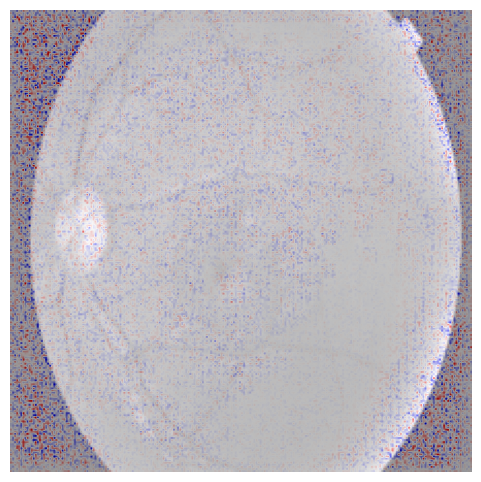}
\end{tabular} &
\begin{tabular}{c}
    $+$ CBAM Layer\\
     \includegraphics[width=.115\linewidth]{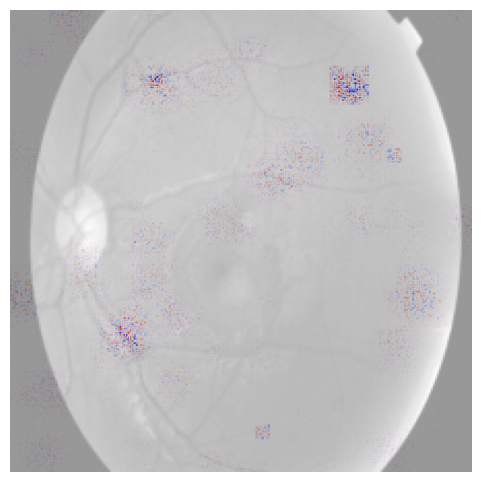}
\end{tabular} & &
\begin{tabular}{c}
     No Attention\\
     \includegraphics[width=.115\linewidth]{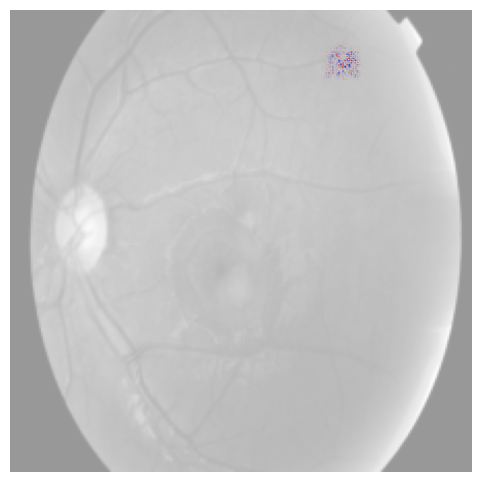}
\end{tabular} &
\begin{tabular}{c}
     $+$ SE Layer\\
     \includegraphics[width=.115\linewidth]{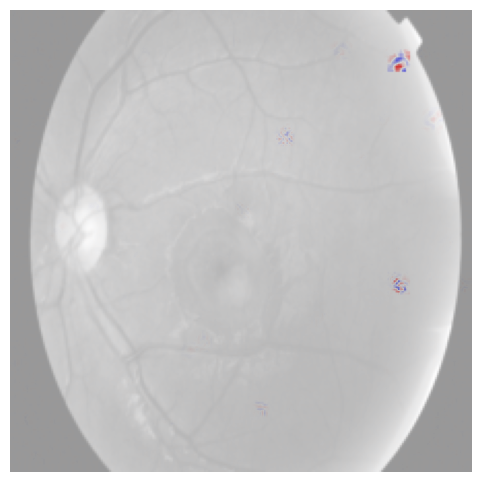}
\end{tabular} &
\begin{tabular}{c}
     $+$ CBAM Layer\\
     \includegraphics[width=.115\linewidth]{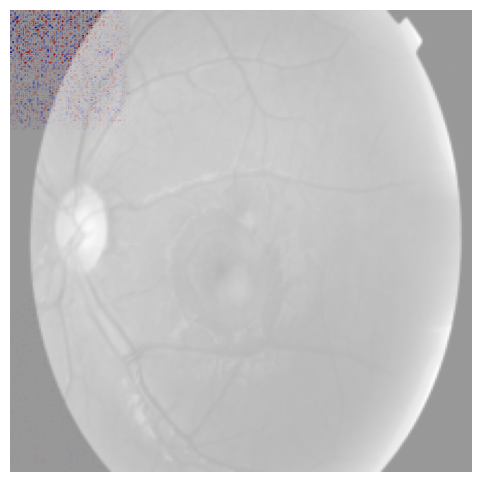}
\end{tabular}
\\
\\
\begin{tabular}{c}
    DeiT (LRP)\\
    \includegraphics[width=.115\linewidth]{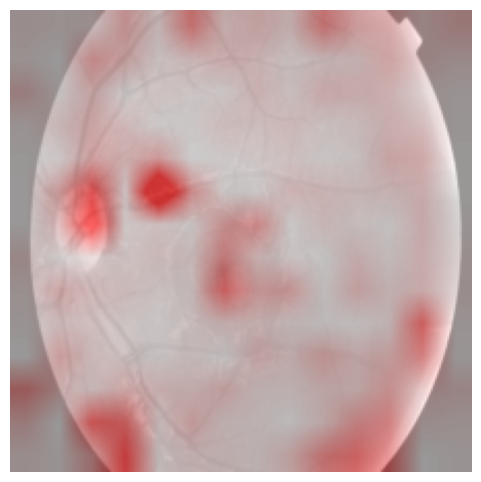}
\end{tabular} & \rotatebox[origin=c]{90}{\textbf{DeepLIFT}} &

\begin{tabular}{c}
    No Attention\\
     \includegraphics[width=.115\linewidth]{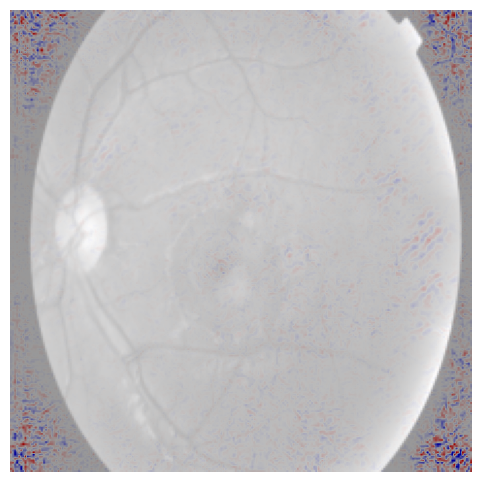}
\end{tabular} &
\begin{tabular}{c}
     $+$ SE Layer\\
     \includegraphics[width=.115\linewidth]{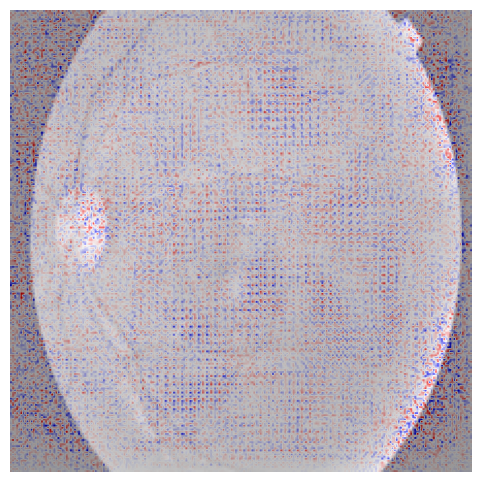}
\end{tabular} &
\begin{tabular}{c}
     $+$ CBAM Layer\\
     \includegraphics[width=.115\linewidth]{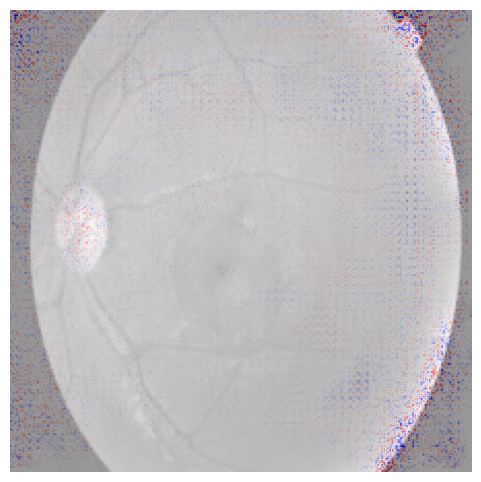}
\end{tabular} & &
\begin{tabular}{c}
     No Attention\\
     \includegraphics[width=.115\linewidth]{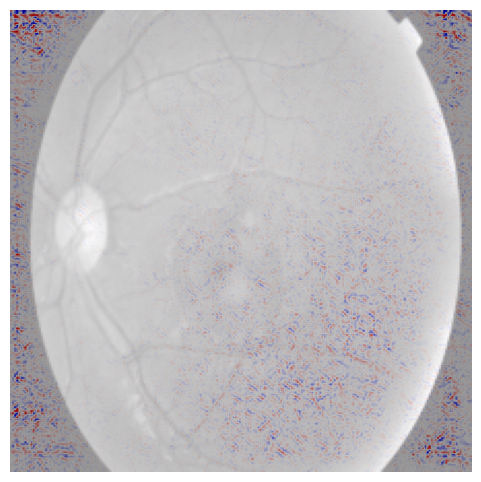}
\end{tabular} &
\begin{tabular}{c}
     $+$ SE Layer\\
     \includegraphics[width=.115\linewidth]{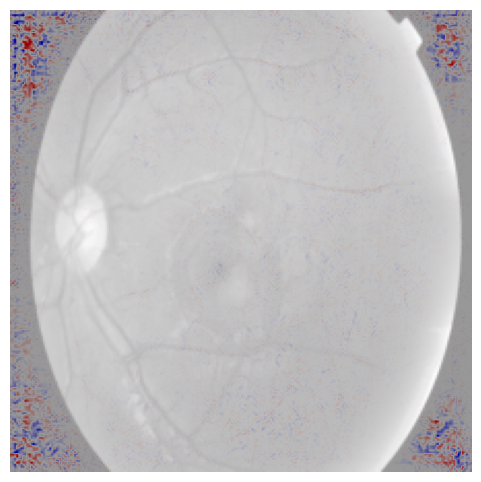}
\end{tabular} &
\begin{tabular}{c}
     $+$ CBAM Layer\\
     \includegraphics[width=.115\linewidth]{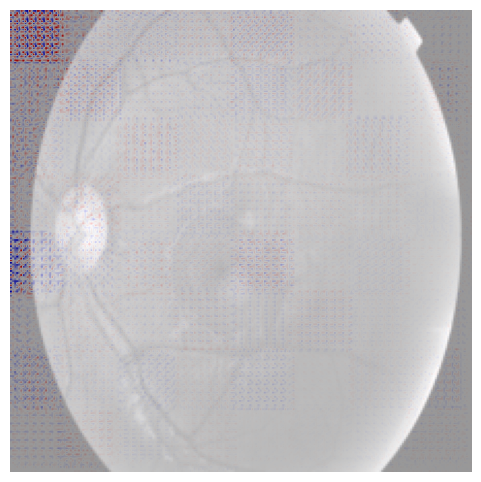}
\end{tabular}

\end{tabular}
\end{table*}

\setlength{\tabcolsep}{3pt}
\begin{table*}[ht]
\sffamily
\centering
\caption{Example of LRP and DeepLIFT \textit{post-hoc} saliency maps for an image of the ISIC2020 data set with the label $0$ correctly classified as $0$ by all models.}
\label{tab:isic-gt0_pred0}
\begin{tabular}{c@{\hspace{.8\tabcolsep}}c@{\hspace{.3\tabcolsep}}cccc@{\hspace{1.5\tabcolsep}}ccc}
& & \multicolumn{3}{c}{\textbf{DENSENET}} & & \multicolumn{3}{c}{\textbf{RESNET}} \\
\cmidrule{3-5} \cmidrule{7-9} \\
\begin{tabular}{c}
     Original Image\\
     \includegraphics[width=.115\linewidth]{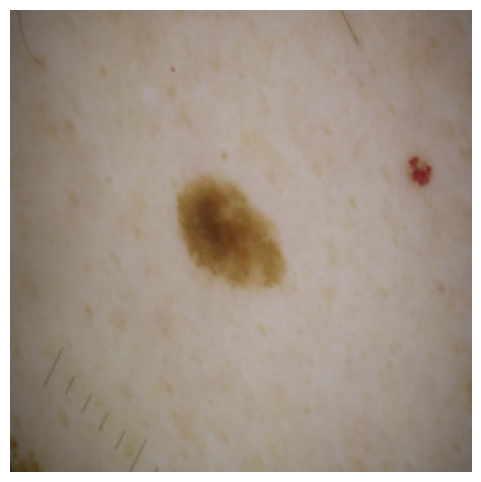}
\end{tabular} & \rotatebox[origin=c]{90}{\textbf{LRP}} & 
\begin{tabular}{c}
     No Attention\\
     \includegraphics[width=.115\linewidth]{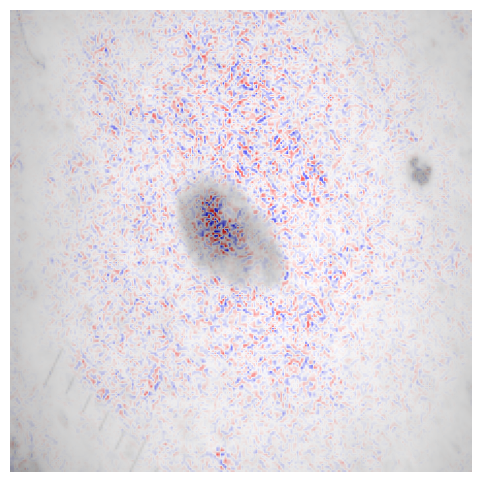}
\end{tabular} & 
\begin{tabular}{c}
     $+$ SE Layer\\
     \includegraphics[width=.115\linewidth]{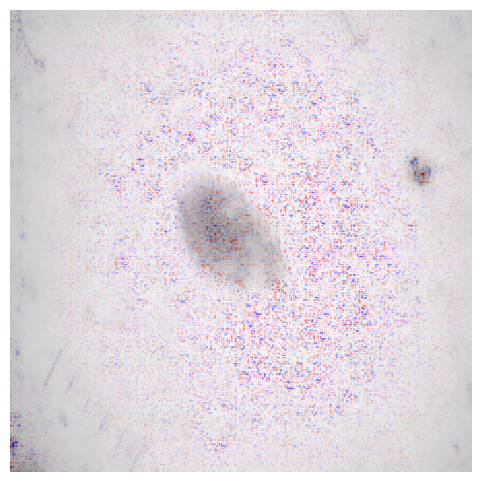}
\end{tabular} &
\begin{tabular}{c}
    $+$ CBAM Layer\\
     \includegraphics[width=.115\linewidth]{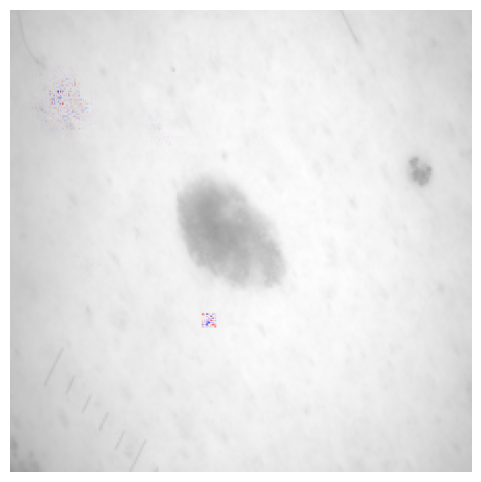}
\end{tabular} & &
\begin{tabular}{c}
     No Attention\\
     \includegraphics[width=.115\linewidth]{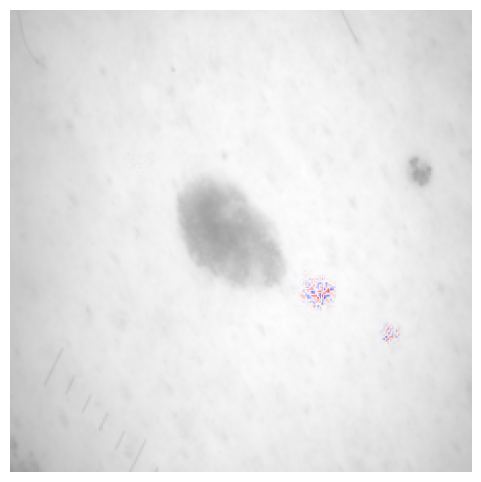}
\end{tabular} &
\begin{tabular}{c}
     $+$ SE Layer\\
     \includegraphics[width=.115\linewidth]{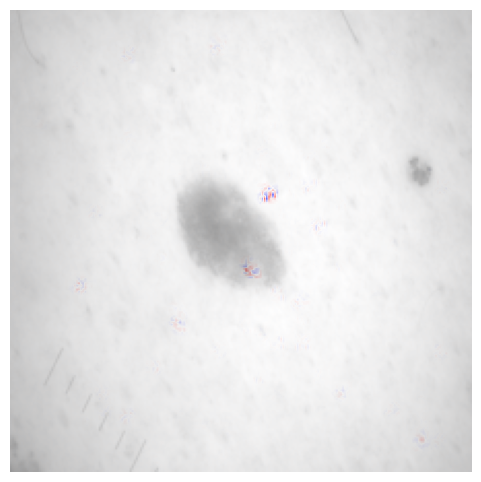}
\end{tabular} &
\begin{tabular}{c}
     $+$ CBAM Layer\\
     \includegraphics[width=.115\linewidth]{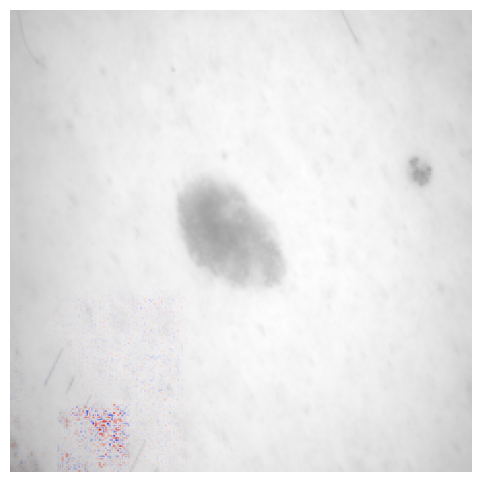}
\end{tabular}
\\
\\
\begin{tabular}{c}
    DeiT (LRP)\\
    \includegraphics[width=.115\linewidth]{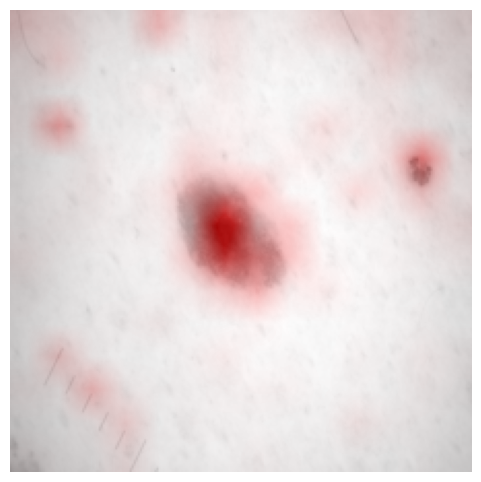}
\end{tabular} & \rotatebox[origin=c]{90}{\textbf{DeepLIFT}} &

\begin{tabular}{c}
    No Attention\\
     \includegraphics[width=.115\linewidth]{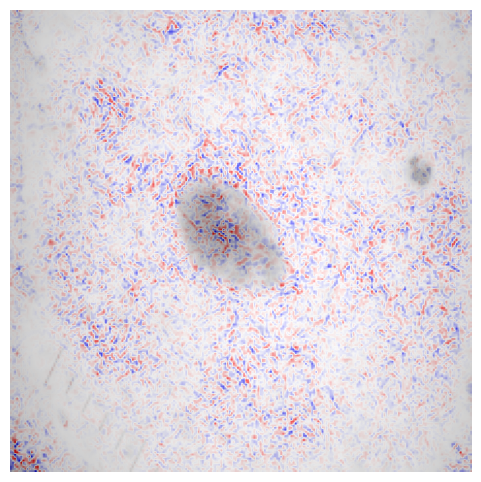}
\end{tabular} &
\begin{tabular}{c}
     $+$ SE Layer\\
     \includegraphics[width=.115\linewidth]{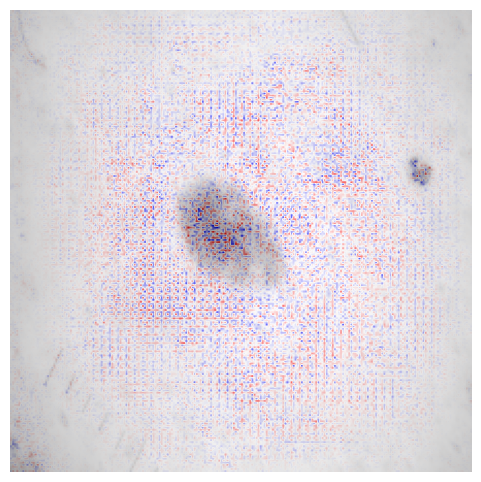}
\end{tabular} &
\begin{tabular}{c}
     $+$ CBAM Layer\\
     \includegraphics[width=.115\linewidth]{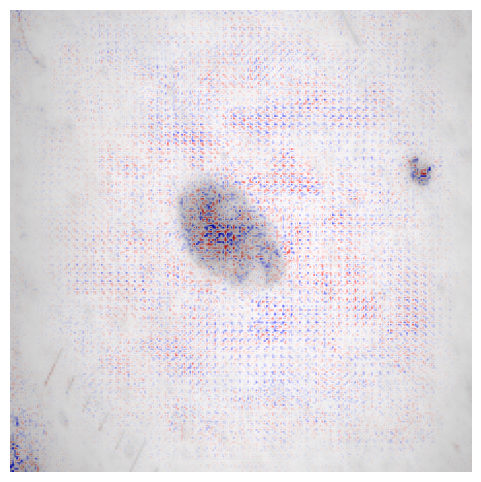}
\end{tabular} & &
\begin{tabular}{c}
     No Attention\\
     \includegraphics[width=.115\linewidth]{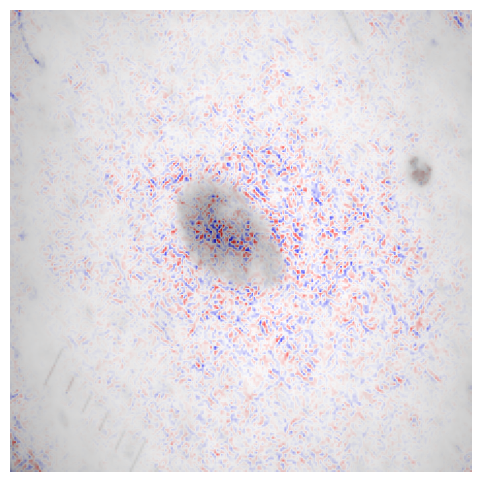}
\end{tabular} &
\begin{tabular}{c}
     $+$ SE Layer\\
     \includegraphics[width=.115\linewidth]{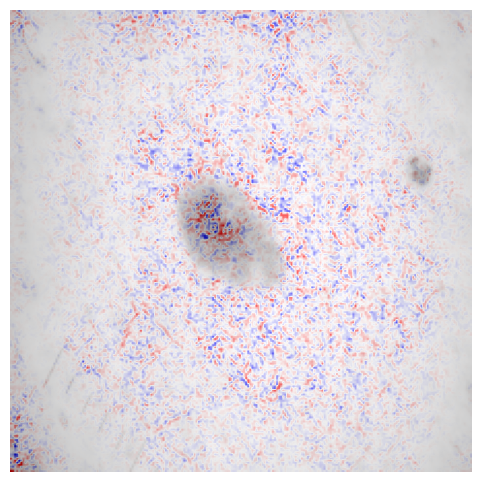}
\end{tabular} &
\begin{tabular}{c}
     $+$ CBAM Layer\\
     \includegraphics[width=.115\linewidth]{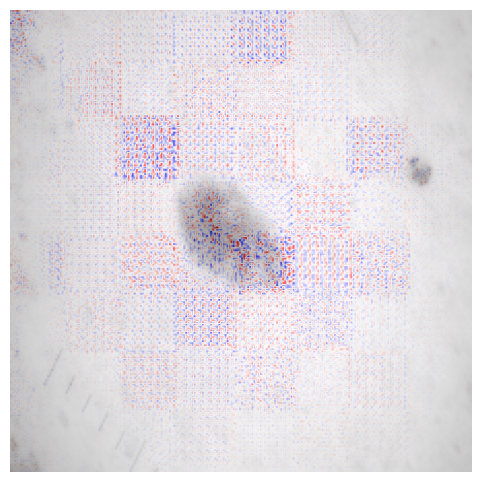}
\end{tabular}

\end{tabular}
\end{table*}

\setlength{\tabcolsep}{3pt}
\begin{table*}[ht]
\sffamily
\centering
\caption{Example of LRP and DeepLIFT \textit{post-hoc} saliency maps for an image of the ISIC2020 data set with the label $1$ correctly classified as $1$ by all models.}
\label{tab:isic-gt1_pred1}
\begin{tabular}{c@{\hspace{.8\tabcolsep}}c@{\hspace{.3\tabcolsep}}cccc@{\hspace{1.5\tabcolsep}}ccc}
& & \multicolumn{3}{c}{\textbf{DENSENET}} & & \multicolumn{3}{c}{\textbf{RESNET}} \\
\cmidrule{3-5} \cmidrule{7-9} \\
\begin{tabular}{c}
     Original Image\\
     \includegraphics[width=.115\linewidth]{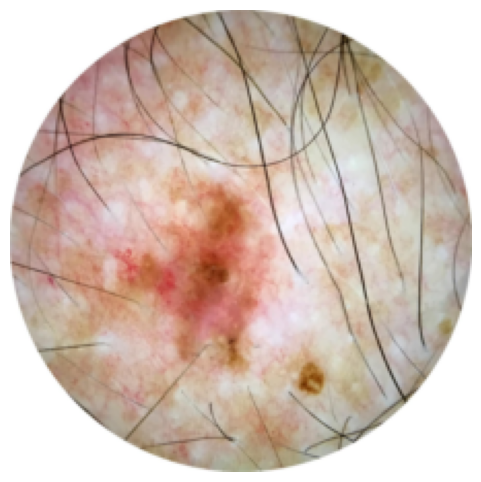}
\end{tabular} & \rotatebox[origin=c]{90}{\textbf{LRP}} & 
\begin{tabular}{c}
     No Attention\\
     \includegraphics[width=.115\linewidth]{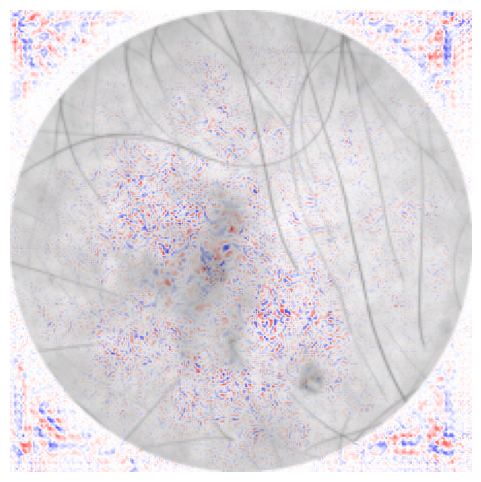}
\end{tabular} & 
\begin{tabular}{c}
     $+$ SE Layer\\
     \includegraphics[width=.115\linewidth]{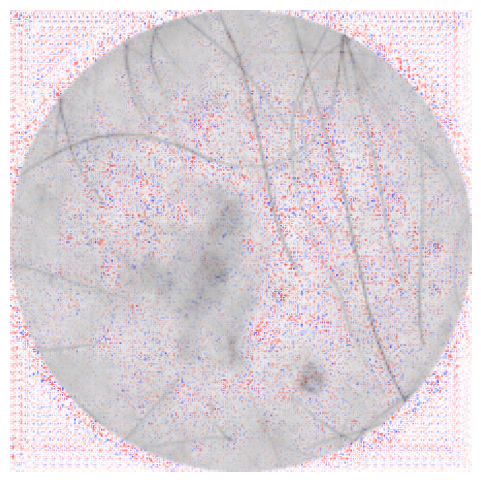}
\end{tabular} &
\begin{tabular}{c}
    $+$ CBAM Layer\\
     \includegraphics[width=.115\linewidth]{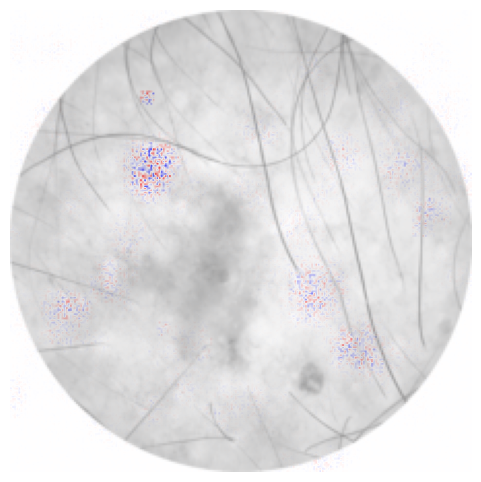}
\end{tabular} & &
\begin{tabular}{c}
     No Attention\\
     \includegraphics[width=.115\linewidth]{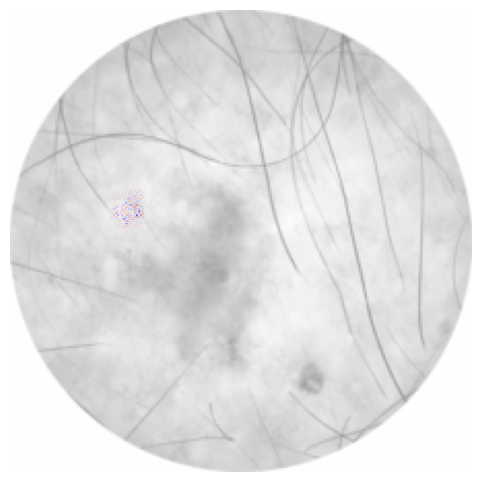}
\end{tabular} &
\begin{tabular}{c}
     $+$ SE Layer\\
     \includegraphics[width=.115\linewidth]{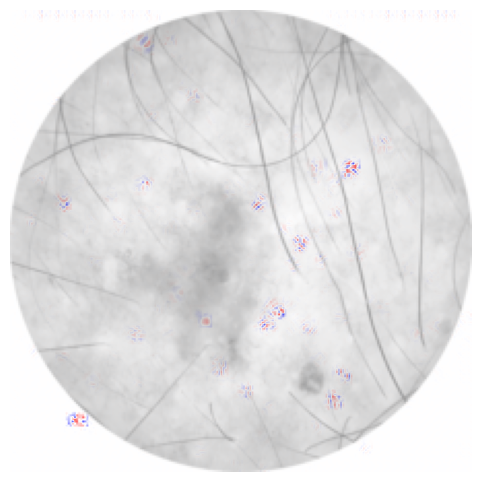}
\end{tabular} &
\begin{tabular}{c}
     $+$ CBAM Layer\\
     \includegraphics[width=.115\linewidth]{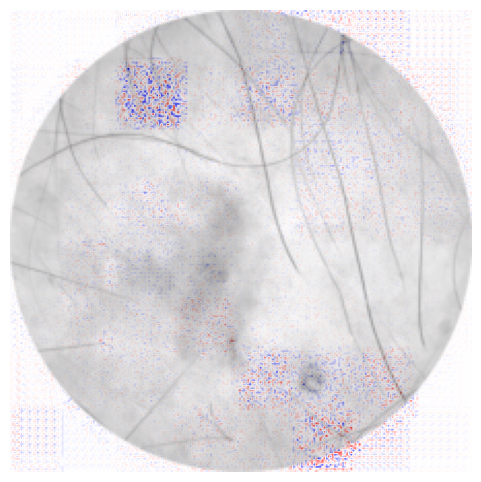}
\end{tabular}
\\
\\
\begin{tabular}{c}
    DeiT (LRP)\\
    \includegraphics[width=.115\linewidth]{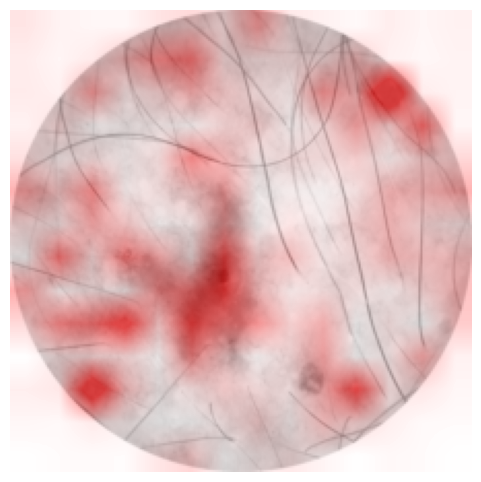}
\end{tabular} & \rotatebox[origin=c]{90}{\textbf{DeepLIFT}} &

\begin{tabular}{c}
    No Attention\\
     \includegraphics[width=.115\linewidth]{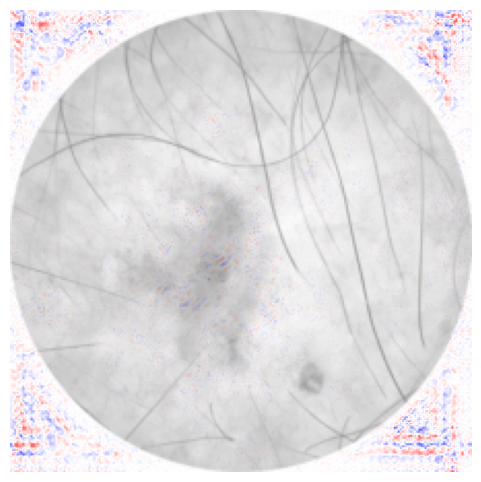}
\end{tabular} &
\begin{tabular}{c}
     $+$ SE Layer\\
     \includegraphics[width=.115\linewidth]{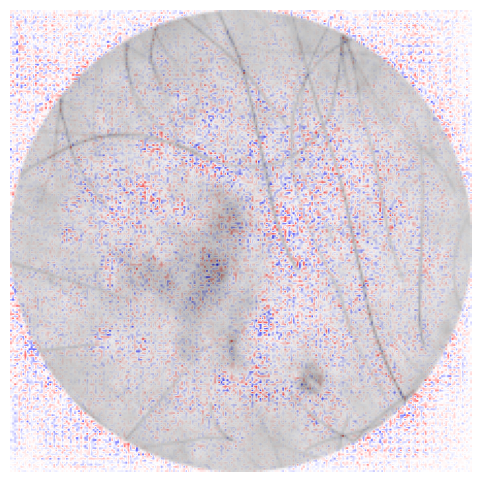}
\end{tabular} &
\begin{tabular}{c}
     $+$ CBAM Layer\\
     \includegraphics[width=.115\linewidth]{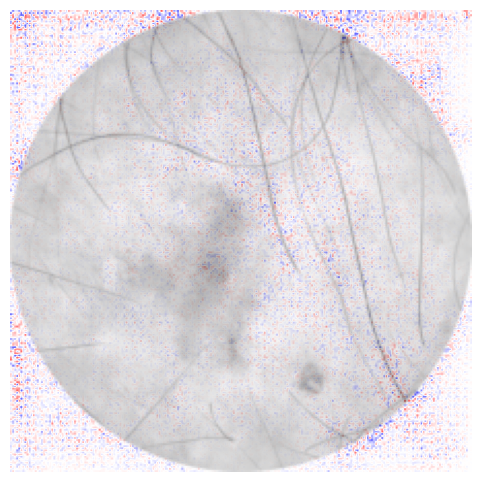}
\end{tabular} & &
\begin{tabular}{c}
     No Attention\\
     \includegraphics[width=.115\linewidth]{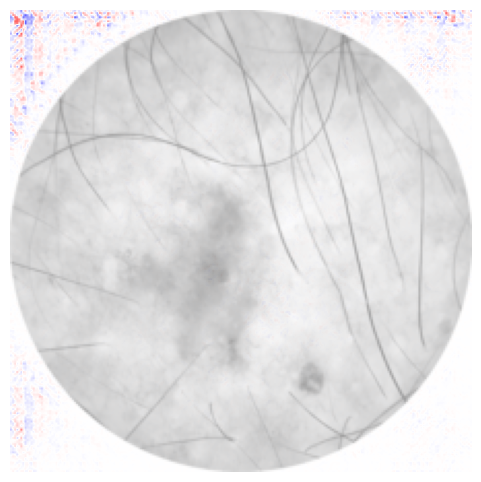}
\end{tabular} &
\begin{tabular}{c}
     $+$ SE Layer\\
     \includegraphics[width=.115\linewidth]{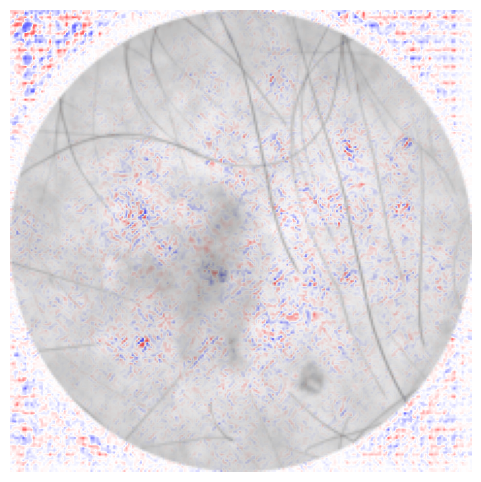}
\end{tabular} &
\begin{tabular}{c}
     $+$ CBAM Layer\\
     \includegraphics[width=.115\linewidth]{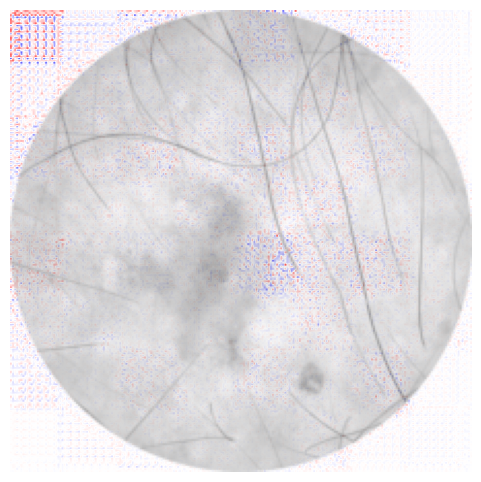}
\end{tabular}

\end{tabular}
\end{table*}

\setlength{\tabcolsep}{3pt}
\begin{table*}[ht]
\sffamily
\centering
\caption{Example of LRP and DeepLIFT \textit{post-hoc} saliency maps for an image of the ISIC2020 data set with the label $0$ incorrectly classified as $1$ by all models.}
\label{tab:isic-gt0_pred1}
\begin{tabular}{c@{\hspace{.8\tabcolsep}}c@{\hspace{.3\tabcolsep}}cccc@{\hspace{1.5\tabcolsep}}ccc}
& & \multicolumn{3}{c}{\textbf{DENSENET}} & & \multicolumn{3}{c}{\textbf{RESNET}} \\
\cmidrule{3-5} \cmidrule{7-9} \\
\begin{tabular}{c}
     Original Image\\
     \includegraphics[width=.115\linewidth]{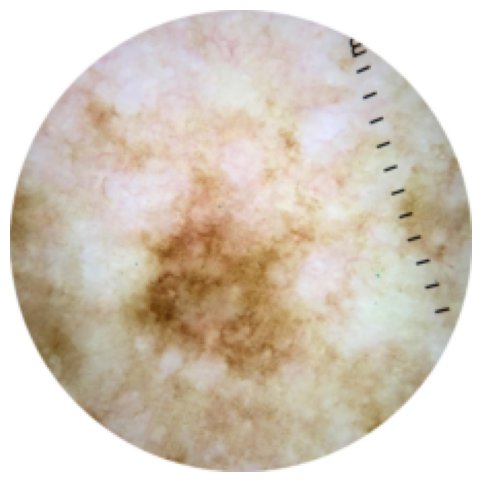}
\end{tabular} & \rotatebox[origin=c]{90}{\textbf{LRP}} & 
\begin{tabular}{c}
     No Attention\\
     \includegraphics[width=.115\linewidth]{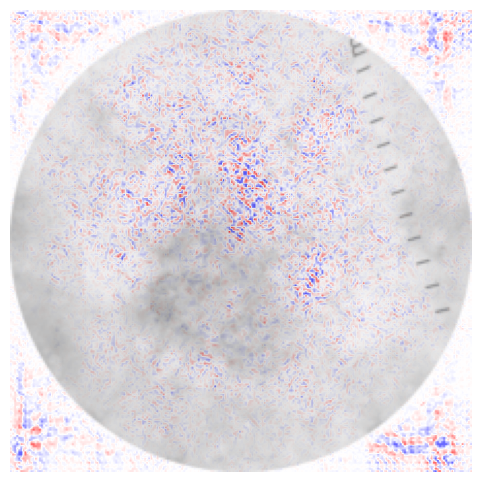}
\end{tabular} & 
\begin{tabular}{c}
     $+$ SE Layer\\
     \includegraphics[width=.115\linewidth]{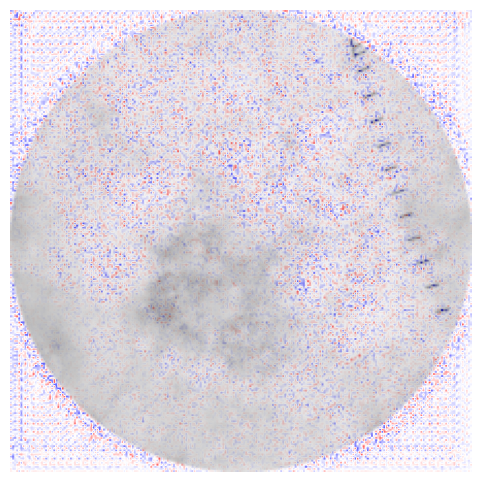}
\end{tabular} &
\begin{tabular}{c}
    $+$ CBAM Layer\\
     \includegraphics[width=.115\linewidth]{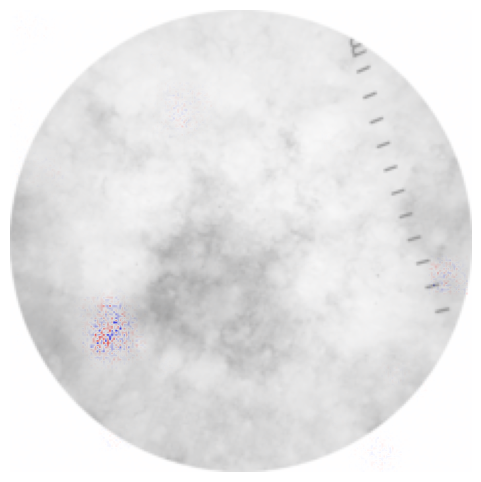}
\end{tabular} & &
\begin{tabular}{c}
     No Attention\\
     \includegraphics[width=.115\linewidth]{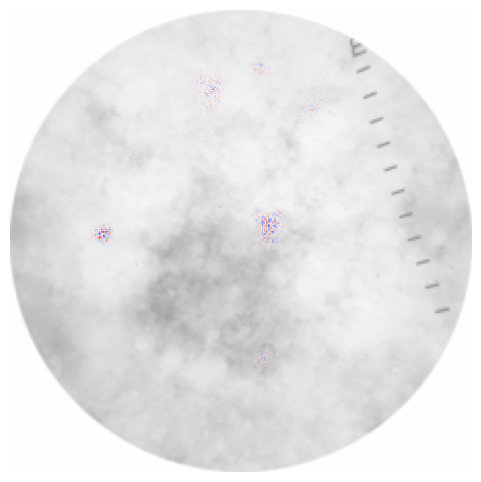}
\end{tabular} &
\begin{tabular}{c}
     $+$ SE Layer\\
     \includegraphics[width=.115\linewidth]{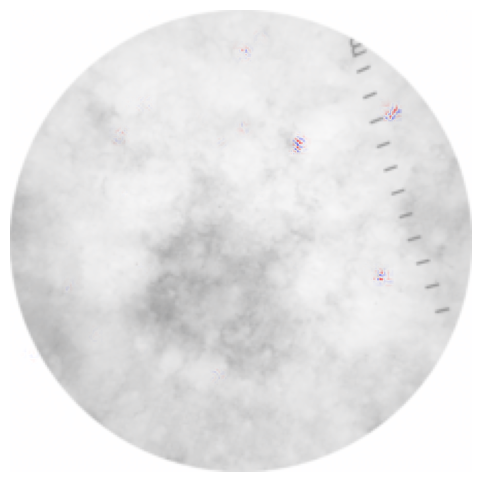}
\end{tabular} &
\begin{tabular}{c}
     $+$ CBAM Layer\\
     \includegraphics[width=.115\linewidth]{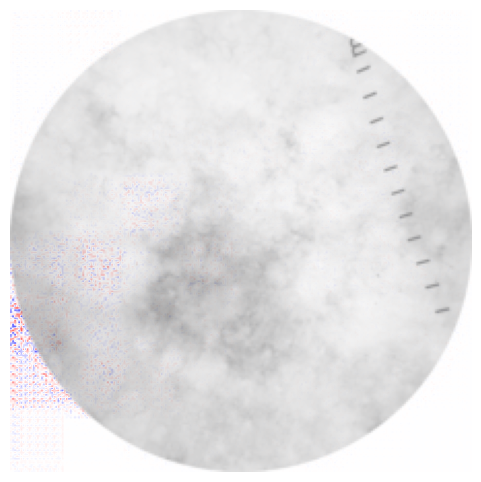}
\end{tabular}
\\
\\
\begin{tabular}{c}
    DeiT (LRP)\\
    \includegraphics[width=.115\linewidth]{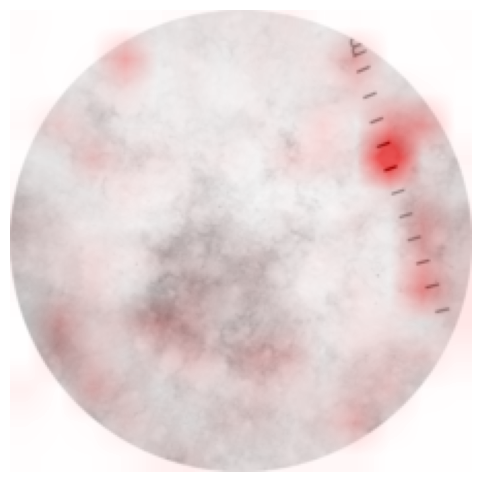}
\end{tabular} & \rotatebox[origin=c]{90}{\textbf{DeepLIFT}} &

\begin{tabular}{c}
    No Attention\\
     \includegraphics[width=.115\linewidth]{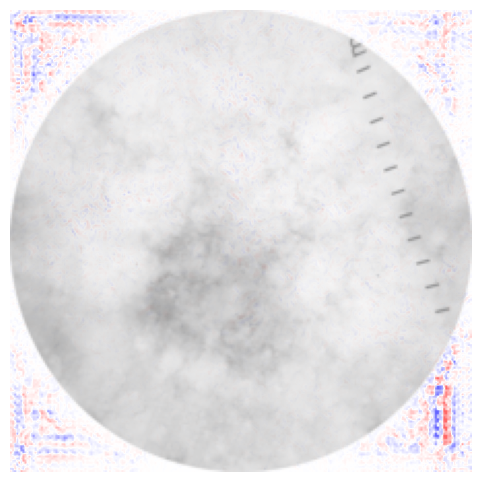}
\end{tabular} &
\begin{tabular}{c}
     $+$ SE Layer\\
     \includegraphics[width=.115\linewidth]{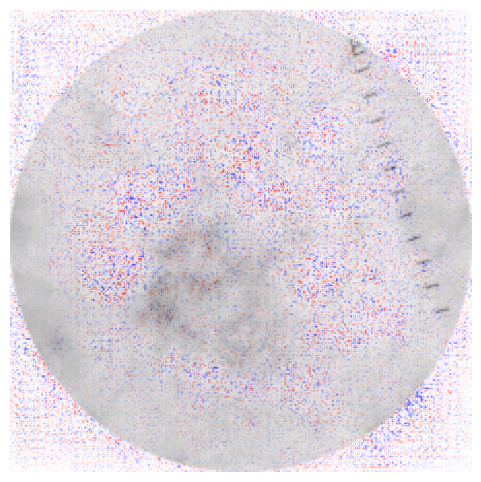}
\end{tabular} &
\begin{tabular}{c}
     $+$ CBAM Layer\\
     \includegraphics[width=.115\linewidth]{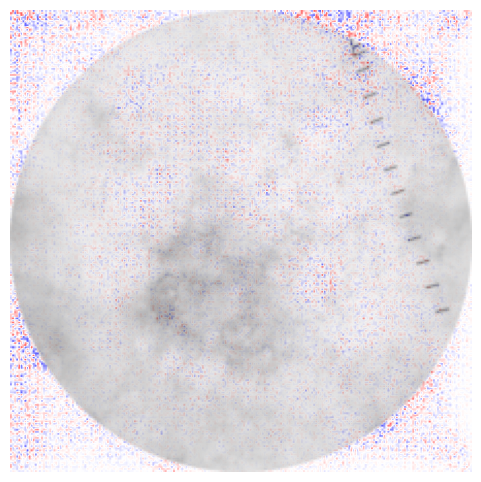}
\end{tabular} & &
\begin{tabular}{c}
     No Attention\\
     \includegraphics[width=.115\linewidth]{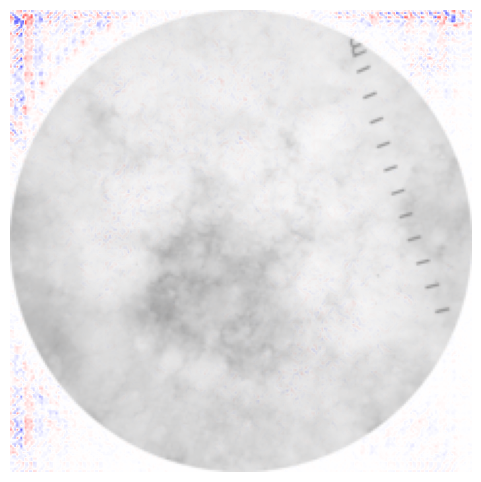}
\end{tabular} &
\begin{tabular}{c}
     $+$ SE Layer\\
     \includegraphics[width=.115\linewidth]{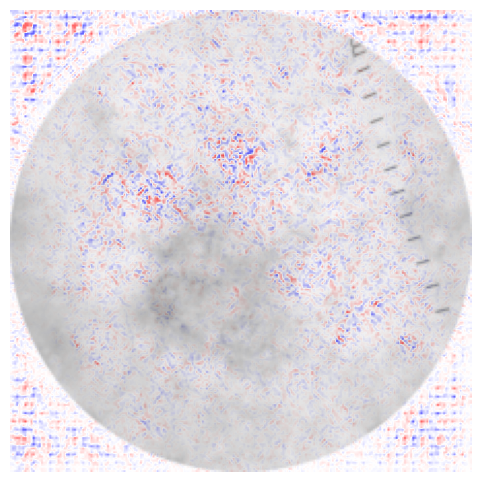}
\end{tabular} &
\begin{tabular}{c}
     $+$ CBAM Layer\\
     \includegraphics[width=.115\linewidth]{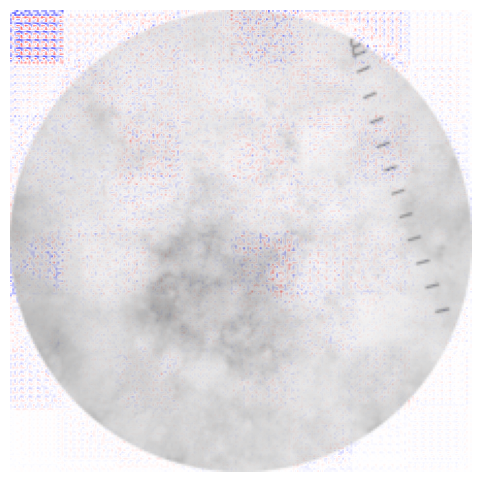}
\end{tabular}

\end{tabular}
\end{table*}

\setlength{\tabcolsep}{3pt}
\begin{table*}[ht]
\sffamily
\centering
\caption{Example of LRP and DeepLIFT \textit{post-hoc} saliency maps for an image of the ISIC2020 data set with the label $1$ incorrectly classified as $0$ by all models.}
\label{tab:isic-gt1_pred0}
\begin{tabular}{c@{\hspace{.8\tabcolsep}}c@{\hspace{.3\tabcolsep}}cccc@{\hspace{1.5\tabcolsep}}ccc}
& & \multicolumn{3}{c}{\textbf{DENSENET}} & & \multicolumn{3}{c}{\textbf{RESNET}} \\
\cmidrule{3-5} \cmidrule{7-9} \\
\begin{tabular}{c}
     Original Image\\
     \includegraphics[width=.115\linewidth]{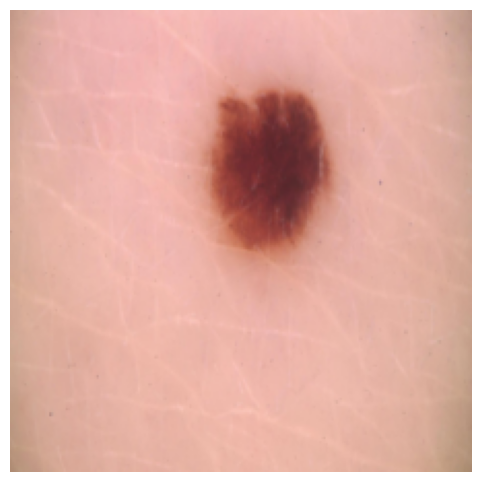}
\end{tabular} & \rotatebox[origin=c]{90}{\textbf{LRP}} & 
\begin{tabular}{c}
     No Attention\\
     \includegraphics[width=.115\linewidth]{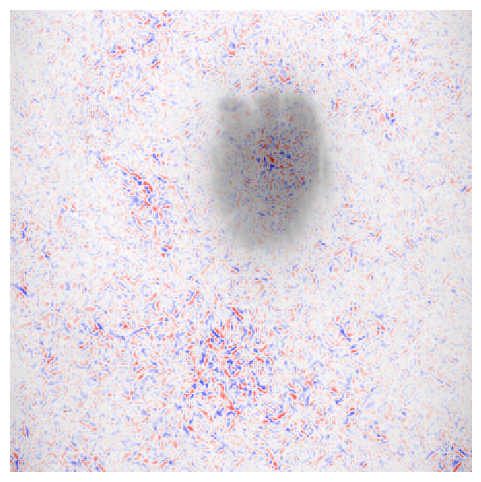}
\end{tabular} & 
\begin{tabular}{c}
     $+$ SE Layer\\
     \includegraphics[width=.115\linewidth]{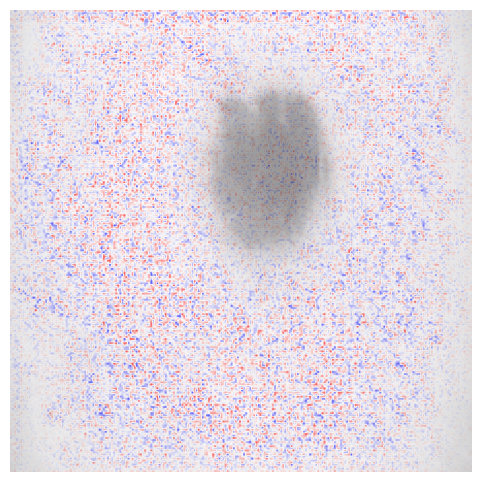}
\end{tabular} &
\begin{tabular}{c}
    $+$ CBAM Layer\\
     \includegraphics[width=.115\linewidth]{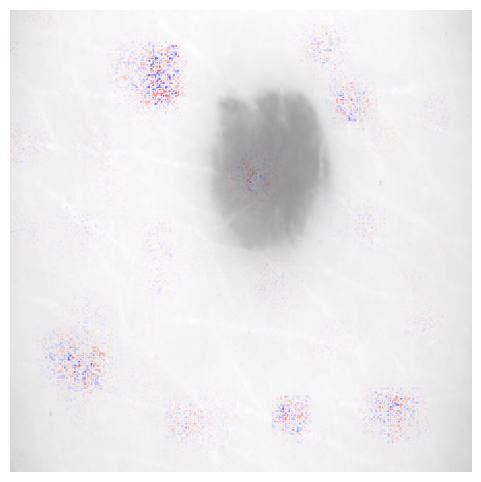}
\end{tabular} & &
\begin{tabular}{c}
     No Attention\\
     \includegraphics[width=.115\linewidth]{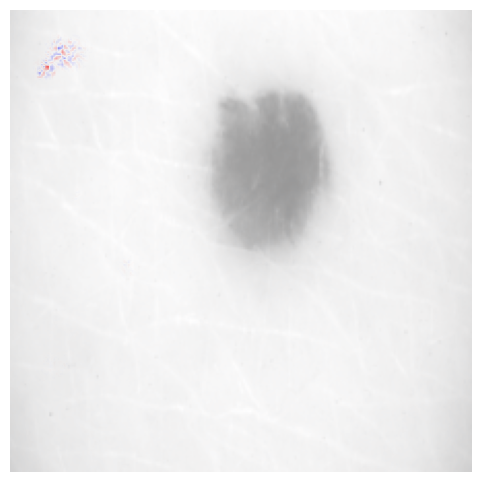}
\end{tabular} &
\begin{tabular}{c}
     $+$ SE Layer\\
     \includegraphics[width=.115\linewidth]{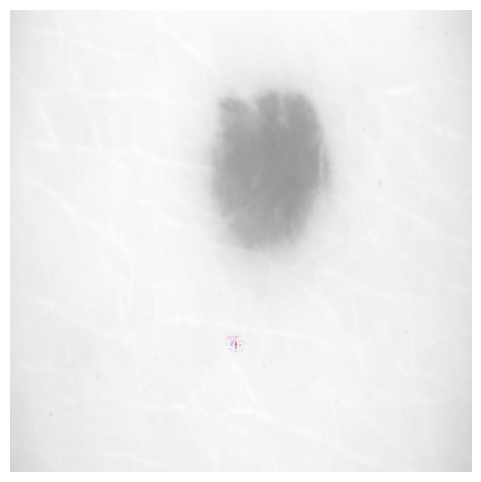}
\end{tabular} &
\begin{tabular}{c}
     $+$ CBAM Layer\\
     \includegraphics[width=.115\linewidth]{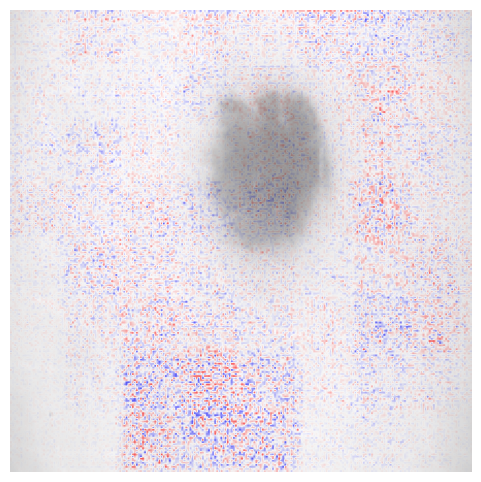}
\end{tabular}
\\
\\
\begin{tabular}{c}
    DeiT (LRP)\\
    \includegraphics[width=.115\linewidth]{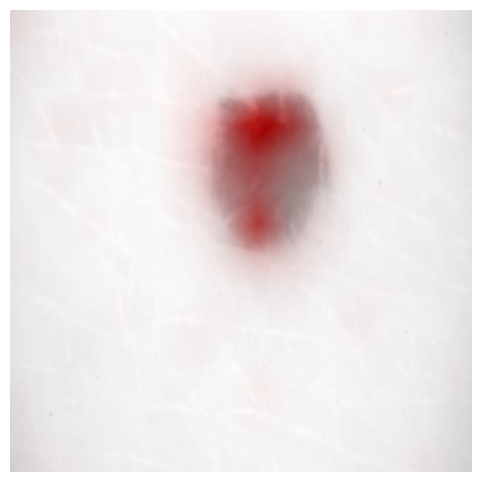}
\end{tabular} & \rotatebox[origin=c]{90}{\textbf{DeepLIFT}} &

\begin{tabular}{c}
    No Attention\\
     \includegraphics[width=.115\linewidth]{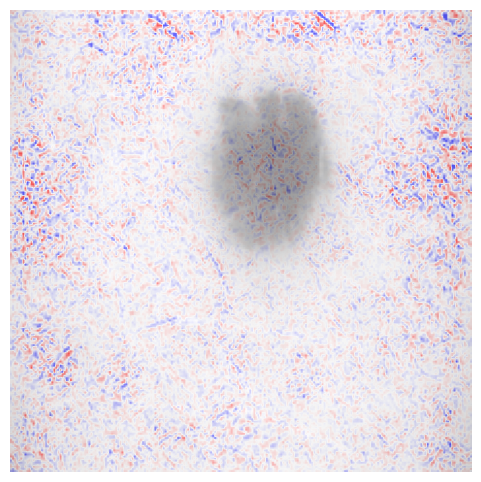}
\end{tabular} &
\begin{tabular}{c}
     $+$ SE Layer\\
     \includegraphics[width=.115\linewidth]{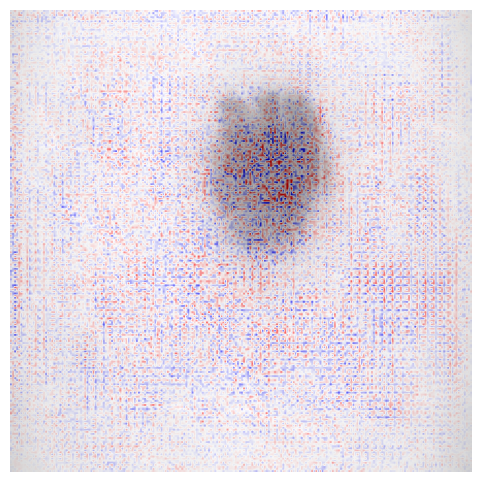}
\end{tabular} &
\begin{tabular}{c}
     $+$ CBAM Layer\\
     \includegraphics[width=.115\linewidth]{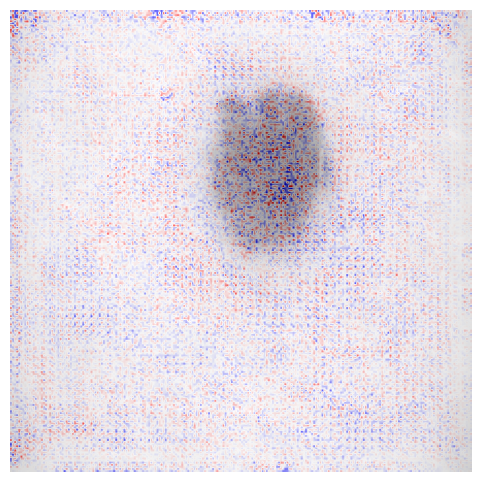}
\end{tabular} & &
\begin{tabular}{c}
     No Attention\\
     \includegraphics[width=.115\linewidth]{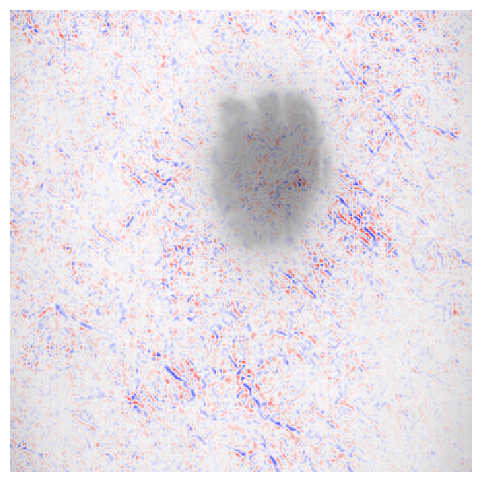}
\end{tabular} &
\begin{tabular}{c}
     $+$ SE Layer\\
     \includegraphics[width=.115\linewidth]{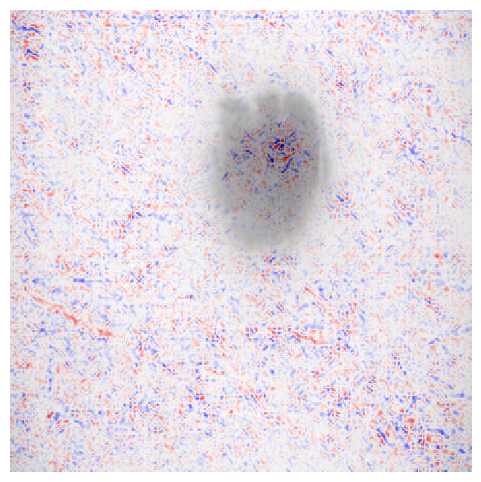}
\end{tabular} &
\begin{tabular}{c}
     $+$ CBAM Layer\\
     \includegraphics[width=.115\linewidth]{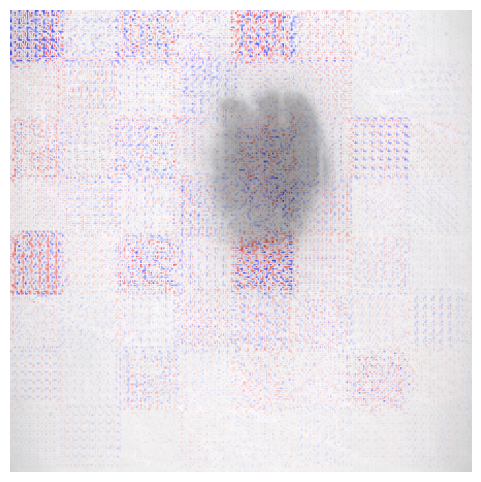}
\end{tabular}

\end{tabular}
\end{table*}

\setlength{\tabcolsep}{3pt}
\begin{table*}[ht]
\sffamily
\centering
\caption{Example of LRP and DeepLIFT \textit{post-hoc} saliency maps for an image of the MIMIC-CXR data set with the label $0$ correctly classified as $0$ by all models.}
\label{tab:mimiccxr-gt0_pred0}
\begin{tabular}{c@{\hspace{.8\tabcolsep}}c@{\hspace{.3\tabcolsep}}cccc@{\hspace{1.5\tabcolsep}}ccc}
& & \multicolumn{3}{c}{\textbf{DENSENET}} & & \multicolumn{3}{c}{\textbf{RESNET}} \\
\cmidrule{3-5} \cmidrule{7-9} \\
\begin{tabular}{c}
     Original Image\\
     \includegraphics[width=.115\linewidth]{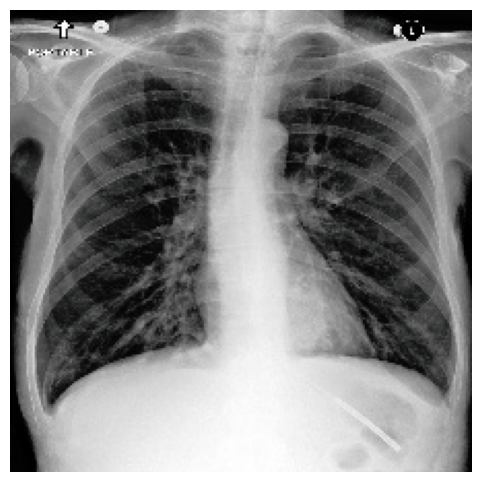}
\end{tabular} & \rotatebox[origin=c]{90}{\textbf{LRP}} & 
\begin{tabular}{c}
     No Attention\\
     \includegraphics[width=.115\linewidth]{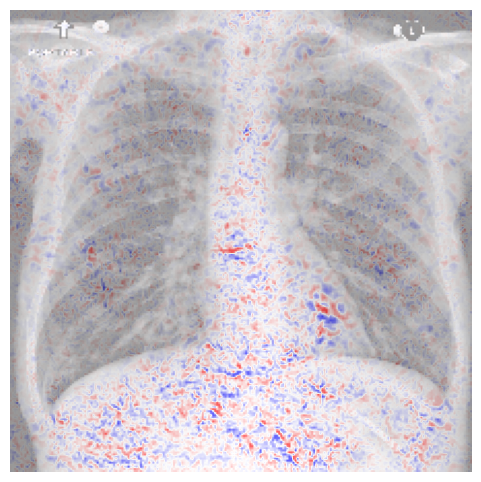}
\end{tabular} & 
\begin{tabular}{c}
     $+$ SE Layer\\
     \includegraphics[width=.115\linewidth]{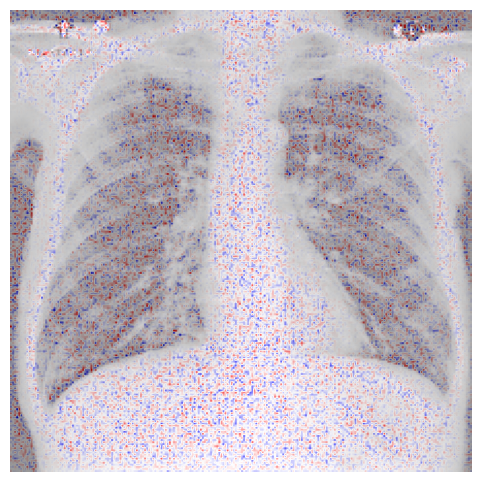}
\end{tabular} &
\begin{tabular}{c}
    $+$ CBAM Layer\\
     \includegraphics[width=.115\linewidth]{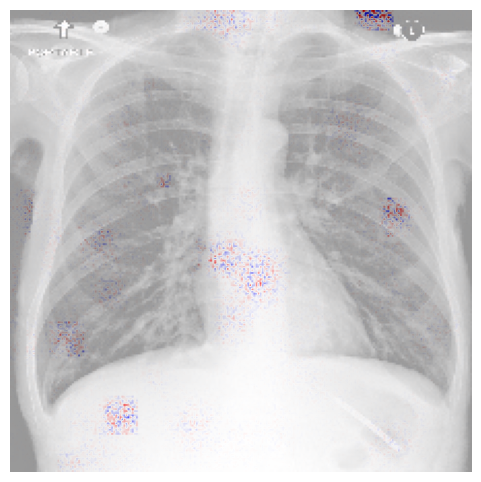}
\end{tabular} & &
\begin{tabular}{c}
     No Attention\\
     \includegraphics[width=.115\linewidth]{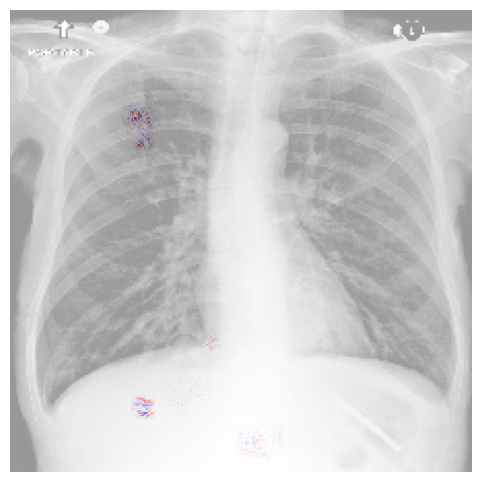}
\end{tabular} &
\begin{tabular}{c}
     $+$ SE Layer\\
     \includegraphics[width=.115\linewidth]{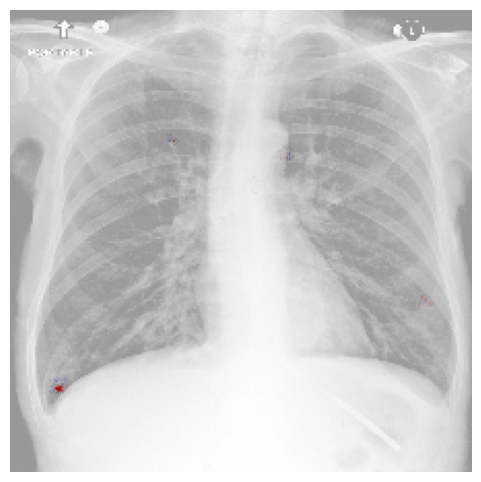}
\end{tabular} &
\begin{tabular}{c}
     $+$ CBAM Layer\\
     \includegraphics[width=.115\linewidth]{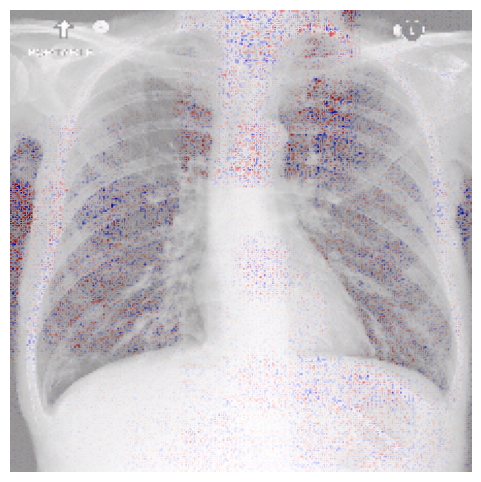}
\end{tabular}
\\
\\
\begin{tabular}{c}
    DeiT (LRP)\\
    \includegraphics[width=.115\linewidth]{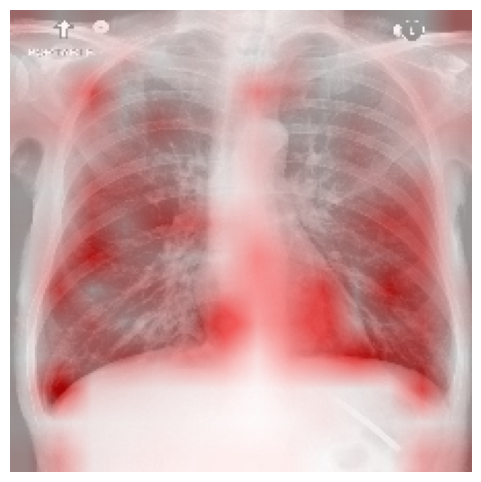}
\end{tabular} & \rotatebox[origin=c]{90}{\textbf{DeepLIFT}} &

\begin{tabular}{c}
    No Attention\\
     \includegraphics[width=.115\linewidth]{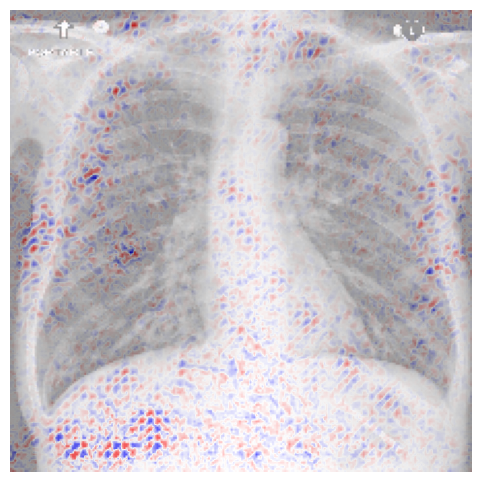}
\end{tabular} &
\begin{tabular}{c}
     $+$ SE Layer\\
     \includegraphics[width=.115\linewidth]{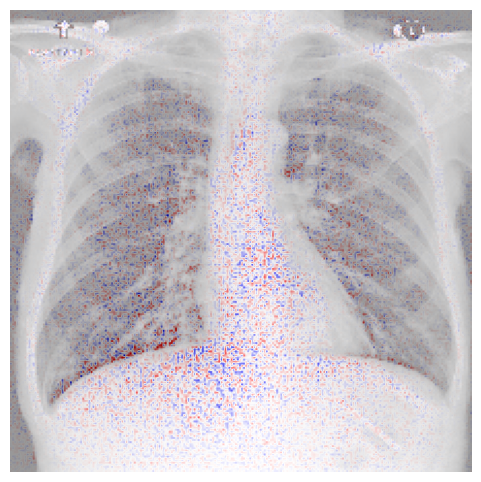}
\end{tabular} &
\begin{tabular}{c}
     $+$ CBAM Layer\\
     \includegraphics[width=.115\linewidth]{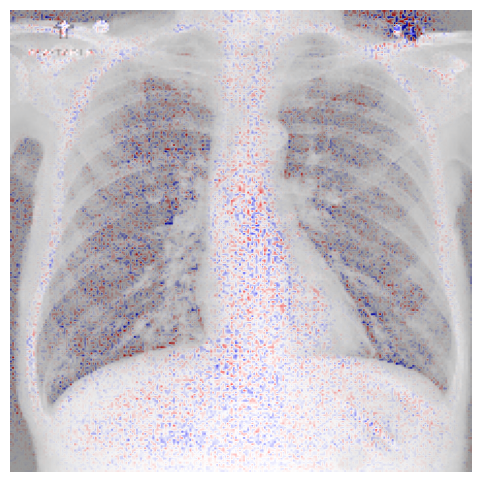}
\end{tabular} & &
\begin{tabular}{c}
     No Attention\\
     \includegraphics[width=.115\linewidth]{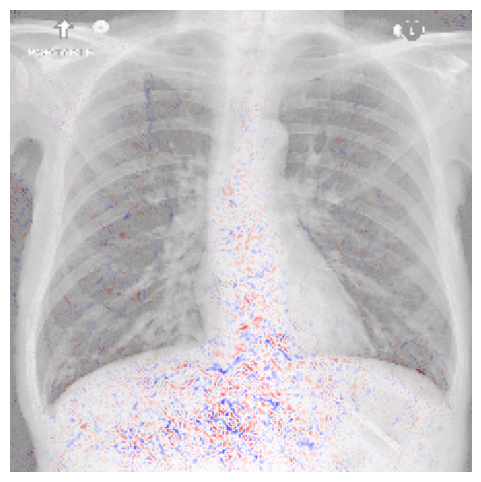}
\end{tabular} &
\begin{tabular}{c}
     $+$ SE Layer\\
     \includegraphics[width=.115\linewidth]{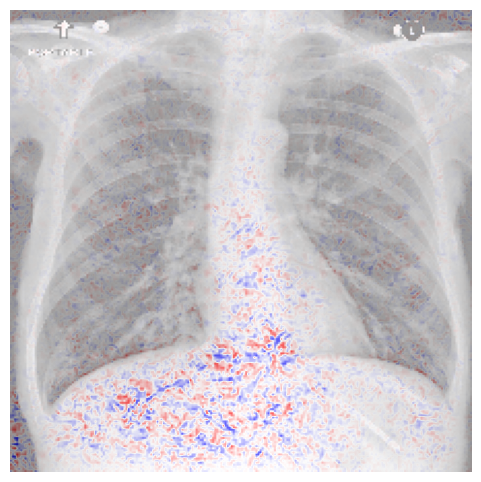}
\end{tabular} &
\begin{tabular}{c}
     $+$ CBAM Layer\\
     \includegraphics[width=.115\linewidth]{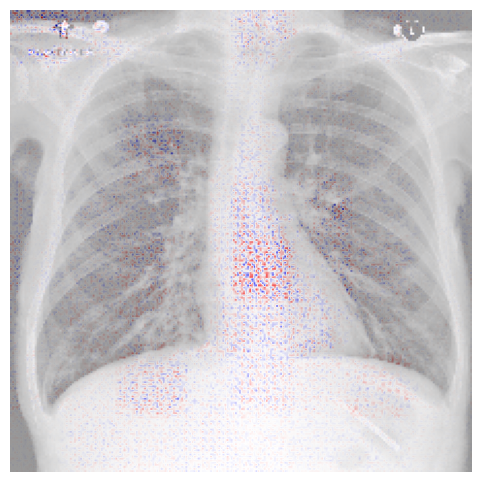}
\end{tabular}

\end{tabular}
\end{table*}

\setlength{\tabcolsep}{3pt}
\begin{table*}[ht]
\sffamily
\centering
\caption{Example of LRP and DeepLIFT \textit{post-hoc} saliency maps for an image of the MIMIC-CXR data set with the label $1$ correctly classified as $1$ by all models.}
\label{tab:mimiccxr-gt1_pred1}
\begin{tabular}{c@{\hspace{.8\tabcolsep}}c@{\hspace{.3\tabcolsep}}cccc@{\hspace{1.5\tabcolsep}}ccc}
& & \multicolumn{3}{c}{\textbf{DENSENET}} & & \multicolumn{3}{c}{\textbf{RESNET}} \\
\cmidrule{3-5} \cmidrule{7-9} \\
\begin{tabular}{c}
     Original Image\\
     \includegraphics[width=.115\linewidth]{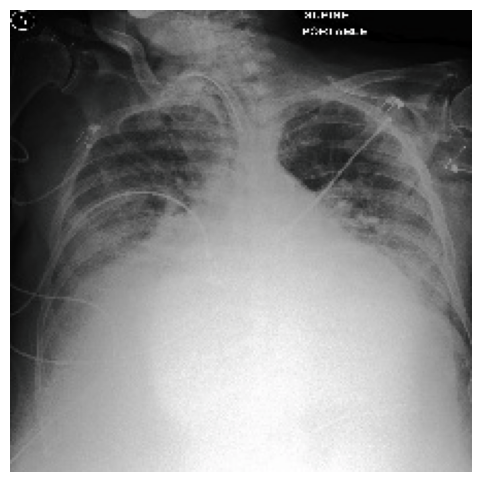}
\end{tabular} & \rotatebox[origin=c]{90}{\textbf{LRP}} & 
\begin{tabular}{c}
     No Attention\\
     \includegraphics[width=.115\linewidth]{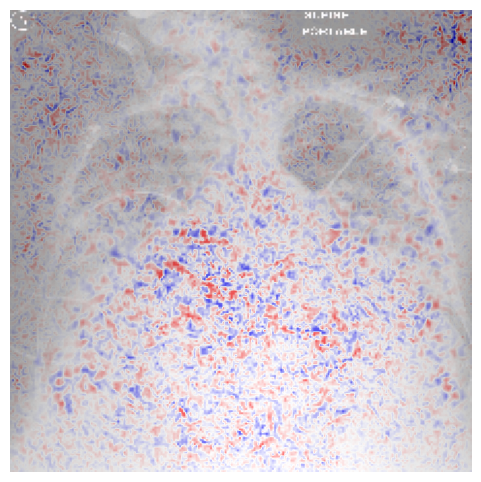}
\end{tabular} & 
\begin{tabular}{c}
     $+$ SE Layer\\
     \includegraphics[width=.115\linewidth]{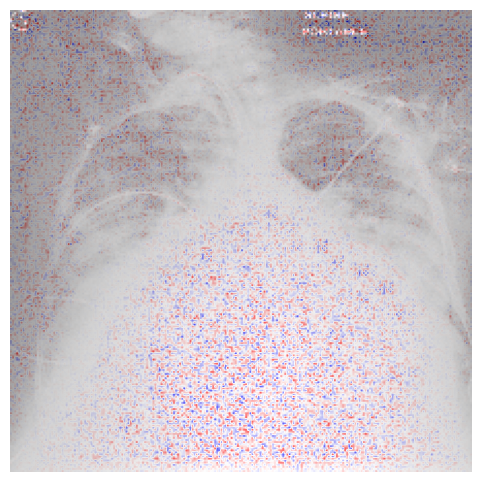}
\end{tabular} &
\begin{tabular}{c}
    $+$ CBAM Layer\\
     \includegraphics[width=.115\linewidth]{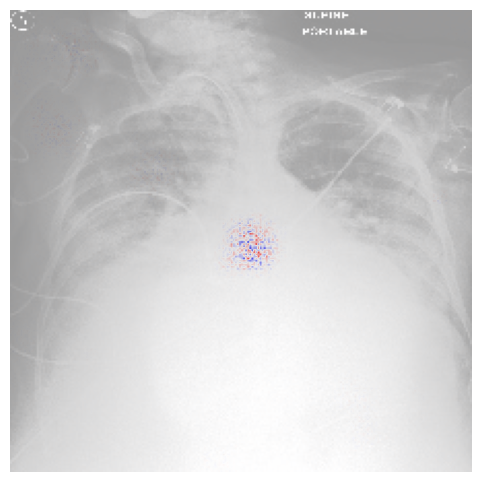}
\end{tabular} & &
\begin{tabular}{c}
     No Attention\\
     \includegraphics[width=.115\linewidth]{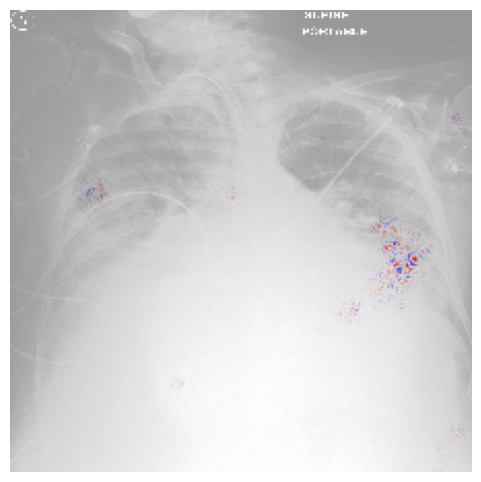}
\end{tabular} &
\begin{tabular}{c}
     $+$ SE Layer\\
     \includegraphics[width=.115\linewidth]{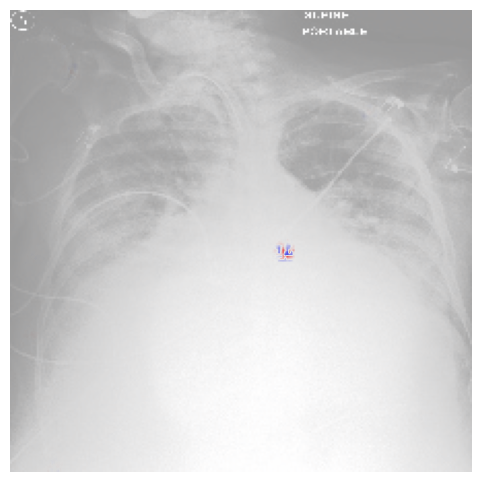}
\end{tabular} &
\begin{tabular}{c}
     $+$ CBAM Layer\\
     \includegraphics[width=.115\linewidth]{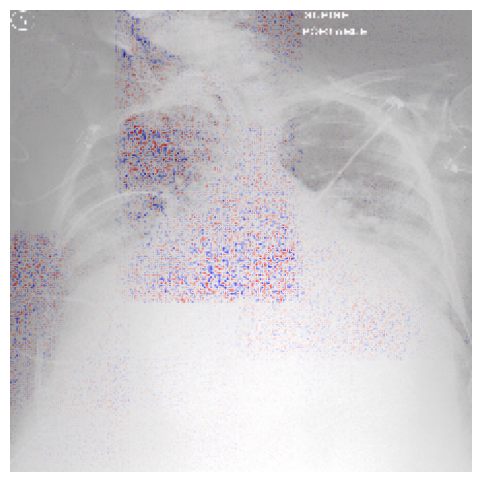}
\end{tabular}
\\
\\
\begin{tabular}{c}
    DeiT (LRP)\\
    \includegraphics[width=.115\linewidth]{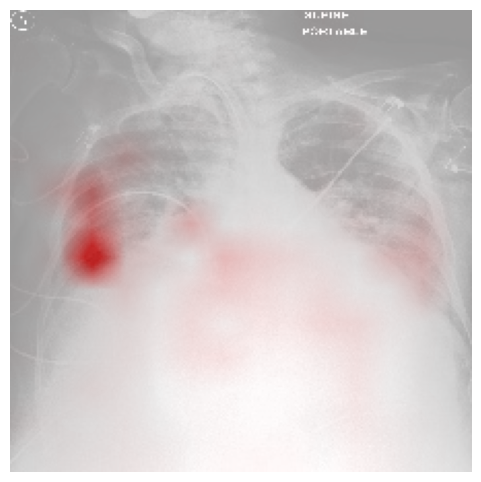}
\end{tabular} & \rotatebox[origin=c]{90}{\textbf{DeepLIFT}} &

\begin{tabular}{c}
    No Attention\\
     \includegraphics[width=.115\linewidth]{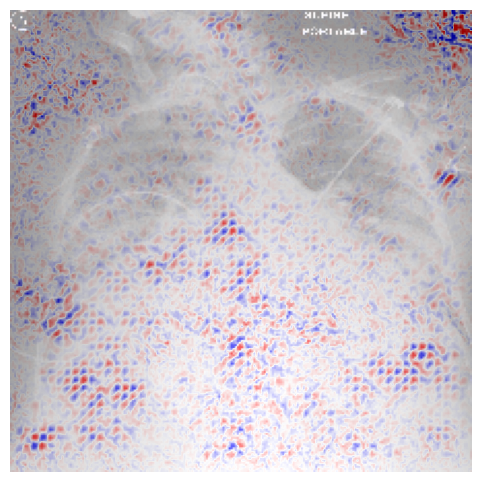}
\end{tabular} &
\begin{tabular}{c}
     $+$ SE Layer\\
     \includegraphics[width=.115\linewidth]{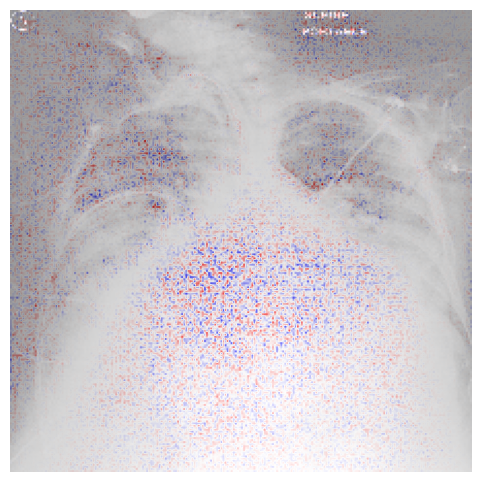}
\end{tabular} &
\begin{tabular}{c}
     $+$ CBAM Layer\\
     \includegraphics[width=.115\linewidth]{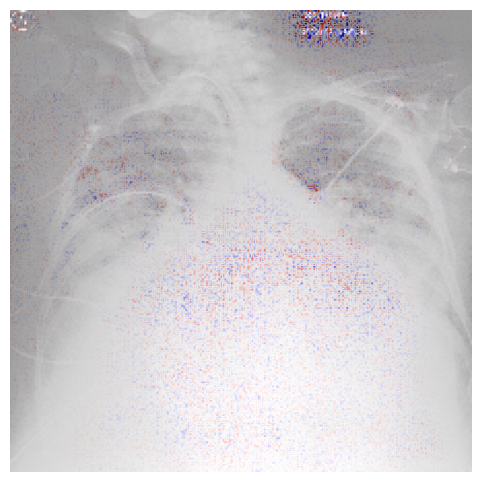}
\end{tabular} & &
\begin{tabular}{c}
     No Attention\\
     \includegraphics[width=.115\linewidth]{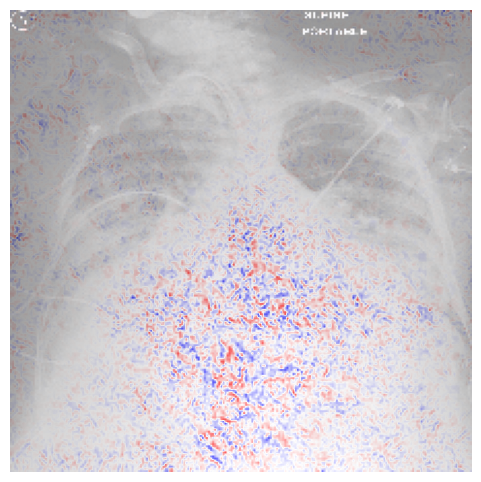}
\end{tabular} &
\begin{tabular}{c}
     $+$ SE Layer\\
     \includegraphics[width=.115\linewidth]{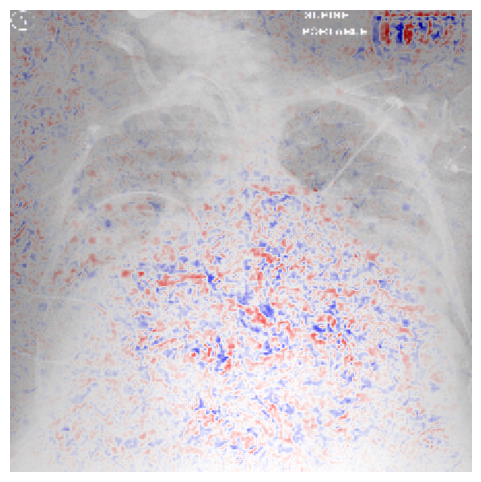}
\end{tabular} &
\begin{tabular}{c}
     $+$ CBAM Layer\\
     \includegraphics[width=.115\linewidth]{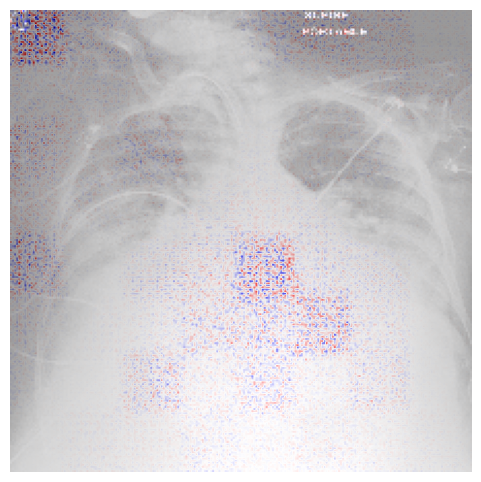}
\end{tabular}

\end{tabular}
\end{table*}

\setlength{\tabcolsep}{3pt}
\begin{table*}[ht]
\sffamily
\centering
\caption{Example of LRP and DeepLIFT \textit{post-hoc} saliency maps for an image of the MIMIC-CXR data set with the label $0$ incorrectly classified as $1$ by all models.}
\label{tab:mimiccxr-gt0_pred1}
\begin{tabular}{c@{\hspace{.8\tabcolsep}}c@{\hspace{.3\tabcolsep}}cccc@{\hspace{1.5\tabcolsep}}ccc}
& & \multicolumn{3}{c}{\textbf{DENSENET}} & & \multicolumn{3}{c}{\textbf{RESNET}} \\
\cmidrule{3-5} \cmidrule{7-9} \\
\begin{tabular}{c}
     Original Image\\
     \includegraphics[width=.115\linewidth]{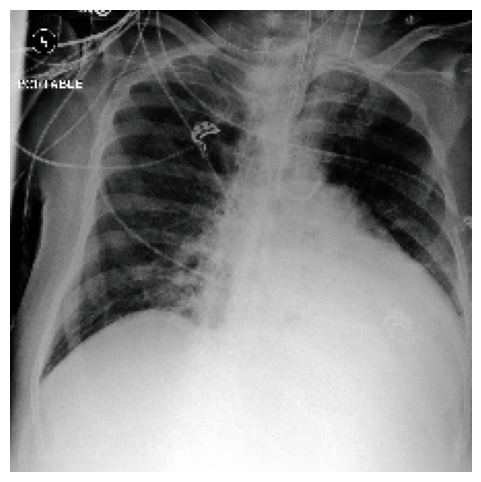}
\end{tabular} & \rotatebox[origin=c]{90}{\textbf{LRP}} & 
\begin{tabular}{c}
     No Attention\\
     \includegraphics[width=.115\linewidth]{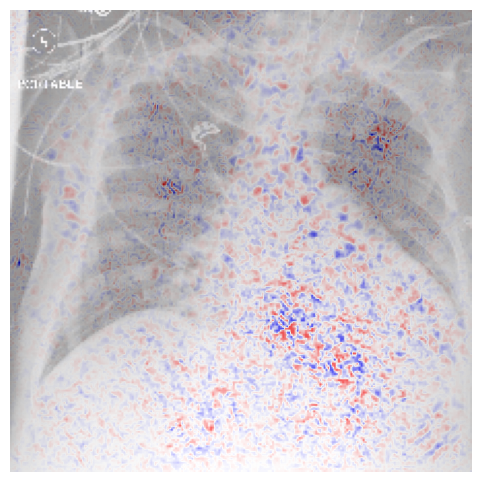}
\end{tabular} & 
\begin{tabular}{c}
     $+$ SE Layer\\
     \includegraphics[width=.115\linewidth]{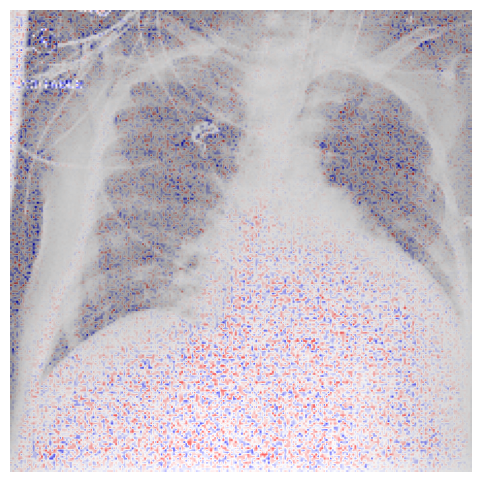}
\end{tabular} &
\begin{tabular}{c}
    $+$ CBAM Layer\\
     \includegraphics[width=.115\linewidth]{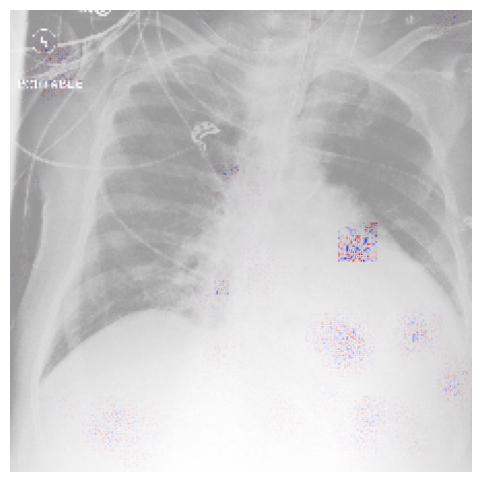}
\end{tabular} & &
\begin{tabular}{c}
     No Attention\\
     \includegraphics[width=.115\linewidth]{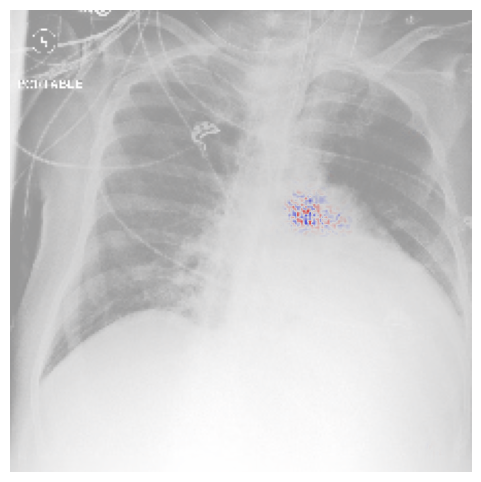}
\end{tabular} &
\begin{tabular}{c}
     $+$ SE Layer\\
     \includegraphics[width=.115\linewidth]{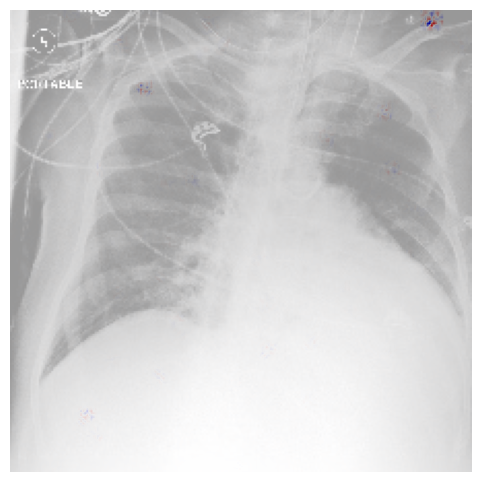}
\end{tabular} &
\begin{tabular}{c}
     $+$ CBAM Layer\\
     \includegraphics[width=.115\linewidth]{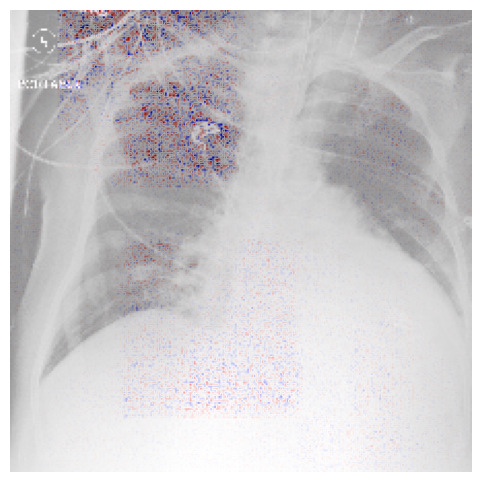}
\end{tabular}
\\
\\
\begin{tabular}{c}
    DeiT (LRP)\\
    \includegraphics[width=.115\linewidth]{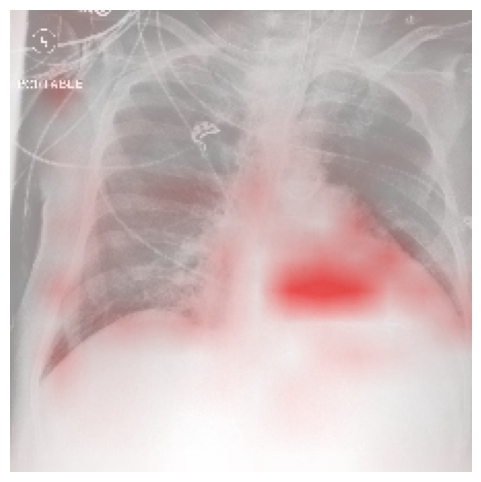}
\end{tabular} & \rotatebox[origin=c]{90}{\textbf{DeepLIFT}} &

\begin{tabular}{c}
    No Attention\\
     \includegraphics[width=.115\linewidth]{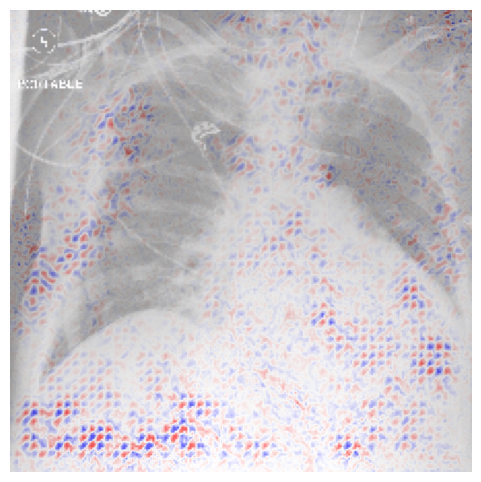}
\end{tabular} &
\begin{tabular}{c}
     $+$ SE Layer\\
     \includegraphics[width=.115\linewidth]{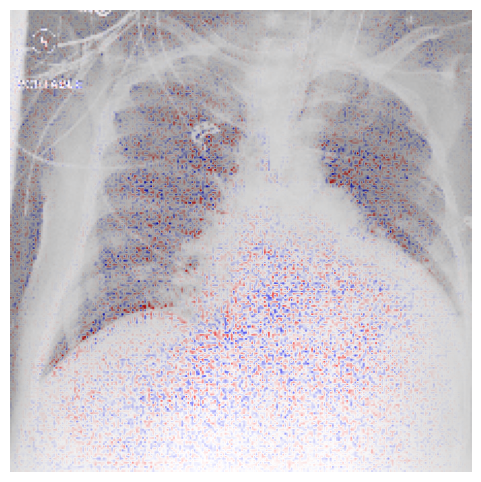}
\end{tabular} &
\begin{tabular}{c}
     $+$ CBAM Layer\\
     \includegraphics[width=.115\linewidth]{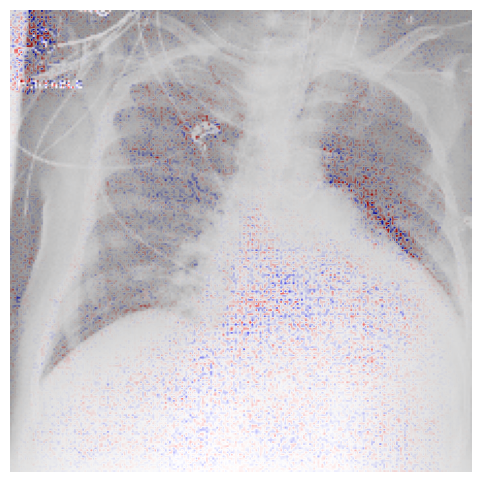}
\end{tabular} & &
\begin{tabular}{c}
     No Attention\\
     \includegraphics[width=.115\linewidth]{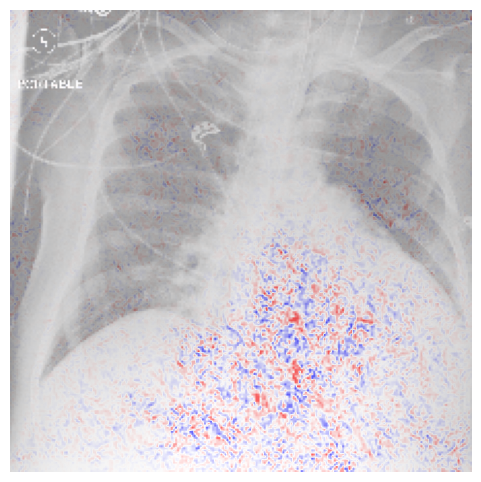}
\end{tabular} &
\begin{tabular}{c}
     $+$ SE Layer\\
     \includegraphics[width=.115\linewidth]{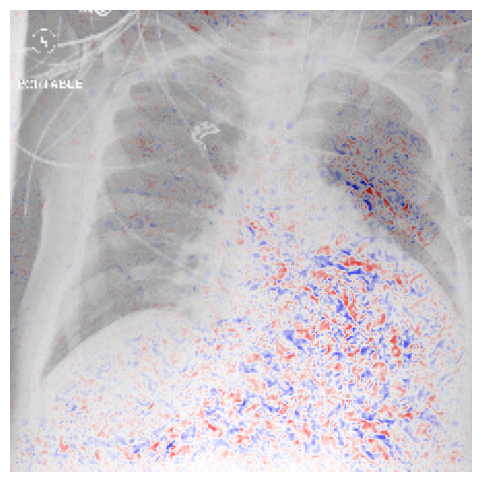}
\end{tabular} &
\begin{tabular}{c}
     $+$ CBAM Layer\\
     \includegraphics[width=.115\linewidth]{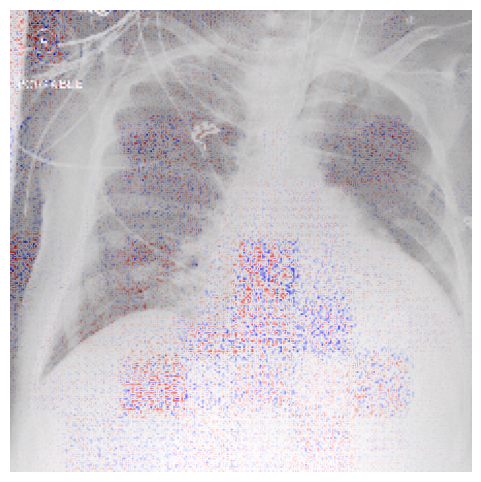}
\end{tabular}

\end{tabular}
\end{table*}

\setlength{\tabcolsep}{3pt}
\begin{table*}[ht]
\sffamily
\centering
\caption{Example of LRP and DeepLIFT \textit{post-hoc} saliency maps for an image of the MIMIC-CXR data set with the label $1$ incorrectly classified as $0$ by all models.}
\label{tab:mimiccxr-gt1_pred0}
\begin{tabular}{c@{\hspace{.8\tabcolsep}}c@{\hspace{.3\tabcolsep}}cccc@{\hspace{1.5\tabcolsep}}ccc}
& & \multicolumn{3}{c}{\textbf{DENSENET}} & & \multicolumn{3}{c}{\textbf{RESNET}} \\
\cmidrule{3-5} \cmidrule{7-9} \\
\begin{tabular}{c}
     Original Image\\
     \includegraphics[width=.115\linewidth]{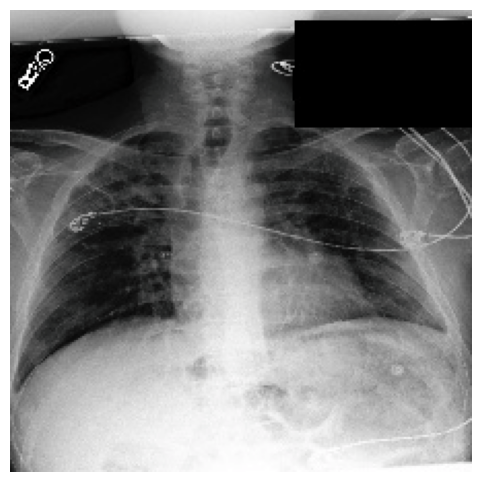}
\end{tabular} & \rotatebox[origin=c]{90}{\textbf{LRP}} & 
\begin{tabular}{c}
     No Attention\\
     \includegraphics[width=.115\linewidth]{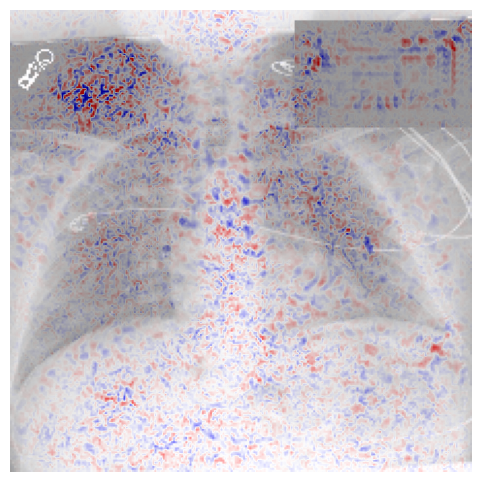}
\end{tabular} & 
\begin{tabular}{c}
     $+$ SE Layer\\
     \includegraphics[width=.115\linewidth]{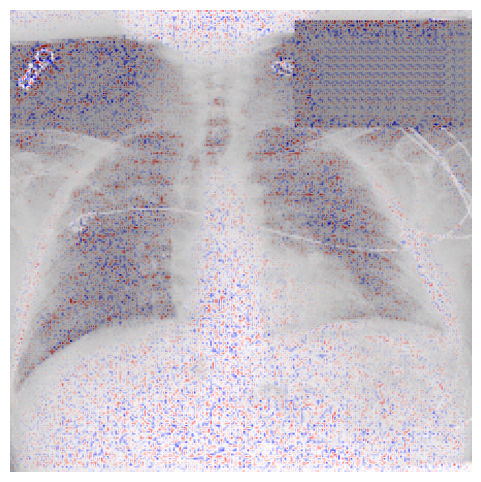}
\end{tabular} &
\begin{tabular}{c}
    $+$ CBAM Layer\\
     \includegraphics[width=.115\linewidth]{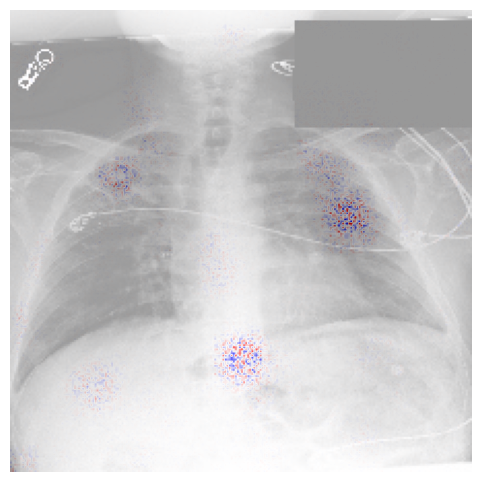}
\end{tabular} & &
\begin{tabular}{c}
     No Attention\\
     \includegraphics[width=.115\linewidth]{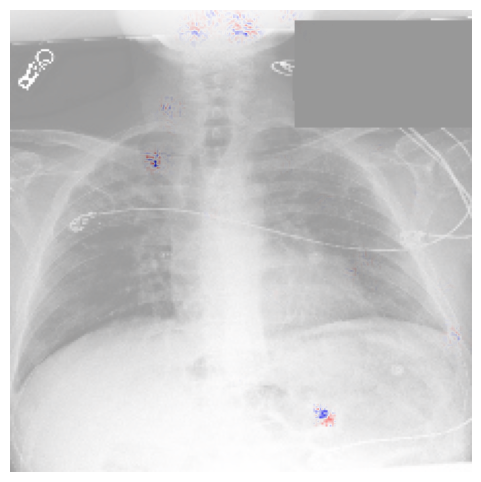}
\end{tabular} &
\begin{tabular}{c}
     $+$ SE Layer\\
     \includegraphics[width=.115\linewidth]{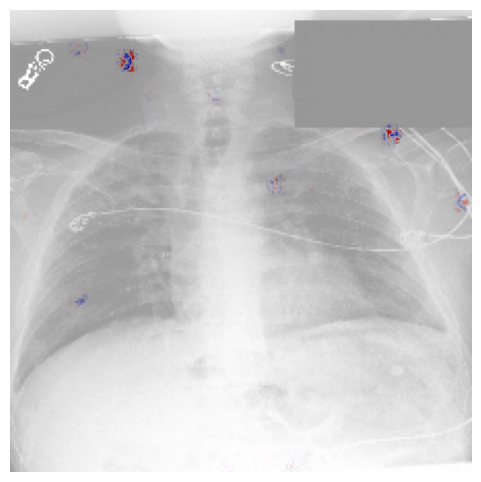}
\end{tabular} &
\begin{tabular}{c}
     $+$ CBAM Layer\\
     \includegraphics[width=.115\linewidth]{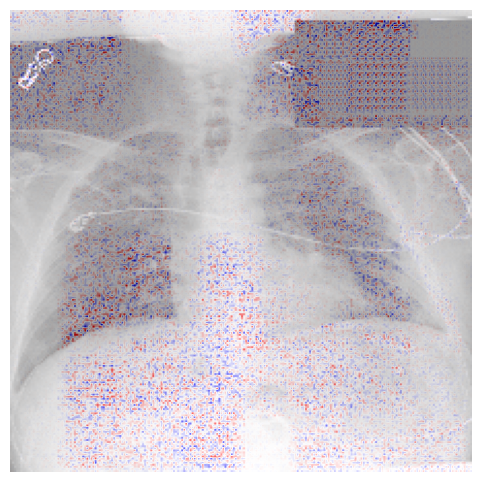}
\end{tabular}
\\
\\
\begin{tabular}{c}
    DeiT (LRP)\\
    \includegraphics[width=.115\linewidth]{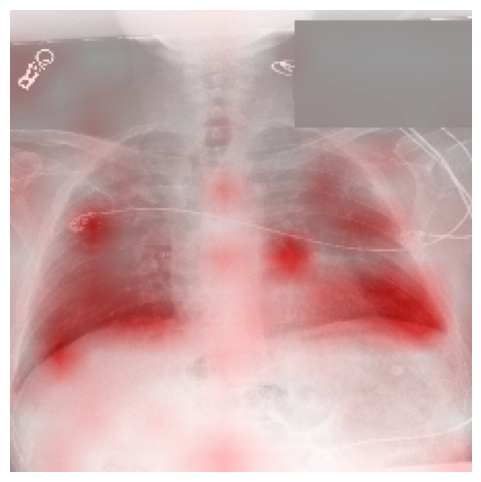}
\end{tabular} & \rotatebox[origin=c]{90}{\textbf{DeepLIFT}} &

\begin{tabular}{c}
    No Attention\\
     \includegraphics[width=.115\linewidth]{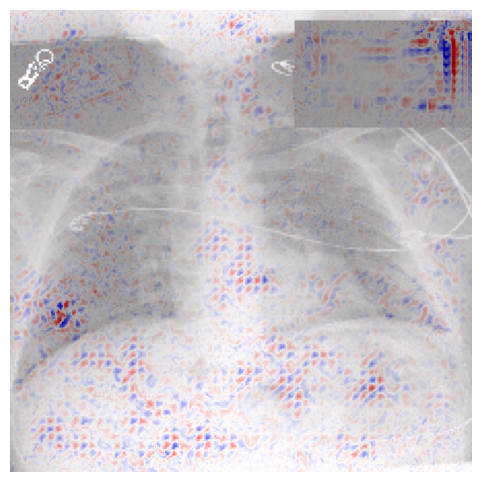}
\end{tabular} &
\begin{tabular}{c}
     $+$ SE Layer\\
     \includegraphics[width=.115\linewidth]{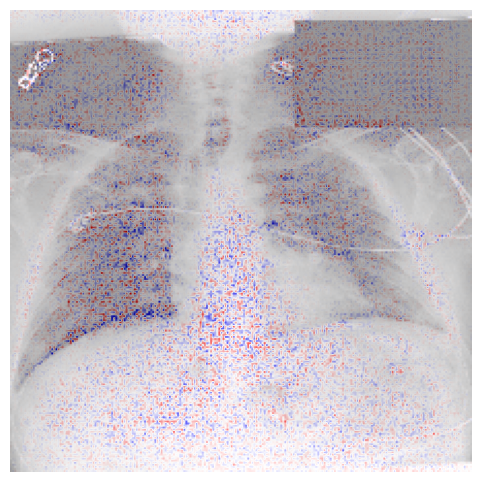}
\end{tabular} &
\begin{tabular}{c}
     $+$ CBAM Layer\\
     \includegraphics[width=.115\linewidth]{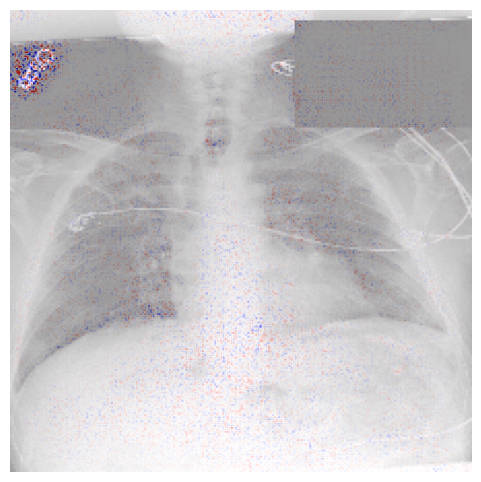}
\end{tabular} & &
\begin{tabular}{c}
     No Attention\\
     \includegraphics[width=.115\linewidth]{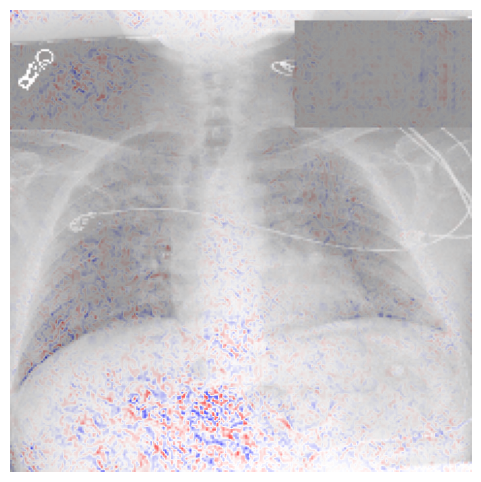}
\end{tabular} &
\begin{tabular}{c}
     $+$ SE Layer\\
     \includegraphics[width=.115\linewidth]{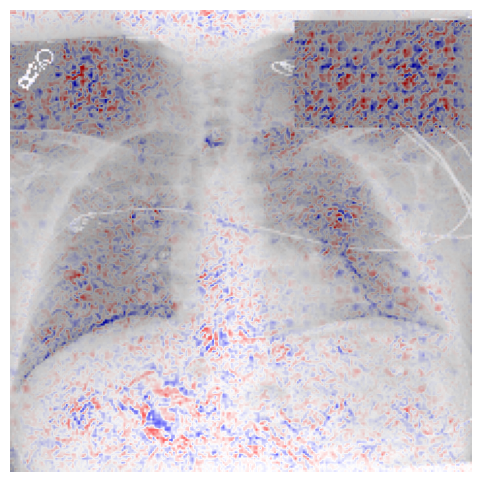}
\end{tabular} &
\begin{tabular}{c}
     $+$ CBAM Layer\\
     \includegraphics[width=.115\linewidth]{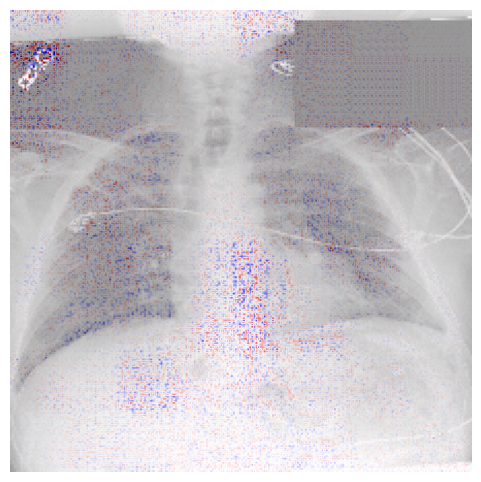}
\end{tabular}

\end{tabular}
\end{table*}

\subsection*{Discussion}
\subsubsection*{Predictive Performance}
All the experiments related to the predictive performance of deep learning models on the different data sets suggest that it is not clear that one should expect improvements in their accuracy when using attention mechanisms. Several attention-based architectures tend to lower their predictive performances against their baseline architectures. Besides, even when directly comparing a Transformer-based architecture (i.e., DeiT), the same baseline architectures seem to achieve comparable results. Given that the intuition behind attention mechanisms is that these end up learning the most relevant features, one might expect that attention-based architectures would perform better when trained in low data regimes. However, results obtained in all data sets suggest that this might not be the case. In fact, we can observe that, apart from the Transformer-based architecture, the baseline backbone models often perform better. Hence, it is reasonable to conclude that the use of attention mechanisms will not \textit{always} bring benefits to the training of deep learning algorithms, an argument that contrasts with a trending narrative in most recent papers. On the other hand, the results reported in the literature often relate to marginal or residual improvements in the state-of-the-art backbone networks. Given that the training of deep learning algorithms is generally a stochastic process, there is a need to assess these reported improvements with a more critical view and with robust statistical tests.

\subsubsection*{Model Complexity}
The integration of attention mechanisms increases the number of parameters of the deep learning models, thus increasing their complexity. This information allows us to conclude that, at least for computer vision applications, it is not necessarily true that the use of attention mechanisms contributes to the decrease of model complexity. While this may be true in several natural language processing applications, it is not correct to generalise such a rule to other types of applications. On the other hand, since these attention mechanisms often rely on simple operations (e.g., matrix multiplications), such as convolutions, we acknowledge that their use may reduce the training time of deep learning algorithms (similarly to what happened when the community started using CNNs). This aspect, though, was not objectively assessed in our experiments. However, since they are particularly efficient at modelling long-range dependencies (e.g., Transformer-based architectures), we expect these mechanisms to keep increasing their popularity and, hence, become widely used in medical image applications. Besides, Cordonnier et al.~\cite{Cordonnier2020On} reported that these mechanisms may operate similarly to CNNs. Another question arises from these results: are these attention-based algorithms allegedly performing well because of the inner-functioning of their attention mechanisms themselves or just because we are increasing the number of model parameters? While one may report that this issue is nonsense, we point out that some Transformer-based architectures have a considerably high number of parameters~\cite{touvron2021training}.

\subsubsection*{Post-model Interpretability}
Although there are already several works that criticise the subjectivity of \textit{post-hoc} saliency maps~\cite{lipton2018mythos,adebayo2018sanity,kindermans2019reliability,hooker2019benchmark,rudin2019stop,kaur2020interpreting}, it is still common to use these methods to justify improvements related to the degree of interpretability of models. In this sense, when analysing the results obtained with the DenseNet-121 and ResNet-50 models, we would expect that, with the increase in complexity of the attention mechanism, the distribution of the important pixels around the image would also be more focused (i.e., would have less variance). However, that does not seem to happen in our use cases. Interestingly, besides the inherent properties of the data and the task (i.e., images of the retina are different from skin images and chest X-ray images), the results drastically change when we use a different backbone. In general, ResNet-based models attain less noisy DeepLIFT and LRP saliency maps, with some exceptions in the ISIC2020 data set (e.g., Table~\ref{tab:isic-gt1_pred0}). On the other hand, the LRP saliency maps obtained for the DeiT seem to highlight somewhat clear regions of the images. Besides, it is also important to remind the reader that these frameworks allow us to generate explanations even for the cases where the model miss-classifies. We also stress that one of the limitations of such analysis is that there does not exist an objective ground truth of what a high-quality visual explanation is. Therefore, it is not trivial, in computer vision tasks such as this, to conclude with complete confidence that, even for the cases where the model succeeds, it learned the right correlations. Hence, can we believe the narrative that attention mechanisms are learning the most relevant features of the image? Or is this type of analysis a result of luck in most cases? Truth is, our results still suggest that there is a high degree of subjectivity (i.e., the interpretation of these saliency maps deeply depends on the human that is interpreting them) and that there is no apparent correlation between the use of attention mechanisms and the visual aspect of the \textit{attributions}. Besides, we stress that these visualisations shown in the literature depend on several parameters (e.g., type of colour map, overlay parameter). Moreover, the method we use to generate such visualisations may often change what we expect users to observe. This is also related to another open challenge in \textit{post-hoc} explanation methods that is the sign of the attributions. What should we expect? Positive attributions for the positive class and negative attributions for the negative class, or the other way around? Should we always normalise the sign of the attributions so it is always positive? This motivates our statement on the need for more objective methods to assess the degree of interpretability of models.

\section*{Conclusions and Future Challenges}\label{sec:conclusion}
This section summarises the main conclusions of this survey and points to future directions toward the study of attention-based algorithms for medical applications. 

\subsection*{Conclusions}
This survey presented an exhaustive overview of the use of attention mechanisms for medical applications and provided an experimental study on medical image classification that approached three different use cases. We found that backbone models can attain equivalent predictive performances to Transformer-based architectures with equivalent model complexity (i.e., number of parameters). Moreover, when using a \textit{post-hoc} framework to visually assess what type of features these models can extract, we can conclude that there is still a high degree of subjectivity in such analysis. The results are very noisy, even for the cases of attention mechanisms, which is counter-intuitive (i.e., with attention mechanisms, these saliency maps should have less noise). The community is moving toward using attention mechanisms (specially Transformer-based ones) and arguing that these frameworks increase the quality, transparency, and interpretability of deep learning architectures. However, we state that this is not true and that there are several open challenges one needs to address to achieve this.

\subsection*{Future Challenges}
\subsubsection*{Attention Mechanisms: Past or Future?}
The research on attention-based models is in constant evolution and requires a validation and maturation step. Besides, as these models keep rising in popularity and keep convincing the community of their benefits, we acknowledge that they will appear as a common block in future architectures. However, the scientific community must keep in mind that high-stake decision areas (e.g., finance, healthcare, justice) require algorithms to be fair and transparent~\cite{rudin2019stop}. Therefore, even if attention mechanisms are pushing deep learning algorithms towards the limits of their predictive power, we must start thinking about creating interpretable frameworks that allow us to audit and assess these algorithms concerning the specific conditions of their domains (i.e., in the case of this work, healthcare).

\subsubsection*{Design and Integration of Attention Mechanisms}
A potential disadvantage of the investment in attention mechanisms is related to the design of novel strategies of attention and their integration into well-studied backbone models. Several works on attention often rely on a specific backbone architecture for their experiments. Besides, if we look at the topographies of these deep learning algorithms, it is not always clear for the users where they should place these modules, and why it makes sense to put them in a specific place. Once again, another question arises: are these attention modules dependent on the backbone into which they are integrated to? Can we consider that the reported improvements related to the predictive performance of deep learning models are solely due to the addition of an attention module? Can we come up with an objective strategy that guides future users in building and integrating attention mechanisms into their models?

\subsubsection*{The Rise of Transformers}
Apart from the \textit{traditional} deep algorithms with attention mechanisms, it is crucial to talk about Transformer-based architectures. The attentive reader might have noticed that most of the recent approaches to the use of attention-based models in medical applications rely on Transformer-based modules. While there is hype on the use of these structures, it is not clear whether they are more interpretable or not, or if their generalization power is superior to the other deep models.

\subsubsection*{Interpretability is the Path to Better Algorithms}
Even if we acknowledge that the Transformer-based algorithms are provoking a shift in the paradigm, it is not clear that they are improving the transparency of algorithms. For instance, in~\cite{touvron2021training}, the authors show that when training a DeiT using \textit{knowledge distillation} techniques, the Transformer ends up learning better when the \textit{teacher model} is a CNN. Hence, if the Transformer is learning with a non-interpretable model by design, how can we trust that the Transformer is inherently interpretable? On the other hand, given the capacity of these attention-based models to learn long-range dependencies, we must create objective techniques to assess the quality of the features learned by these frameworks, while moving towards the design of intrinsically interpretable attention-based architectures. Even if we intend to keep using visual saliency maps to explain our models, in a high stake decision field such as healthcare, we must achieve a clear standard, validated by the clinical community, of what these maps should look like and what is their effective meaning.

\bibliographystyle{unsrtnat}
\bibliography{template}

\section*{Acknowledgements}
This work, developed within the scope of the project ``TAMI - Transparent Artificial Medical Intelligence'' (NORTE-01-0247-FEDER-045905), is co-financed by ERDF - European Regional Fund through the Operational Program for Competitiveness and Internationalisation - COMPETE2020, the North Portugal Regional Operational Program - NORTE 2020 and by the Portuguese Foundation for Science and Technology - FCT under the CMU - Portugal International Partnership and the Ph.D. grants ``2020.06434.BD'' and ``2020.07034.BD''.




\section*{Code availability}
The code is publicly available in a GitHub repository (\url{https://github.com/TiagoFilipeSousaGoncalves/survey-attention-medical-imaging}).

\end{document}